\documentclass[12pt,a4paper]{article}

\usepackage[utf8]{inputenc} % allow utf-8 input
\usepackage[T1]{fontenc}    % use 8-bit T1 fonts
\usepackage{amsfonts}       % blackboard math symbols
\usepackage{nicefrac, array}       % compact symbols for 1/2, etc.
\usepackage{amsmath, amssymb, mathrsfs, amsthm, mathabx}
\usepackage[autostyle]{csquotes}
\usepackage[sort, compress]{natbib}
\usepackage[colorlinks = true,citecolor=blue]{hyperref}
\usepackage{graphicx, comment, url, hyperref, xurl}
\usepackage{listings,wrapfig,enumitem}
\usepackage{float, parskip}
\usepackage{color, xcolor}
\usepackage[margin=1in]{geometry}
\usepackage{subfig}
\usepackage[english]{babel}
\usepackage[nodisplayskipstretch]{setspace}
\usepackage{bbm} 
\usepackage{booktabs}       % professional-quality tables
\usepackage{algorithm} 
\usepackage{algpseudocode} 
\usepackage[hang,flushmargin]{footmisc}
\usepackage[toc,page,titletoc]{appendix}
\usepackage[capitalise,nameinlink]{cleveref}
\crefname{supp}{Supplement}{Supplements}
\usepackage{multirow}
\usepackage{cellspace}
\usepackage{stackengine} 
\newcommand\oast{\stackMath\mathbin{\stackinset{c}{0ex}{c}{0ex}{\ast}{\bigcirc}}}
\usepackage[subfigure]{tocloft}
\allowdisplaybreaks
\newtheorem{thm}{Theorem}
\newtheorem{defn}{Definition}
\newtheorem{lem}[thm]{Lemma}
\newtheorem{cor}[thm]{Corollary}
\newtheorem{pro}[thm]{Proposition}

\newtheorem{rem}{Remark}
\newcommand{\M}{\mathcal{M}}

\title{Enhancing Trade-offs in Privacy, Utility, and Computational Efficiency through MUltistage Sampling Technique (MUST)}

\author{%
  Xingyuan Zhao, Ruyu Zhou, Fang Liu\footnote{Zhao was a doctoral student when this work was conducted, and Zhu is a currently doctoral student, and Liu is Professor in the Department of Applied and Computational Statistics and Mathematics of the University of Notre Dame, IN 46656, USA. Corresponding may be directed to fliu2@nd.edu.}
}
\date{}

\begin{document}
\maketitle

\begin{abstract}
Applying a randomized algorithm to a subset rather than the entire dataset amplifies privacy guarantees. We propose a class of subsampling methods ``MUltistage Sampling Technique  (MUST)'' for privacy amplification (PA) in the context of differential privacy (DP). We conduct comprehensive analyses of the PA effects and utility for several 2-stage MUST procedures through newly introduced concept including strong vs weak PA effects and aligned privacy profile. %namely, MUSTwo, MUSTow, and MUSTww that respectively represent sampling with (W)/without (O)/with (W) replacement from the original dataset in stage I and sampling without (O)/with (W)/with (W) replacement in stage II.  
We provide the privacy loss composition analysis over repeated applications of MUST via the Fourier accountant algorithm.  Our theoretical and empirical results suggest that %MUSTow and MUSTww 
MUST offers stronger PA in $\epsilon$ than the common one-stage sampling procedures including Poisson sampling, sampling without replacement, and sampling with replacement, while the results on $\delta$ vary case by case. % (MUSTwo is equivalent to sampling with replacement).  
Our experiments show that MUST is non-inferior in the utility and stability of privacy-preserving (PP) outputs to one-stage subsampling methods at similar privacy loss while enhancing the computational efficiency of algorithms that require complex function calculations on distinct data points.  MUST can be seamlessly integrated into stochastic optimization algorithms or procedures that involve parallel or simultaneous subsampling when DP guarantees are necessary.
\end{abstract}

\noindent\textbf{keywords}: 
 differential privacy, Fourier transform,  multi-stage subsampling, privacy profile, privacy loss distribution, privacy loss composition,   weak or strong privacy amplification (PA).

\newpage
\setstretch{1.05}
\section{Introduction}
\subsection{Background}\label{sec:background}
Differential privacy (DP) \cite{dwork2006calibrating,dwork2006our} is a state-of-the-art privacy notion and provides formal privacy guarantees for individuals when privacy-preserving (PP) procedures are used to release information from a dataset. If a DP procedure is applied to a subset of the data rather than to the entire dataset, its privacy guarantees can be amplified, owing to the inherent randomness introduced by subsampling.  

Some commonly used subsampling methods in the context of DP include Poisson sampling, sampling without replacement (WOR), and sampling with replacement (WR). %  A subsampling scheme that offer stronger privacy assurances if it results in a better balance between privacy and utility or if it leads to more computational efficiency while maintaining the same privacy-utility tradeoff. This is especially important
Subsampling is often used to reduce the overall privacy loss when some base DP mechanisms are applied multiple times to query the same dataset, such as iterative differentially private stochastic optimization procedures \cite{abadi2016deep} or the subsample-and-aggregate framework in DP \citep{nissim2007smooth,smith2011privacy}. Subsampling is also employed in various conventional statistical and machine learning methods outside of DP, such as subsampling bootstrap \citep{politis1994large, politis1999subsampling, bickel1997resampling,bickel2008choice} and bagging \citep{breiman1996bagging}. Should formal privacy assurances be necessary for the application of these procedures to disclose information, the inherent subsampling nature of these procedures can be utilized to enhance the privacy guarantees provided by the base DP mechanisms designed to ensure the privacy of these procedures.  %the cumulative privacy loss resulting from multiple applications of a differential privacy mechanism to data subsets will also be taken into consideration. Therefore,
                                        
\subsection{Related Work}
%Work that quantifies the privacy guarantees of a DP mechanism on subsampled data exists. 
The  privacy amplification (PA) effect of subsampling is first studied in \citep{smith2009differential} for a specific case. It shows that a $(1,0)$-DP mechanism preceded by Poisson subsampling with rate   $\gamma\!\in\!(0,1)$, satisfies $(2\gamma,0)$-DP, which offers PA when $\gamma\!<\!0.5$. This result is used implicitly in designing a private probabilistically approximately correct learner \citep{valiant1984theory}  in \citep{kasiviswanathan2011can}. \cite{balle2018privacy} develops a general methodology to analyze the PA effects of different subsampling methods. \cite{wang2019subsampled} provides upper bounds for amplified privacy guarantees for WOR in the R{\'e}nyi DP (RDP)  framework \citep{mironov2017renyi}. \cite{zhu2019poission} bounds the RDP amplification effect for Poisson subsampling. %Some works study subsampling under the R{\'e}nyi differential privacy (RDP) \citep{mironov2017renyi} framework, where RDP is a relaxation form of DP based on R{\'e}nyi divergence \citep{van2014renyi}. 

In terms of privacy loss compositions over multiple applications of DP mechanisms, \cite{mcsherry2007mechanism} provides the first results on bounding the privacy loss including sequential and parallel compositions. The advanced composition theorem obtains a tighter bound for $\epsilon$ under $k$-fold adaptive composition in the $(\epsilon,\delta)$-DP framework \citep{dwork2010boosting}. Various DP extensions are also proposed to allow for tighter composition on total privacy loss over multiple applications of DP mechanism, such as %$(\epsilon,\delta)$-probabilistic DP (pDP) \citep{machanavajjhala2008privacy}, 
$(\epsilon,\tau)$-concentrated DP  \citep{dwork2016concentrated}, zero-concentrated DP (zCDP) \citep{bun2016concentrated}, truncated concentrated DP (tCDP) \citep{bun2018composable}, RDP, and Gaussian DP (GDP) \citep{dong2022gaussian}. %Meanwhile, a sequence of composition theorems is developed to more precisely evaluate the overall privacy loss after compositions. 
%While these approaches derive the overall privacy loss given single $(\epsilon, \delta)$ values in each release, 

Combined with Poisson sampling, the moment accountant method \citep{abadi2016deep} achieves tighter bounds than the advanced composition results \citep{dwork2010boosting} on the overall privacy loss of the differentially private stochastic gradient descent (DP-SGD) algorithm, leveraging the composition properties of RDP. \cite{bu2020deep} derives closed-form and tighter privacy bounds of using DP-SGD and DP-Adam \citep{kingma2014adam} with Poisson subsampling than the moment accountant method in the GDP framework. %DP-SGD releases a perturbed average of the clipped gradients evaluated over a mini-batch in each iteration. 
\cite{koskela2020computing, koskela2021tight} apply the Fourier accountant algorithm to derive numerical privacy bounds for various subsampled Gaussian mechanisms based on the privacy loss distribution concept \citep{sommer2018privacy}, which are tighter than those given by the moment accountant method with Poisson sampling.

%------------------------------
                                              
\subsection{Our Work and Contributions}
We introduce a new subsampling family -- MUltistage Sampling Technique (MUST) -- for PA.  MUST is a hierarchical sampling procedure, allowing for multiple and flexible implementations by mixing different sampling schemes at various stages (e.g., WOR in stage I and WR in stage II). 

MUST is not merely a new alternative to single-step sampling procedures; it is particularly relevant for statistical and machine learning procedures that inherently involve multiple sampling stages, like bag of little bootstraps \citep{kleiner2014scalable} and double bootstrap \citep{lee1999class}. Consequently, it is essential to study the PA effects of MUST in these procedures when developing their DP versions for PP information release.

While from an implementation standpoint, MUST  can be viewed as a  sequential composition of multiple single-stage sampling procedures, it is more complex analytically and conceptually when it comes to PA effect analysis than single-stage sampling.  Analytically, MUST needs to account for the uncertainty in the composition of a subset obtained in an earlier stage and derive the final subset composition given each possible subset scenario in the earlier stages. In contrast, the final set is generated from a fixed sample set in  single-stage sampling. In other words,  there are multiplicatively more possible scenarios for the final set  and more sampling hyperparameters in MUST compared to single-stage sampling.  In addition, the sampling procedure used in each stage of MUST  does not have to be the same; and the derivation of privacy profile in a  MUST procedure needs careful consideration of the relative sizes of the subsets in different stages. Conceptually, analyzing a MUST procedure leads us to introducing two new concepts (strong vs weak PA effects and aligned privacy profile) so to better understand and more effectively explain the PA effects of MUST. 

 %Let $b$ be the intermediate subset size after the first subsampling stage from the original dataset of size $n$, and $m$ be the final released subset size after the second subsampling stage given the $b$ samples.

%We define the MUST subsampling framework and quantify the PA effects for three members of the MUST family   %In practice, the WR$(m,b)$ step in a MUST procedure can be replaced with sampling the occurrence frequency for each distinct sampled data point in the set of size $b$ via multinomial($m,\mathbf{p}$), where the probability vector $p=1/b$ is the same over the $b$ data points (though some of the data points are the same if the 1st step is also WR).
We derive the privacy loss profiles for $(\mbox{MUST}\circ\M)^k$ for three 2-stage MUST procedures (MUSTwo: WR/WOR in stage I/II, respectively; MUSTow: WOR/WR in stage I/II;  MUSTww: WR in both stages) when they are composed with a general base randomized mechanism $\M$ for $k\ge1$ times. %using the conceptual framework of privacy loss distribution \citep{sommer2018privacy}. 
We present numerical results on the privacy loss profiles for  Laplace and  Gaussian mechanisms $\M$, respectively. %obtained via the Fourier accountant algorithm \citep{koskela2020computing, koskela2021tight}. 
We conduct experiments to compare the utility of $(\mbox{MUST}\circ\M)^k$ in prediction and statistical inference to some commonly used one-stage subsampling methods. Our contributions and the main findings are summarized below.

\begin{itemize}%[leftmargin=9pt]
\item We propose the concepts of \emph{strong vs weak (Type I and Type II) PA} for a subsampling procedure, based on the impact the procedure has on $\epsilon$ and $\delta$ in the framework of $(\epsilon,\delta)$-DP.  We also propose \emph{aligned privacy profile} to effectively capture the PA effect of a subsampling procedure on a base mechanism of $(\epsilon,\delta)$-DP.
\item We show theoretically and empirically that MUSTww and MUSTow offer stronger PA effects on $\epsilon$ for baseline mechanisms of $\epsilon$-DP and $(\epsilon,\delta)$-DP compared to Poisson subsampling,  WOR, and WR.  We prove MUSTwo and WR are equivalent subsampling procedures and have the same PA effects. 
%\item  We derive the privacy loss distribution of the subsampled Gaussian mechanism with MUST and use the Fourier Accountant algorithm to numerically evaluate the overall privacy loss bounds over multiple applications of MUST and a DP mechanism (e.g., DP-SGD algorithms or procedures that use parallel/simultaneous subsamples such as bagging, subsampled bootstrap, or random forest.).
\item We show that MUST-subsampled Gaussian mechanisms generate PP outputs of similar or better utility and stability compared to Poisson sampling, WR, and WOR at similar privacy loss.
\item MUSTow and MUSTww offer computational advantages as the final subset they generate contains notably fewer distinct data points on average than those of generated by Poisson sampling or WOR. This will improve the computational efficiency of algorithms that evaluate complex functions at each distinct data point\footnote{e.g.  stochastic gradient descent via back-propagation in deep neural networks, where high-dimensional gradients are calculated per data point; other examples  included likelihood evaluation in Bayesian statistics and the Expectation-Maximization algorithm, Gaussian processes.}.
\end{itemize}%
By providing a broader range of subsampling options compared to single-step procedures, MUST has the potential to enhance the trade-offs between privacy, utility, and computational efficiency, as shown in the rest of this work.

%In summary, MUST can be a viable and practical alternative for subsampling in the DP setting, alongside the current subsampling methods like Poisson, WR, and WOR, to achieve PA while offering similar or better utility and computational advantages. 

%The rest of the paper is organized as follows. Sec.~\ref{sec:prelim} reviews the DP concepts used in this work. Sec.~\ref{sec:method} defines the MUST procedure, examines its PA effect analytically, and derives the privacy loss distribution of the MUST-subsampled Gaussian mechanism after $k$-fold composition and presents the FA algorithms for calculating privacy loss composition. Sec.~\ref{sec:experiment} runs multiple experiments to examine the PA of the 2-stage MUST, the sensitivity of the  MUST procedures to its hyperparameters, and the utility and stability of PP inferential and prediction tasks. The paper concludes in Sec.~\ref{sec:conclusion} with some final remarks. 

\section{Preliminaries}\label{sec:prelim}
%\subsection{Differential Privacy (DP) and Privacy Profile}\label{sec:dp}\vspace{-3pt}

\begin{defn}[$(\epsilon,\delta)$-DP \citep{dwork2006calibrating,dwork2006our}]\label{defn:dp}
A randomized algorithm $\M$ is of $(\epsilon,\delta)$-DP if for any pair of neighboring data sets $X\simeq X'$ and any measurable subset $\mathcal{O}\subseteq$ image$(\M)$,
\begin{equation}\label{eqn:DP}
\Pr(\mathcal{M}(X)\in \mathcal{O}) \leq e^{\epsilon} \Pr(\mathcal{M}(X')\in \mathcal{O})+\delta.
\end{equation}
\end{defn} 
Two datasets $(X, X')$ are neighbors, denoted as $X\simeq X'$, if they differ by one element.  The neighboring relations include remove/add  $X\simeq_RX'$ ($X$ can be obtained by adding or removing one element from $X'$) and substitute $X\simeq_SX'$. $X\simeq_RX'$ ($X$ can be obtained by substituting an element in $X'$ with a different element). The sizes of two neighboring datasets under  $X\simeq_RX'$ are different by one, while those under  $X\simeq_SX'$ are the same. We focus on the PA effects under $X\simeq_SX'$ in Section \ref{subsec:amplification}.

$\epsilon$ and $\delta$ in Eqn \eqref{eqn:DP} are the privacy loss parameters. When $\delta=0$, $(\epsilon,\delta)$-DP reduces to $\epsilon$-DP (pure DP). %$\delta$ can be interpreted as the probability that pure DP is violated. 
The smaller $\epsilon$ and $\delta$ are, the stronger the privacy guarantees there are for the individuals in a dataset when output from $\mathcal{M}$ is released.
% In other words, one cannot distinguish the outputs released by a differentially private mechanism based on neighboring datasets to some extent. 
A randomized algorithm $\M$ is called \emph{tightly $(\epsilon, \delta)$-DP if there is no $\delta'<\delta$ such that $\M$ is $(\epsilon, \delta')$-DP} \citep{sommer2018privacy}. 

$(\epsilon, \delta)$-DP can be expressed in terms of $\alpha$-divergence $D_{\alpha}$ between two probability measures \citep{barthe2013beyond}\footnote{\hspace{1pt}$\alpha$-divergence ($\alpha\geq1$) measures the distance between two probability measures $\mu, \mu'\in\mathbb{P}(\mbox{image}(\mathcal{M}))$ and is defined as  $D_\alpha(\mu||\mu')=\sup_\mathcal{O}(\mu(\mathcal{O})-\alpha\mu'(\mathcal{O}))$, where $\mathcal{O}$ is any measurable subset of $\mbox{image}(\mathcal{M})$.}. A randomized algorithm $\M$ is of $(\epsilon,\delta)$-DP if and only if 
\begin{equation}
D_{e^\epsilon}(\mathcal{M}(X)||\mathcal{M}(X'))\leq\delta.
\end{equation}
In other words, the $e^\epsilon$-divergence between the outputs from the randomized algorithm $\M$ run on two neighboring datasets is upper bounded by $\delta$. Based on the $\alpha$-divergence formulation of $(\epsilon,\delta)$-DP,  \cite{balle2018privacy}  defines the \emph{privacy profile}  
\begin{equation}\label{eqn:profile}
\delta_{\mathcal{M}}(\epsilon)=\textstyle\sup_{X\simeq X'}D_{e^\epsilon}(\mathcal{M}(X)||\mathcal{M}(X'))
\end{equation}
that represents the set of all privacy loss parameters $(\epsilon,\delta)$ and depicts their relations in the the domain of $[0,\infty)\times[0,1]$.

Theorem \ref{thm:LapGau} lists the functional privacy profiles of the Laplace \citep{dwork2006calibrating}  and Gaussian mechanisms \citep{dwork2014algorithmic, liu2018generalized, balle2018improving}, respectively, which are two commonly used DP mechanisms.

\begin{thm}\label{thm:LapGau}\textbf{(privacy profiles of Laplace and Gaussian mechanisms)} 
Let $f$ be an output function with $\ell_1$ global sensitivity $\Delta_1$ and $\ell_2$ global sensitivity $\Delta_2$\footnote{\hspace{1pt}The $\ell_p$ global sensitivity \citep{liu2018generalized} of a function  $f: X\to \mathbb{R}^d$ is $\Delta_p(f)=\sup_{X\simeq X'} ||f(X)-f(X')||_p\mbox{ for } p>0.$ The $\ell_1$ and $\ell_2$ sensitivities are special cases of the $\ell_p$ sensitivity.}. For any $\epsilon>0$, \newline
\emph{(a)} \citep{balle2018privacy} the privacy profile of the Laplace mechanism $\mathcal{M}: f(X)+\mbox{Lap}(0,\sigma)$ with scale parameter $\sigma$ is $\delta_{\mathcal{M}}(\epsilon)=[1-\mbox{exp}((\epsilon-\theta)/2)]_{+}$, where $\theta=\Delta_1/\sigma$ and  function $[\cdot]_{+}$ takes the maximum of the input value and $0$.\newline
\emph{(b)} \citep{balle2018improving}  the privacy profile of the Gaussian mechanism $\mathcal{M}: f(X)+\mathcal{N}(0, \sigma^2 I)$ is $\delta_{\mathcal{M}}(\epsilon)=\Phi(\theta/2-\epsilon/\theta)-e^\epsilon\Phi(-\theta/2-\epsilon/\theta)$, where $\theta=\Delta_2/\sigma$ and $\Phi(\cdot)$ is the cumulative distribution function of $\mathcal{N}(0,1)$. \cite{dwork2014algorithmic} presents a different  privacy profile  for the Gaussian mechanism of  $(\epsilon, \delta)$-D, where the scale parameter is $\sigma^2\geq2\Delta_2^2\log(1.25/\delta)/\epsilon^2$  for $\epsilon<1$. Comparatively, the Gaussian mechanism in  \cite{balle2018improving} is exact and does not constrain  $\epsilon$ to be $<1$.
\end{thm}
%by adding Laplace noise and  Gaussian noise, respectively, that are calibrated to the $\ell_1$ and $\ell_2$ GS of an output function $f$. 
Theorem \ref{thm:LapGau} suggests that the Laplace mechanism can achieve  $(\epsilon, \delta>0)$-DP when $\delta=\delta_{\mathcal{M}}(\epsilon)$ for $\epsilon<\Delta_1/\sigma$, and it includes the classical result that the Laplace mechanism with $\sigma\geq\Delta_1/\epsilon$ is of $(\epsilon,0)$-DP \citep{dwork2006calibrating} given $\delta_{\mathcal{M}}(\epsilon)=0$ for any $\epsilon\geq\theta$.

The \emph{group privacy profile} of $\mathcal{M}$, as defined in Eqn \eqref{eqn:group}, when two datasets differ in $j\ge1$ elements, that is, $d(X,X')\leq j$, is a natural extension of the privacy profile defined for two datasets differing in one element ($d(X,X')=1$) \citep{balle2018privacy}. 
\begin{equation}\label{eqn:group}
\delta_{\mathcal{M}, j}(\epsilon)=\textstyle\sup_{d(X,X')\leq j}D_{e^\epsilon}(\mathcal{M}(X)||\mathcal{M}(X')).
\end{equation}
$\delta_{\M,j}(\epsilon)$ increases with $j$ and may reach  $1$ for large $j$. For an arbitrary $\mathcal{M}$, \cite{vadhan2017complexity} derives an upper bound on the group privacy profile $\delta_{\mathcal{M}, j}(\epsilon)\leq((e^\epsilon-1)/(e^{\epsilon/j}-1))\delta_\mathcal{M}(\epsilon/j)$, known as the \emph{black-box} group privacy analysis   \citep{balle2018privacy}. Given  a specific mechanism, one may apply the \emph{white-box} group privacy analysis by factoring in the mechanism information to obtain tighter privacy loss bounds than the black-box analysis. For the Laplace and Gaussian mechanisms, the group privacy profiles can be obtained by substituting $\Delta_p$ in Theorem \ref{thm:LapGau} with $j\Delta_p$ for $p=1,2$ \citep{balle2018privacy}. We apply the white-box analysis when examining the PA effect of a subsampling scheme that outputs a \emph{multiset} with multiple copies of the same element in the subsampled data space where a DP mechanism $\mathcal{M}$ is performed.

\begin{defn}[privacy loss random variable \citep{dwork2016concentrated}]\label{defn:PRV}
Let $\M$ be a randomized algorithm, $f_{X}(t)$ and $f_{X'}(t)$  denote the  probability density functions (pdf) of  output $t$ of $\M(X)$ and $\M(X')$, respectively, where $X\!\simeq\!X'$. The privacy loss random variable with respect to $f_{X}$ over $f_{X'}$  is  \vspace{-3pt}
\begin{equation}
\mathcal{L}_{X/X'}(t)=\log\left(\frac{f_{X}(t)}{f_{X'}(t)}\right). 
\end{equation}
\end{defn} 
$\mathcal{L}_{X'/X}$, the privacy loss random variable with respect to $f_{X'}$ over $f_{X}$, can be defined in a similar manner to Definition \ref{defn:PRV}; that is, $\mathcal{L}_{X'/X}(t)=\log\left(\frac{f_{X'}(t)}{f_{X}(t)}\right)$.
%The PLD for a differentiable privacy loss function $\mathcal{L}_{X/X'}$  is given in Definition \ref{defn:PLD}

\begin{defn}[{privacy loss distribution (PLD)} \citep{sommer2018privacy}] \label{defn:PLD}
If $\mathcal{L}_{X/X'}$ is a differentiable bijective function and  the derivative of the inverse function  $\mathcal{L}^{-1}$ is integrable, then the pdf of the privacy loss random variable $s$ is %of $\M(X)$ over $\M(X')$ has the density
\vspace{-3pt}
    \begin{align*}
        \omega_{X/X'}(s)=
        \begin{cases}
        f_{X}(g(s))g'(s),  & s\in\mathcal{L}_{X/X'}(\mathbb{R})\\
        0, & \mbox{otherwise}
        \end{cases},  
    \end{align*}
where $g(s)=\mathcal{L}_{X/X'}^{-1}(s)$ is the inverse of function $\mathcal{L}_{X/X'}(t)$.   
\end{defn} 
The PLD of the cumulative privacy loss from applying $k$ DP mechanisms (the $k$-fold composition) is the convolution of $k$ PLDs $\omega_{1}*\cdots*\omega_{k}$, where $\omega_{i}$ represents the PLDs of the $i$-th DP procedure for $i=1,\ldots,k$. Similarly, we can define the PLD $\omega_{X/X'}(s)$ of privacy loss random variable $s\in\mathcal{L}_{X'/X}(\mathbb{R})$ by switching $X$ and $X'$ in Definition \ref{defn:PLD}.  $\omega_{X/X'}$ and $\omega_{X'/X}$ are not equivalent.  Lemma \ref{lem:omega} depicts the relations between the two.
\begin{lem}[relation between $\omega_{X/X'}$ and $\omega_{X'/X}$\citep{sommer2018privacy}]\label{lem:omega}
Under $k$-fold convolution for $k\ge1$, $(\omega_{1, X/X'}\!*\!\cdots*\!\omega_{k, X/X'})(s)\!=\!e^s(\omega_{1, X'/X}\!*\cdots*\!\omega_{k, X'/X})(-s)$.
\end{lem}
Based on the above definitions and results, one may define the privacy profiles of a DP mechanism $\M$ and the composition of $k$ DP mechanisms, respectively (Theorem \ref{thm:delta_kfold}).

\begin{thm}[privacy profile in $k$-fold composition \citep{sommer2018privacy}\hspace{-3pt}]\label{thm:delta_kfold}
%Suppose $(\epsilon, \infty)\subset\mathcal{L}_{X/X'} (\mathbb{R})$.
\textit{(a)} let $\M$ be a randomized mechanism with  corresponding PLDs $\omega_{X/X'}$ and $\omega_{X'/X}$. $\M$ is tightly $(\epsilon,\delta(\epsilon))$-DP with $\delta(\epsilon)=\max\{\delta_{X/X'}(\epsilon),\delta_{X'/X}(\epsilon)\}$, where $\delta_{X/X'}(\epsilon)=\int_\epsilon^\infty(1-e^{\epsilon-s})\omega_{X/X'}(s)ds$ and $\delta_{X'/X}(\epsilon)=\int_\epsilon^\infty(1-e^{\epsilon-s})\omega_{X'/X}(s)ds$. \newline
\textit{(b)} let $\M_1, \dots, \M_k$ be $k$ randomized mechanisms with  corresponding PLDs $\omega_{i, X/X'}$ and $\omega_{i, X'/X}$ for $i=1,\ldots,k$. The composited mechanism $\M=\M_1\circ\cdots\circ\M_k$  is tightly $(\epsilon,\delta(\epsilon))$-DP with  
$\delta(\epsilon)=\max\{\delta_{X/X'}(\epsilon),\delta_{X'/X}(\epsilon)\},$
where
$\delta_{X/X'}(\epsilon)=\int_\epsilon^\infty(1-e^{\epsilon-s})(\omega_{1, X/X'}*\cdots*\omega_{k, X/X'})(s)ds$, and $\delta_{X/X'}(\epsilon)=\int_\epsilon^\infty(1-e^{\epsilon-s})(\omega_{1, X'/X}*\cdots*\omega_{k, X'/X})(s)ds$.
\end{thm} 
%In the context of privacy loss random variable $\mathcal{L}$ and its pdf  $\omega$, 
For $\M$ to be tightly $(\epsilon, \delta(\epsilon))$-DP, the supports of its privacy loss random variables $\mathcal{L}_{X/X'}(t)$ and $\mathcal{L}_{X'/X}(t)$ should contain $(\epsilon, \infty)$, and $\delta(\epsilon)=\max\{\delta_{X/X'}(\epsilon), \delta_{X'/X}(\epsilon)\}$ (see Sec.~4.4 in \cite{sommer2018privacy} and Lemma 5 in \cite{koskela2020computing}).

\vspace{-3pt}\section{MUST Subsampling and Its PA Effect}\label{sec:method}
Applying a DP mechanism $\mathcal{M}$  to a random subsample rather than the entire dataset amplifies the privacy guarantees of $\mathcal{M}$ as there is a non-zero probability that an individual does not appear in the subsample. Formally speaking, one first performs subsampling $\mathcal{S}$ on a dataset $X$ of size $n$ to obtain a subset $Y$ of size $m$ ($m\leq n$), i.e., $\mathcal{S}: X\to\mathbb{P}(Y)$, and then applies $\M: Y\to\mathbb{P}(Z)$ to the subset $Y$ and release privatized information $Z$. Suppose $\mathcal{M}$ is $(\epsilon, \delta)$-DP, the composed mechanism $\mathcal{M}\circ\mathcal{S}$ satisfies $(\epsilon', \delta')$-DP with $\epsilon'\leq\epsilon$ and $\delta'$ being a function of $\delta$. \cite{balle2018privacy} provides a general framework to quantify PA by subsampling using techniques including couplings, advanced joint convexity, and privacy profiles, and derive the PA results for Poisson sampling, sampling WOR, and sampling WR.

\begin{figure}[!htb]
\vspace{-12pt}\centering
\includegraphics[width=0.65\textwidth]{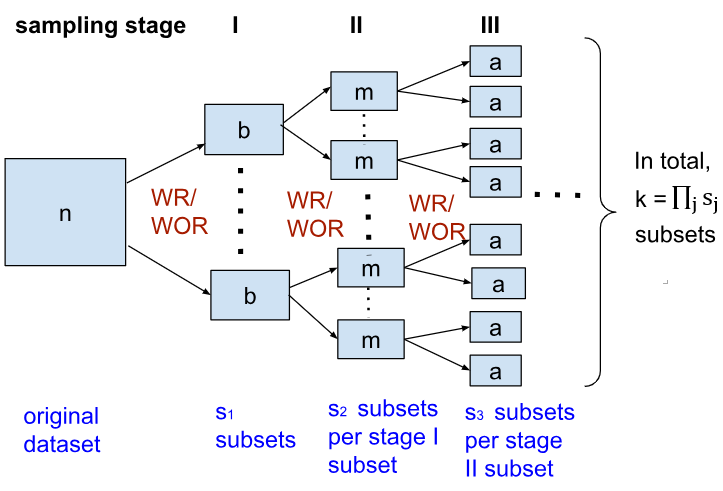}
\caption{A general MUST procedure. WR and WOR refer to sampling with and without replacement, respectively; the number in each box is the respective dataset size. }\label{fig:MUST}
\end{figure}

We introduce a hierarchical subsampling procedure MUST (MUlti-stage sUbsampling technique) for PA, as depicted in Fig.~\ref{fig:MUST}. In stage I, $s_1$ sets of subsamples of size $b$ are drawn from the original data of size $n$; in stage II,  $s_2$ subsamples of size $m$ are drawn in each of the $s_1$ subsets, so on so forth, leading to a  total of $\prod_j s_j$ subsets of samples.
%inspired by the subsampling bootstrap technique \citep{politis1999subsampling}. 

The motivation behind MUST is twofold. Firstly, incorporating additional stages into the subsampling process expands the subsampling configuration and thereby alters the composition of the subsamples, permitting greater flexibility in the trade off among privacy, utility, and computational efficiency. Second, certain statistical or machine learning techniques inherently involve multi-stage sampling, such as the double bootstrap procedure \citep{lee1999class}  for confidence interval construction and the BLB procedure \citep{kleiner2014scalable} for computationally efficient bootstrap in large-scale data. The PA effects of these multi-stage sampling procedures will need to be studied and leveraged when DP versions of these procedures are developed to release PP information. 

%----------------------------------------------
\vspace{-12pt}
\subsection{2-stage MUST}
\begin{figure}[!htb]
\centering\includegraphics[width=0.65\textwidth]{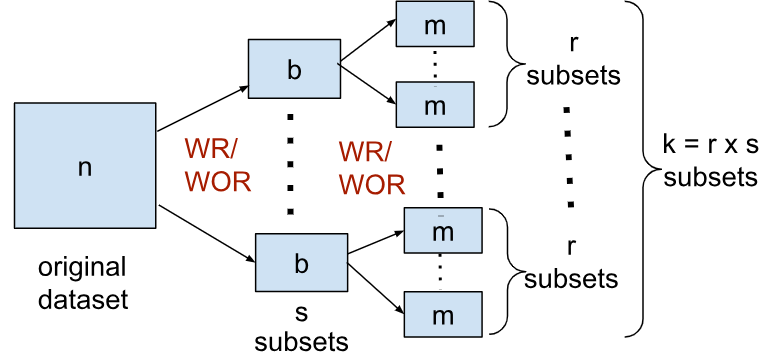}
%\subfloat[\centering a special case of 2-step MUST procedure with $r=1$] {{\includegraphics[width=0.35\textwidth]{figures/MUST2s.PNG}}}\\%

\caption{2-stage MUST procedure (numbers in boxes are dataset sizes)}\label{fig:MUST2}\vspace{-12pt}
\end{figure}

We introduce the 2-stage MUST procedure in this section (Fig.~\ref{fig:MUST2}) and briefly discuss MUST with $>2$ stages in Section \ref{sec:MUST3}. 
% While a MUST procedure may encompass more than two stages, from the practicality,  PA, and utility perspectives, 2-stage MUST would be sufficient, which is the focus of this work (Fig.~\ref{fig:MUST2}). 
We focus on three types of 2-stage MUST. % --  MUSTwo (WR in the first stage and WOR in the second stage), MUSTow (WOR in the first stage and WR in the second stage) and MUSTww (WR in both the first and second stages). MUSToo  -- WOR in both stages -- is a trivial case and not examined as it yields the same subset as WOR. Each MUST scheme can be expressed as a composition of two subsampling schemes. For example, $\mbox{MUSTwo}(n, b, m)=\mbox{WR}(n,b)\circ\mbox{WOR}(b,m)$), where  $n$ is the original dataset size, $b<n$ is the subset size in the first stage, and $m$ is the subset size in the second stage. 
We suggest $m < n$ (or $\ll n$), a large $s$\footnote{\hspace{1pt}For example, in SGD, $s$ is the number of iterations for the algorithm to converge, often in the thousands for complex models}, and $r=1$ for MUST procedures to be used in procedures like stochastic optimization and double bootstrap.\footnote{\hspace{1pt} The MUST procedure in Fig.~\ref{fig:MUST2} includes the sampling scheme in the bag of bootstrap (BLB) procedure as a special case when Stage I is WOR and Stage II is WR. However, the specification of hyperparameter $s, r,b$, and $m$ in the regular BLB procedure can be very different given that the BLB procedure was developed to efficiently obtain bootstrap-based statistical inference in large-scale data and driven by this, $m$ is set to be $n$, $s$ is usually small, and $r$ is often large, depending on the value of $b$, in a BLB procedure. In the experiments in \cite{kleiner2014scalable}, $s$ ranges from 1$\sim$2 when $b=n^{0.9}$ to 10 $\sim$ 20 when $b=n^{0.6}$ and $r$ is often on the scale of 50 to 100.  That said, the PA effect of a typical BLB procedure can be examined in a similar matter as presented in the rest of the paper, a subject that will be explored in future research.} There are 4 possible 2-stage MUST procedures, as listed in Definition  \ref{defn:MUST}. %Tab.~\ref{tab:MUST}. 
%The formal definition of 2-stage MUST is provided in Definition  \ref{defn:MUST} and a summary of 4 possible MUST procedures are provided in Tab.~\ref{tab:MUST}.

\begin{defn}[2-step MUST]\label{defn:MUST}
MUST$(n,b,m): X\to\mathbb{P}(Y)$ is a sampling scheme to release a multiset $Y$ of size $m$ from a dataset $X$ of size $n$ via a two-step procedure:\\ 
(a) MUSTow$(n,b,m)$ first samples $b$ ($b<n$) points without replacement from $X$, then samples the occurrence frequency of each distinct data point obtained in the first step via Multinomial$(m,\mathbf{1}_b/b)$ ($\mathbf{1}_b$ is a vector with $b$ entries of 1).  \\
(b) MUSTww$(n,b,m)$ first draws $b$ points with replacement from $X$ of size $n$, and then draws $m$ points with replacement from the $b$ points sampled in the first step. \\
(c)  MUSTwo$(n,b,m)$ first draws $b$ points with replacement from $X$, and then samples $m$ ($m<b$) points without replacement from the subset obtained in step 1. MUSTwo$(n, b, m)$ is equivalent to WR($n,m$) (see Cor.~\ref{cor:MUSTwo=WR}).\\
(d) MUSToo$(n, b, m)$ is equivalent to WOR($n,m$) and thus not discussed.
\end{defn}

%---------------------------------------
\subsection{PA via 2-step MUST}\label{subsec:amplification}
Suppose $\M$ is a base mechanism of $(\epsilon, \delta)$-DP. If $\M$ is applied to a subset from a dataset obtained via a subsampling scheme $\mathcal{S}$, then $\mathcal{S}\!\circ\!\M$ is of $(\epsilon', \delta')$-DP, where $\epsilon'$ is a function of $\epsilon$ and the hyperparameters associated with $\mathcal{S}$, and $\delta'$ is a function of $\epsilon,\delta$, and the hyperparameters in $\mathcal{S}$.  Contrasted with single-step sampling procedures, MUST introduces more randomness through additional sampling stages, potentially enhancing privacy guarantees.  We derive the privacy profiles $(\epsilon', \delta')$ of MUSTow, MUSTwo, and MUSTww below when they are composited with a general base mechanism $\M$ of $(\epsilon, \delta)$-DP under the $\simeq_{S}$ relation for both $\mathcal{M}$ and $\mathcal{S}$.  
The main results are given in Theorem \ref{thm:RUB}. Corollary \ref{cor:MUSTwo=WR} shows that MUSTwo$(n,b,m)$ and WR$(n,m)$ are equivalent in their PA effect though they differ in how subsampling is implemented. The proofs are provided in the  Supplementary Materials (Suppl. Mat.).  

\begin{thm}[privacy profile of $\M\circ$MUST]\label{thm:RUB}
Let $\mathcal{M}$ be a $(\epsilon, \delta)$-DP mechanism.  For any $\epsilon\!\!>\!\!0$, with respect to the $\simeq_{S}$ relation for $\mathcal{M}$ and MUST,  $\mathcal{M}\circ \mbox{MUST}(n,b,m)$ is of $(\epsilon', \delta')$-DP, where $\epsilon'\!=\!\log(1+\eta(e^\epsilon-1))$ and for different MUST schemes, the $\eta$ and $\delta'$ are listed in Table \ref{tab:amplification}.

\begin{table}[!htb]
\vspace{-3pt}\centering
\caption{PA for MUSTwo, MUSTow and MUSTww on randomized mechanism $\mathcal{M}$ of $(\epsilon,\delta)$-DP. $\epsilon'=\log(1\!+\!\eta(e^\epsilon\!-\!1))$ for all sampling schemes. }\label{tab:amplification} \vspace{-3pt}
\resizebox{1\textwidth}{!}{
\begin{tabular}{@{}c@{}c@{\hspace{3pt}}c}
\toprule
$\mathcal{S}$&  $\eta$ & $\delta'$  \\
\midrule
\shortstack[c]{MUSTwo\\($n,b,m$)}  &\shortstack[c]{\small{$\displaystyle\sum_{j=1}^b\binom{b}{j}\left(\frac{1}{n}\right)^j\left(1-\frac{1}{n}\right)^{b-j}$}\\\small{$\times\left(1-\frac{\binom{b-j}{m}}{\binom{b}{m}}\right)$}}&  \shortstack[c]{\small{$\displaystyle\sum_{j=1}^m\binom{b}{j}\left(\frac{1}{n}\right)^j\left(1-\frac{1}{n}\right)^{b-j}\sum_{u=1}^j\frac{\binom{j}{u}\binom{b-j}{m-u}}{\binom{b}{m}}\delta_u$}\\\small{$\displaystyle+\!\!\!\sum_{j=m+1}^b\binom{b}{j}\left(\frac{1}{n}\right)^j\left(1-\frac{1}{n}\right)^{b-j}\!\sum_{u=1}^m\frac{\binom{j}{u}\binom{b-j}{m-u}}{\binom{b}{m}}\delta_u$}}\\ 
& $=1-(1-\frac{1}{n})^m$ & \small{=$\displaystyle\sum_{j=1}^m\! \binom{m}{j}\!\left(\!\frac{1}{n}\!\right)^j\!\left(1-\frac{1}{n}\right)^{m-j}\delta_j$}\\
\hline
\shortstack[c]{\\MUSTow\\($n,b,m$)}  &\shortstack[c]{$\frac{b}{n}(1\!-\!(1\!-\!\frac{1}{b})^m)$} & \shortstack[c]{\small{$\displaystyle\frac{b}{n}\sum_{j=1}^m \!\binom{m}{j}\!\left(\!\frac{1}{b}\!\right)^j\!\left(\!1\!-\!\frac{1}{b}\!\right)^{m-j}\!\!\!\delta_j$}} \\
\hline
\shortstack[c]{MUSTww\\($n,b,m$)}   &\shortstack[c]{\small{$\sum_{j=1}^b\binom{b}{j}\left(\frac{1}{n}\right)^j\left(1-\frac{1}{n}\right)^{b-j}$}\\
\small{$\times\left(1-\left(1-\frac{j}{b}\right)^m\right)$}} &  \shortstack[c]{\small{$\sum_{j=1}^b\binom{b}{j}\left(\frac{1}{n}\right)^j\left(1-\frac{1}{n}\right)^{b-j}$}\\\small{$\times\left(\sum_{u=1}^m\binom{m}{u}\left(\frac{j}{b}\right)^u\left(1-\frac{j}{b}\right)^{m-u}\delta_u\right)$}}\\
\midrule
\multicolumn{3}{p{1\linewidth}}{\footnotesize $\eta\in(0,1)$ is the probability that an element appears in the final subset.
$\delta'$ for WR and MUST involve group privacy terms $\delta_j$ and $\delta_u$ as they all output multisets.
For MUSTwo and MUSTww, $j$ is the number of an element after subsampling $b$ elements
in stage I  ($1\leq j\leq b$) and $u$ is the number of an element in the final subsample. 
The two terms in the summation of $\delta'$ for MUSTwo refers to the cases of  $j\leq m$ and 
$j>m$,  respectively  (when $j\leq m$, $1\leq u\leq j$; when $j>m$, $1\leq u\leq m$).
$\binom{b-j}{m}=0$ 
if $b\!-\!j\!<\!m$ in $\eta$ and  $\binom{b-j}{m-u}=0$ if $b\!-\!j\!<\!m\!-\!u$ in $\delta'$.}\\
\bottomrule
\end{tabular}}
\end{table} 
\end{thm}

\begin{cor}[equivalence between MUSTwo and WR]\label{cor:MUSTwo=WR}
MUSTwo$(n,b,m)$ and WR$(n,m)$ are equivalent subsampling procedures in terms of generated samples and their PA effects with the $X\simeq_SX'$  relation; i.e., $\eta_{\text{MUSTwo}}=\eta_{\text{WR}}$ and $\delta'_{\text{MUSTwo}}=\delta'_{\text{WR}}$.
\end{cor}
We focus on the $\simeq_{S}$ neighboring relation in $\mathcal{M}$ and $\mathcal{S}$ for the following reasons. For  $\mathcal{M}$, since the final subset size $m$ in the three MUST procedures, to which $\mathcal{M}$ is applied, is fixed, it only makes sense when the relation considered for $\mathcal{M}$ is substitution as removal implies the neighboring sets are of different size. This would not affect the generalization of the results as many base mechanisms are applicable in both neighboring relation cases and substitution is sufficient when it applies to DP applications and interpretation. As for $\mathcal{S}$, Both MUSTow and MUSTwo involve WOR, the PA effect of which is more sensible with substitution due to the reasons outlined in \cite{balle2018improving} (i.e., protecting the original data size $n$ if needed and the possibility to derive a meaningful privacy profile for $\mathcal{S}$ across all inputs sets). As for MUSTww, it involves WR in both stages; the mathematical formulations of its PA effect on both $\epsilon'$ and $\delta'$ are the same regardless of whether the relation is removal or substitution \cite{balle2018improving}, implying that our results on  MUSTww derived under substitution should also be applicable to the removal relation.  %Finally, focusing on the substitution relation   \citet{balle2018privacy} only presents the results when both $\mathcal{M}$ and $\mathcal{S}$ are based on the substitution relation WR and WOR

%since the involved WOR and WR steps in MUST are known for their PA results concerning the substitution relation . %\cite{balle2018privacy} also notes that some subsampling methods and neighboring relation combinations can be more natural than others, e.g., the remove/add relation is typically selected for analyzing the PA effect of Poisson sampling, and the substitution relation is typically used for analyzing the PA effect of WOR. 
For easy comparison, we also list the existing PA results for Poisson sampling, WOR, and WR in Table \ref{tab:existing}. 
For WR and MUST, $\delta'$ is the weighted summation of $\delta_j(\epsilon)$ for $j=1, 2, \dots, m$, where the weight equals the probability of having $j$ copies of an element in the final subset. For example, the probability of sampling an element without replacement from the original dataset $X$ of size $n$ is $b/n$ in the WOR stage of MUSTow and the probability that this element is sampled at least once with replacement in $m$ trials from the subset obtained in the first stage is $\left(1-\left(1-\frac{1}{b}\right)^m\right)$ in the WR stage; the product of the two probabilities yields the final weight $\frac{b}{n}\binom{m}{j}\left(\frac{1}{b}\right)^j\left(1-\frac{1}{b}\right)^{m-j}$ for $\delta_j$.  
\begin{table}[!htb]
\vspace{-8pt}\centering
\setlength{\tabcolsep}{1pt}
\caption{PA for Poisson sampling, WOR, and WR on $\mathcal{M}$ of $(\epsilon,\delta)$-DP \cite{balle2018improving}). $\epsilon'=\log(1\!+\!\eta(e^\epsilon\!-\!1))$ for all sampling schemes.  }\label{tab:existing} \vspace{-3pt}
\resizebox{0.75\textwidth}{!}{
\begin{tabular}{lcc}
\toprule
$\mathcal{S}$ & $\eta$  & $\delta'$  \\
\midrule
Poisson($\eta$) ($X\simeq_RX'$) & -- & $\eta\delta$  \\
%\cline{1-1}\cline{3-4}
WOR ($X\simeq_SX'$) & $m/n$ & $\eta\delta$  \\
%\cline{1-1}\cline{3-4}
WR ($X\simeq_SX'$) &$(1\!-\!\frac{1}{n})^m$ & $\sum_{j=1}^m\! \binom{m}{j}\!\left(\!\frac{1}{n}\!\right)^j\!\left(1\!-\!\frac{1}{n}\right)^{m-j}\delta_j$\\
\bottomrule
\multicolumn{3}{p{0.8\linewidth}}{\footnotesize The PA  result of Poisson  first appears in \cite{li2012sampling} and can be reproduced
using the method in \cite{balle2018privacy}.}\\
\bottomrule
\end{tabular}}
\end{table}

\begin{figure}[!htb]
\centering
{\includegraphics[width=0.5\textwidth]{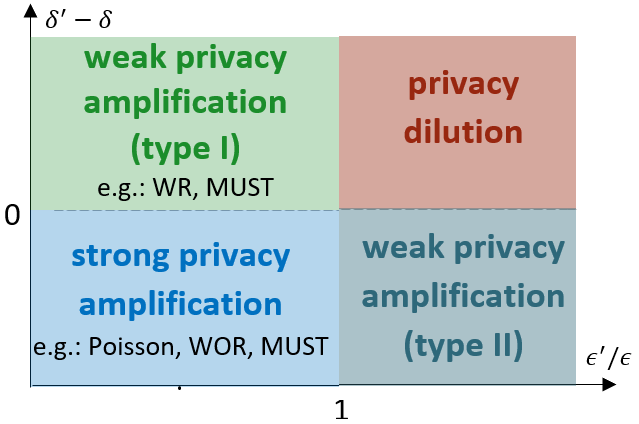}}\\

\caption{PA effect of a sampling procedure $\mathcal{S}$ on a base randomized mechanism, When $\epsilon'/\epsilon<1$ and $\delta'-\delta<0$, $\mathcal{S}$ yields strong PA; when one of the conditions is not satisfied, $\mathcal{S}$ yields weak PA; when both conditions fail, $\mathcal{S}$ leads privacy dilution.} \label{fig:PAeffect}
\vspace{-20pt}\end{figure}

\begin{defn}[PA types]\label{def:PA}
Let $\mathcal{M}$ be a  base mechanism of $(\epsilon, \delta)$-DP, $\mathcal{S}$ be a subsampling scheme, and the composition of $\mathcal{S}\circ\M$ satisfies $(\epsilon',\delta')$-DP.  
\small

\noindent\emph{(a)} If $\epsilon'\!<\!\epsilon$ and $\delta'(\epsilon')\!<\!\delta(\epsilon)$, then $\mathcal{S}$ offers \textit{strong PA}; 

\noindent\emph{(b)} if $\epsilon'\!<\!\epsilon$ and  $\delta'(\epsilon')\!>\!\delta(\epsilon)$, then $\mathcal{S}$ has a  \textit{type I weak  PA} effect;  

\noindent\emph{(c)} if $\epsilon'\!>\!\epsilon$ and  $\delta'(\epsilon')\!<\!\delta(\epsilon)$, then $\mathcal{S}$ has a  \textit{type II weak  PA} effect; 

\noindent\emph{(d)} if $\epsilon'\!>\!\epsilon$ and  $\delta'(\epsilon')\!>\!\delta(\epsilon)$, then $\mathcal{S}$ has a  \textit{privacy dilution} effect. 
\end{defn}
\normalsize
Fig.~\ref{fig:PAeffect} summarizes four possible scenarios that a sampling procedure may have on the privacy guarantees of a base mechanism of $(\epsilon,\delta)$-DP. The $x$-axis is the ratio $\epsilon'/\epsilon$ and the $y$-axis is the difference $\delta'-\delta$. Among the examined sampling procedures,  Poisson sampling, WOR, and  MUST at certain values of $(b,m)$ can yield strong PA; MUST  at certain values of $(b,m)$ may lead to type I weak PA. We do not have a sampling procedure of type II weak PA in this work; however, it doesn't mean such procedures do not exist.  The privacy dilution category  should be avoided when designing a sampling procedure to be used together with a base randomized mechanism. 

%\begin{rem}[PA effects on $\epsilon$ and $\delta$]\label{rem:PA}
%Not all commonly used $\mathcal{S}$ yield strong PA --  Poisson and WOR do, but not WR. 
Intuitively speaking, Poisson and WOR have strong PA effects because both output a subset with either no or just one copy of a distinct element from the original dataset, implying that two subsets sampled from two neighboring datasets (whether $X\simeq_{S}X'$ or $X\simeq_{R}X'$), respectively, may be identical or differ by at most one observation. WR and the MUST procedures have Type I weak PA effects because they generate subsets that may contains multiple copies of the same individual. Therefore two subsets sampled from two neighboring datasets, respectively, may be identical or differ by $j\ge1$ observations, in which case, the group privacy $\delta_j$  in Eqn \eqref{eqn:group} will apply. While $\delta$ in the classical DP setting is often close to 0 %(e.g., $o(n^{-a})$ for $a\ge1$), 
$\delta_j$ in group privacy can be as large as 1, depending on $j$. Therefore, for WR and MUST, $\delta'$ of $\mathcal{S}\circ\M$ may be larger than $\delta$ of the base mechanism  $\M$.
%\end{rem}

\begin{rem}[aligned privacy profile]
We refer to the relation between $\epsilon'/\epsilon$ and $\delta'-\delta$ for the composited $\mathcal{S}\circ{M}$ mechanism as the \emph{aligned privacy profile}. %where $(\epsilon,\delta)$ are the privacy loss parameters associated with a base mechanism $\M$ of $(\epsilon,\delta)$-DP, and  $(\epsilon',\delta')$ are the privacy loss parameters of $\mathcal{S}\circ{M}$. 
Compared to the privacy profile  $(\epsilon',\delta')$, the aligned privacy profile is a more accurate and effective measure of the PA effect of a sampling procedure $\mathcal{S}$ on $\M$, adjusted for the baseline privacy guarantees. In addition, it effectively compresses a 4-tuple $(\epsilon,\delta,\epsilon',\delta')$ into a 2-tuple $(\epsilon'/\epsilon, \delta'-\delta)$, without throwing away information on the PA effect while allowing visualization of the PA effect in a 2-D plot.  Example plots of the adjusted privacy loss profiles for different subsampling schemes with a base Laplace and Gaussian mechanisms are provided in Sec.~\ref{sec:experiment}.
\end{rem}
% \vspace{-12pt}
%$\delta$, if small, can be viewed as the probability an individual is fully exposed (the information of the individual in the dataset is completely released or released). An individual in the whole dataset under the Poisson and WOR  scheme either appears in a subset or does not; if the individual does, there is only one copy of it. In other words, the probability of an individual being fully disclosed with the Poisson and WO sampling is reduced to $\eta\delta$, where $\eta$ is the probability of being sampled. With WR, a person can be repeated 
\begin{rem}[amplification parameter $\eta$]\label{rem:eta} $\eta\in(0,1)$ in Tables \ref{tab:amplification} and \ref{tab:existing} is the probability that an element appears in the final subsample and measures the PA effect of $\mathcal{S}$ on $\mathcal{M}$ of $(\epsilon,\delta)$-DP. $\eta$ captures fully the relation between $\epsilon$ and $\epsilon'$ for Poisson, WOR, WR, and MUST, as well as the %$\epsilon'\!=\!1+\eta(e^\epsilon\!-\!1)\!<\!\epsilon$. 
relation between $\delta$ and $\delta'$ for Poisson and WOR. %i.e., $\delta'\!=\!\eta\delta\!<\!\delta$. 
For WR and MUST that output multi-sets, the relation between $\delta'$ and $\delta$ is complicated and cannot be completely summarized by a single  parameter $\eta$. % Generally speaking, the smaller the subsampling rate, the smaller $\eta$ and $\epsilon'$ are, and the larger the PA effect is. 
\end{rem}

%We also include the PA results for the MUST method investigated in this work in\ref{thm:existing}. The mathematical relations involve an additional parameter $b$ of MUST. Since MUST also outputs multisets, its  $\delta'$ formula contains $\delta_j$ terms.

%We use subscripts to denote the subsampling method the parameters under investigation are associated with, such as $\eta$, $\epsilon'$, and $\delta'$. 

Remark \ref{rem:eta} implies that to compare the $\epsilon'$ for different subsampling scheme $\mathcal{S}$, it suffices to compare $\eta$ among them. $\eta$ can be rather complicated mathematically. We focus on MUSTow, WR, WOR, and Poisson since it is not straightforward to analytically compare  $\eta_{\text{MUSTww}}$ with others. In addition, it is analytically difficult to compare $\delta'_{\text{WOR}}, \delta'_{\text{WR}}$, $\delta'_{\text{MUSTow}}$, and $\delta'_{\text{MUSTww}}$  given their complex expression. %\footnote{\hspace{1pt}e.g.,  $\delta'_{\text{WR}}-\delta'_{\text{MUSTow}}=$ \\ $\sum_{j=1}^m \binom{m}{j}\frac{1}{n}\left(\left(\frac{1}{n}\right)^{j-1}\left(1\!-\!\frac{1}{n}\right)^{m-j}\!-\!\left(\frac{1}{b}\right)^{j-1}\left(1\!-\!\frac{1}{b}\right)^{m-j}\right)\delta_j$, where $\left(\left(\frac{1}{n}\right)^{j-1}\left(1-\frac{1}{n}\right)^{m-j}-\left(\frac{1}{b}\right)^{j-1}\left(1-\frac{1}{b}\right)^{m-j}\right)$ is positive for $j=1$ and negative for $j=2,\dots,m$. The sign of the summation of the $j$ terms can be + or - and vary case by case for different $b,m,n,$ and $\delta_j$ values.}. 
We will provide a numerical comparison of the $\eta$ and $\delta'$ in different subsampling methods in Secs \ref{sec:eg1} and \ref{sec: utility}. 

\begin{pro}[comparison of PA effect on $\epsilon$]\label{prop:epsH}
Among MUSTow$(n,b,m)$, WOR$(n,m)$, and WOR$(n,m)$, and Poisson($n,\gamma=m/n$), where $1\leq b\leq n$, $m\geq 1$, $\eta_{\text{MUSTow}}\leq\eta_{\text{WR}}\leq\eta_{\text{WOR}}=\eta_{\text{Poisson}}$ for all $b$ and $m$, where the equality only holds between $\eta_{\text{MUSTow}}$ and $\eta_{\text{WR}}$ in the trivial case of $b=n$ and between $\eta_{\text{WR}}$ and $\eta_{\text{WOR}}$ when $m=1$ .
\end{pro}
The proof of Prop.~\ref{prop:epsH} is given in the suppl. mat.. In pratice, $b<n$ (MUSTow reduces to WR when $b=n)$ and $m$ is often $>1$ and thus $\eta_{\text{MUSTow}}<\eta_{\text{MUSTwo}}=\eta_{\text{WR}}<\eta_{\text{WOR}}=\eta_{\text{Poisson}}$ and so is  $\epsilon'_{\text{MUSTow}}<\epsilon'_{\text{MUSTwo}}=\epsilon'_{\text{WR}}<\epsilon'_{\text{WOR}}=\epsilon'_{\text{Poisson}}$; in other words,  MUSTow yields the stronger PA effect on $\epsilon$ compared to the three 1-step sampling procedures. %despite that Poisson and WOR provide strong PA while WR and MUST offer weak PA. %On the other hand, the PA effect on $\epsilon$  for MUST  can be stronger compared to the other 3 sampling techniques, which is explored in more detail below. %To sum up, per Remarks \ref{rem:PA} and \ref{rem:eta} and Definition \ref{def:PA}, 

%-----------------------------------------  
\vspace{-12pt}
\subsection{\texorpdfstring{$k$}{}-fold Privacy Composition of 2-step MUST\texorpdfstring{$\circ\M$}{}}\label{subsec:composition}
MUST can be combined with DP procedures that employ repeated applications of a randomized mechanism, such as DP-SGD and subsampled bootstrapping. In the former case, the Gaussian mechanism is often applied to gradients calculated in subsamples in an iterative optimization procedure; in the latter, all subsamples are generated simultaneously and a DP mechanism can be applied in a parallel fashion to each of the subsamples to achieve DP guarantees. 

We provide an analysis of the overall privacy loss from $k$-fold composition of the MUST\hspace{1pt}$\circ$\hspace{1pt}Gaussian mechanism\footnote{The subsampled Gaussian mechanism is studied or applied more often than the subsampled Laplace mechanism in both research and practice. Interested readers may refer to \cite{wang2019subsampled, zhu2019poission, wang2022analytical} for the privacy loss analysis of the $k$-fold composition of the subsampled Laplace mechanism.} below, in a similar manner as in \cite{abadi2016deep} and \cite{koskela2020computing}, using the privacy loss accounting method for WR\hspace{1pt}$\circ$\hspace{1pt}Gaussian in \cite{koskela2020computing}\footnote{\hspace{1pt}\cite{wang2022differentially} suggest that the approach as outlined in Sec.~6.3 of \cite{koskela2020computing} is invalid by providing a counter-example. However, the counter-example does not satisfy the conditions needed for tightly $(\epsilon,\delta(\epsilon)$)-DP as the support for the privacy loss random variable $\mathcal{L}_{X/X'}(\cdot)$ is $(\log(0.25), \infty)$ and that for $\mathcal{L}_{X'/X}(\cdot)$ is $(-\infty, \log(4))$ in the counter-example.  Per Lemma 5 in \cite{koskela2020computing}, which is also noted at the end of  Sec.~\ref{sec:prelim}, the privacy profile $\delta(\epsilon)$ in their counter-example is not guaranteed to be tightly ($\epsilon,\delta$)-DP, which would require $(\epsilon,\infty)\subset$ the supports of $\mathcal{L}_{X/X'}(\cdot)$ and $\mathcal{L}_{X'/X}(\cdot)$, respectively.}.

The steps in deriving the overall privacy loss in the $k$-fold composition of MUST\hspace{1pt}$\circ$\hspace{1pt}Gaussian are as follows. First, we obtain the privacy loss random variable $\mathcal{L}_{X/X'}(t)$ from outputting $t$ in one application of MUST\hspace{1pt}$\circ$\hspace{1pt}Gaussian. Second, we derive the pdf $\omega_{X/X'}(s)$ for  $\mathcal{L}_{X/X'}(t)$ per Definition \ref{defn:PLD}. Lastly, we apply the Fourier account (FA) algorithm \citep{koskela2020computing,koskela2021tight}  to approximate the convolution over $k$ pdfs $\omega_{X/X'}(s)\oast^k\omega_{X/X'}(s)$ and obtain the composited privacy profile  $\delta(\epsilon)$.
% Definition \ref{defn:PLD} for the PLD $\omega_{X/X'}(s)$ requires the privacy loss random variable $\mathcal{L}_{X/X'}(t)$ to be monotonically increasing, while the $\mathcal{L}_{X/X'}(t)$ designed in the counter-example in \cite{wang2022differentially} is a monotonically decreasing function.
The FFA algorithm, presented in the suppl. mat., uses the 
Fast Fourier Transform (FFT) \citep{cooley1965algorithm} to obtain an approximate $\delta(\epsilon)$ and an interval where the exact $\delta(\epsilon)$ resides, taking into account all sources of approximation errors in this process. The precision of the solution, i.e., the width of the interval, depends on the convolution truncation bound  $L$ and the number of grid points $r$ used in the FFT. The FFA algorithm achieves tighter privacy bounds than the moment accountant method for Poisson$\circ$Gaussian in  \cite{koskela2021tight}.   

We present the composition result for MUSTow$\circ$Gaussian below; the steps for  MUSTww$\circ$\\Gaussian are similar and provided in the suppl. mat., along with the main results for the composition analysis of Poisson$\circ$Gaussian, WOR$\circ$Gaussian, and WR$\circ$Gaussian \citep{koskela2020computing}. 
For MUSTow\hspace{1pt}$\circ$\hspace{1pt}Gaussian under $\simeq_S$, the probability that the element that is different between $X$ and $X'$  occurs $l$ times in a subset $Y$ of size $m$ is $\frac{b}{n}\binom{m}{l}\left(\frac{1}{b}\right)^l\left(1-\frac{1}{b}\right)^{m-l}$ for $l=1,\dots,m$ and $\frac{b}{n}\left(1-\frac{1}{b}\right)^{m}+\left(1-\frac{b}{n}\right)$ for $l=0$. Following \citep{koskela2020computing},  we derive the privacy loss random variable from outputting $t$, which is
\begin{align}\label{eqn:loss.MUST}
\mathcal{L}_{X/{X'}}(t)=\log\left(\frac{f_{X}(t)}{f_{X'}(t)}\right)=\log\!\left(\!\frac{\sum_{l=0}^m\frac{b}{n}\binom{m}{l}\left(\frac{1}{b}\right)^l\left(1\!-\!\frac{1}{b}\right)^{m-l}e^{-\frac{l^2}{2\sigma^2}}e^{\frac{tl}{\sigma^2}}\!+\!1\!-\!\frac{b}{n}}{\sum_{l=0}^m\frac{b}{n}\binom{m}{l}\left(\frac{1}{b}\right)^l\left(1\!-\!\frac{1}{b}\right)^{m-l}e^{-\frac{l^2}{2\sigma^2}}e^{-\frac{tl}{\sigma^2}}\!+\!1\!-\!\frac{b}{n}}\!\right), 
\end{align}
where  $\sigma$ is the scale of the Gaussian mechanism.  Since $\frac{b}{n}\binom{m}{l}\left(\frac{1}{b}\right)^l\left(1-\frac{1}{b}\right)^{m-l}e^{-\frac{l^2}{2\sigma^2}}\!>\!0$ for $l\!=\!0,\dots,m$, the numerator in Eqn \eqref{eqn:loss.MUST} monotonically increases with  $t$ and the denominator monotonically decreases with $t$. Therefore, 
$\mathcal{L}_{X/X'}(t)\!\to\! -\infty \mbox{ as }t\!\to\! -\infty \mbox{ and }\mathcal{L}_{X/X'}(t)\!\to\! \infty \mbox{ as }t\!\to\! \infty$; that is,  the support of the pdf of the privacy loss random variable $\omega_{X/X'}$ is $\mathbb{R}$. Since  $\mathcal{L}_{X/{X'}}(t)$ is a continuously differentiable bijection function in $\mathbb{R}$, we can apply Definition \ref{defn:PLD} to obtain its pdf $\omega_{X/X'}(s)=f_X(\mathcal{L}^{-1}_{X/{X'}}(s))d\mathcal{L}^{-1}_{X/{X'}}(s)/ds$ for  $s\in\mathbb{R}$. Since the closed-form formula for $\mathcal{L}^{-1}_{X/{X'}}(s)$ does not exist for MUST  (similarly  for WR, but does exist for Poisson sampling and WOR; see \cite{koskela2020computing}), one may numerical approaches, such as the Newton's method, to find root $t$ of equation $\mathcal{L}_{X/{X'}}(t)=s$ for any given $s$ . In addition, since there is no closed-form formula for $d\mathcal{L}^{-1}_{X/{X'}}(s)/ds$, one may numerically approximate the derivatives over grid points using the difference quotient. As the final step, we can apply the FFA algorithm to numerically approximate the convolution over $k$ PLDs and calculate the $\delta(\epsilon)$ for a given $\epsilon$ per Theorem \ref{thm:delta_kfold}. Given Lemma \ref{lem:omega} and the property $f_{X'}(-t)\!=\!f_{X}(t)$, we have $\delta_{X'/X}(\epsilon)\!=\!\delta_{X/X'}(\epsilon)\!=\!\delta(\epsilon)$. 

\subsection{MUST with more than 2 stages}\label{sec:MUST3}
As suggested by Figure \ref{fig:MUST}, a MUST procedure may encompass more than two stages, the PA effect of which would be complicated and tedious to derive. From the privacy, utility, computational perspectives in practical applications,  2-stage MUST would be sufficient. First, the weak PA effect exhibited by MUST will get weaker as the number of stages increases due to the decreased number of unique data points given a pre-specified final subset size. Second, the utility of the output of $\mathcal{S}\circ\mathcal{M}$ may also be negatively affected by the reduced unique data points in the final subset as the empirical distribution of the subset becomes more discrete with reduced representative of the underlying population if it is a continuous distribution.  Some analysis procedures can be adversely affected by ties, so are the estimates, predictions, or statistical inferences. Lastly, adding additional layers of sampling would increase computational overhead. All taken together, it may not be practically appealing to go beyond 2 stages in MUST though conceptually, the general MUST procedure could be an interesting problem to study.\footnote{This is somewhat similar to Bayesian hierarchical modelling, where, in theory, $>2$ layers can be employed, including priors for parameters, hyperpriors for those priors, hyper-hyperpriors, and so forth but in practice, a Bayesian hierarchical model typically employs two layers.}  

\section{Experiments}\label{sec:experiment}
We run three types of experiments and there are multiple experiments within each type. The main observations in the experiments are summarized below. \vspace{-3pt}
\begin{itemize}
    \item \emph{privacy:} The experiments in Sec.~\ref{sec:eg1} provide numerical evidence to support the theoretical results on PA in Sec.~\ref{sec:method}. The PA effect of MUST on $\epsilon$ decreases with increasing $b$, increases with $m$ for MUSTwo and decreases with $m$ for MUSTww. In general, the PA effect of MUST is relatively insensitive to $(b,m)$ in the examined settings. 
    \item \emph{utility: } We performed four experiments in Sec.~\ref{sec: utility} to examine the utility of outputs from a DP mechanism composited with different subsampling procedures. The results suggests MUST provides similar or superior results (e.g, improved accuracy of parameter estimation and outcome prediction with higher stability) compared to single-step sampling. %For DP-SGD, all the procedures converged with about the same number of iterations. 
    \item \emph{computation: } We compared the computational efficiency of MUST with single-step sampling in Sec.~\ref{sec:computation}.  MUST has the  fewest  distinct data points in a subsample, implying the least computation time based on each distinct data point and  thus saving in computational time.  The time needed to generate a subsample in MUST  is larger than WOR but significantly smaller than Poisson sampling.
\end{itemize} 
Due to space  constraints, the implementation details of  each experiment are provided in the suppl. mat..

\subsection{Privacy Amplification Comparison}\label{sec:eg1}  
We provide several examples to demonstrate the PA of MUST vs. WOR and WR  based on Tables \ref{tab:amplification} and \ref{tab:existing} in the case of Laplace and Gaussian base mechanism with  $\Delta_1/\sigma\!=\!1$ and  $\Delta_2/\sigma\!=\!0.25$, respectively.  We set the original dataset size at $n=1000$, the subset size at $m=400$ for WR, WOR, and MUST, and $b=500$ for MUST. For Poisson sampling, its PA effect at $\gamma=m/n$ is the same as that for WOR per Table \ref{tab:existing} and is thus not examined.  

\begin{figure}[!htb]
\centering
(a) subsampled Laplace mechanism\\
\includegraphics[width=0.45\textwidth] {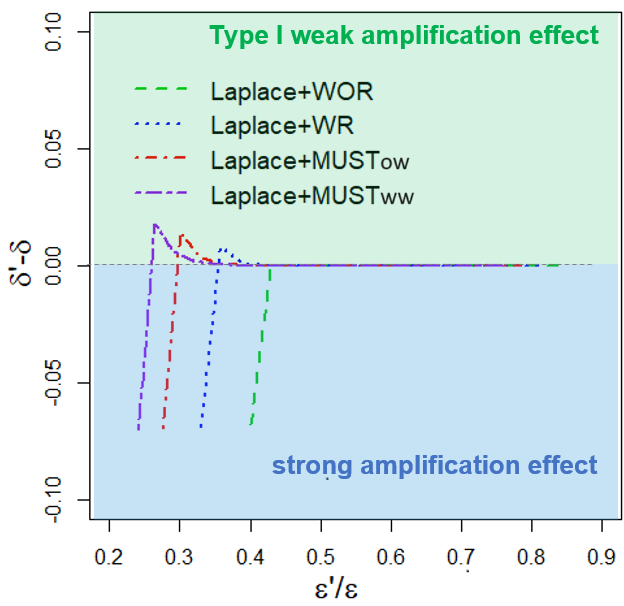}\hspace{6pt}
\includegraphics[width=0.45\textwidth]{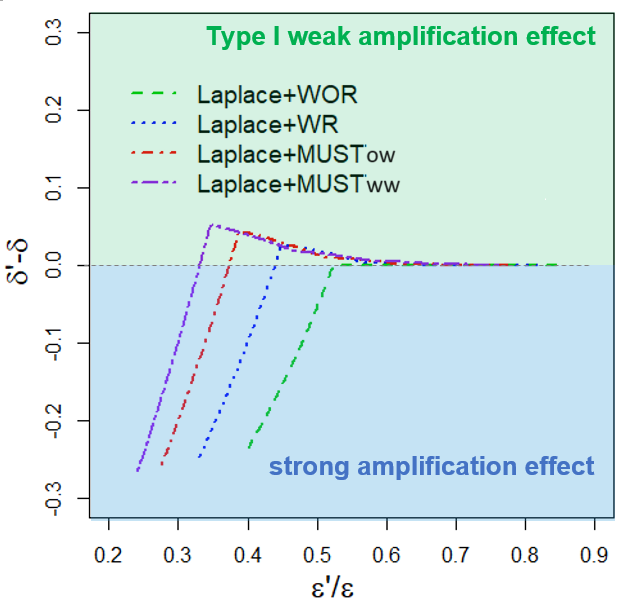}\\
 \hspace{0.1in}$\Delta_1/\sigma\!=\!0.25$ \hspace{2.1in}$\Delta_1/\sigma\!=\!1$ \\
(b) subsampled Gaussian mechanism\\
\includegraphics[width=0.45\textwidth] {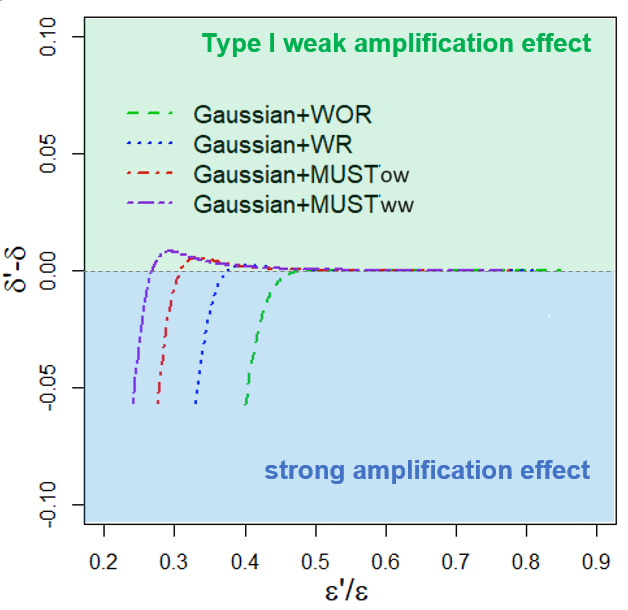}\hspace{6pt} 
\includegraphics[width=0.45\textwidth]{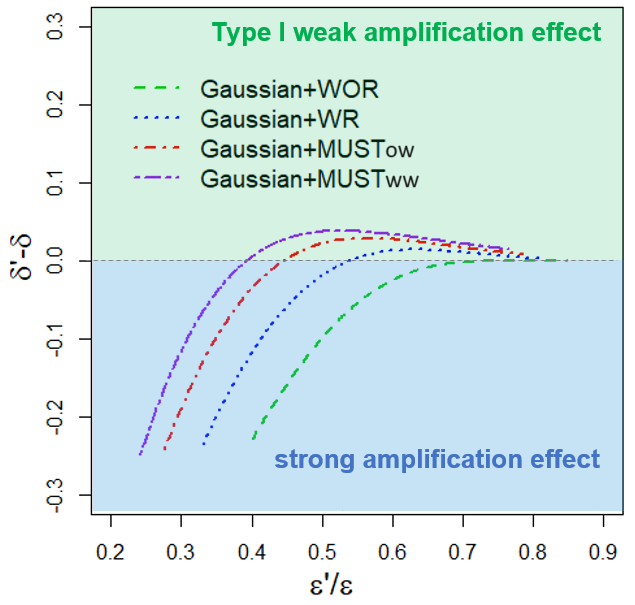}\\
\hspace{0.1in}$\Delta_2/\sigma\!=\!0.25$ \hspace{2.1in} $\Delta_2/\sigma\!=\!1$\vspace{-3pt}
\caption{Aligned privacy profile plots of $\epsilon'/\epsilon$ vs $(\delta'-\delta)$ for the Laplace and Gaussian mechanisms with subsampling schemes WOR (Poisson), WR, MUSTow and MUSTww  when $n=1000, m=400$, and $b=500$.}\label{fig: pairs}
\end{figure}

Fig.~\ref{fig: pairs} depicts the aligned privacy profiles for each subsampling scheme composited with the base Laplace and Gaussian mechanisms.  First, WOR (and Poisson) is the only subsampling procedure that leads to a 100\% strong PA effect in the examined setting whereas the rest (the 2 MUST procedures and WR) visit the Type I weak PA effect quadrant for a minor portion of each curve.   MUSTww stays in the weak-effect quadrant the longest, followed by MUSTow, then WR/MUSTwo, and finally WOR; but the difference in how long the curves stay in the weak-effect quadrant becomes less pronounced as $\Delta/\sigma$ gets smaller. Second, all mechanisms start in the strong-effect quadrant and eventually ``converge'' to the line of $\delta'-\delta=0$. The smaller $\Delta/\sigma$ is, the quicker the convergence is. Third, the base Gaussian mechanism leads to ``smoother'' peaks in the weak-effect quadrant than the Laplace mechanism. 
\begin{comment}
\begin{rem}\label{rem:MUST>WO}
We conjecture that the PA effect of MUSTwo is the same as WR. As the second WOR stage in MUSTwo is only used to shrink the stage-I subset of size $b$ to size $m$ ($<b$) in a random manner and does not change the number of an element appears in the final subset after the first WR stage.    
\end{rem}    
\end{comment}

%Fig.~\ref{fig: PA} seems to suggest that for relatively large $\delta$, we have $\delta'_{\text{MUSTww}}<\delta'_{\text{MUSTow}}<\delta'_{\text{MUSTwo}}=\delta'_{\text{WR}}<\delta'_{\text{WOR}}$ while $\delta'_{\text{MUSTww}}>\delta'_{\text{MUSTow}}>\delta'_{\text{MUSTwo}}=\delta'_{\text{WR}}>\delta'_{\text{WOR}}$  for small $\delta$. Since $\delta$ is set on the order of $o(1/n)$, the large $\delta'$ values raise questions about the applicability of subsampling schemes that involve sampling with replacement  (WR and MUST) in real applications. However, the corresponding $\epsilon'$ values are not the same for the five subsampling schemes for the same $\delta$ value. As a matter of fact, there is a negative relation between the PA effects of WR and MUST on $\epsilon$ and $\delta$ of the base mechanism.  To better examine the trade-off between $\epsilon'$ and $\delta'$ in the privacy profile for these five subsampling methods, 
Table S1 in the suppl. mat. lists some randomly selected pairs of $(\epsilon', \delta')$ from the privacy profiles of Laplace and Gaussian mechanisms, respectively. In all cases, $\epsilon'_{\text{MUSTww}}<\epsilon'_{\text{MUSTow}}<\epsilon'_{\text{WR}}<\epsilon'_{\text{WOR}}$, while MUST has a larger $\delta'$ than WOR and WR in general 
 except for small $\epsilon$ and large $\delta$ when $\Delta/\sigma\!=\!1$. WOR always has a strong PA effect while WR and MUST have a strong PA effect only at small $\epsilon$ and shif to Type I weak PA effect as $\epsilon$  gets larger.

\begin{figure}[!htb]
\centering
\footnotesize{\textcolor{teal}{\hspace{0.75in}$\eta$ vs $(b,m)$  \hspace{1in}  $\delta'-\delta$ vs $(b,m)$\hspace{1in}  $\delta'-\delta$ vs $(b,m)$ }}\\ \vspace{-2pt}
\footnotesize{\textcolor{teal}{\hspace{0.75in}MUST$\circ\M$ \hspace{1in} MUST$\circ$Laplace\hspace{1in} MUST$\circ$Gaussian }}\\
\vspace{-2pt}
\raisebox{0.5in}{\textcolor{teal}{\footnotesize{\rotatebox{90}{MUSTwo}}}}{
\includegraphics[width=0.3\textwidth,trim= 0pt 0pt 12pt 0pt, clip]{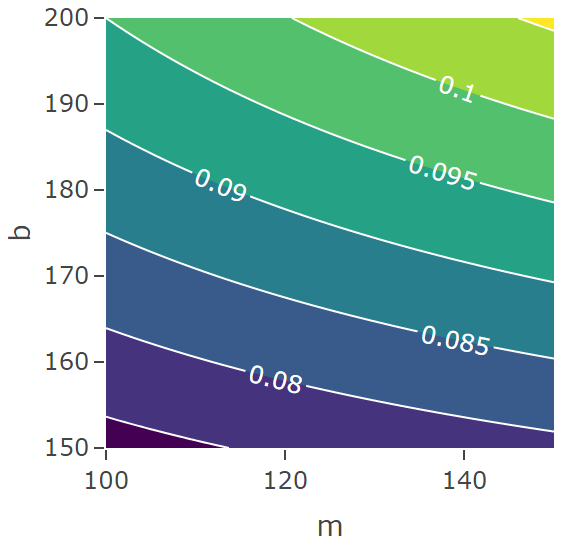}
\includegraphics[width=0.3\textwidth,trim= 0pt 0pt 12pt 0pt, clip]{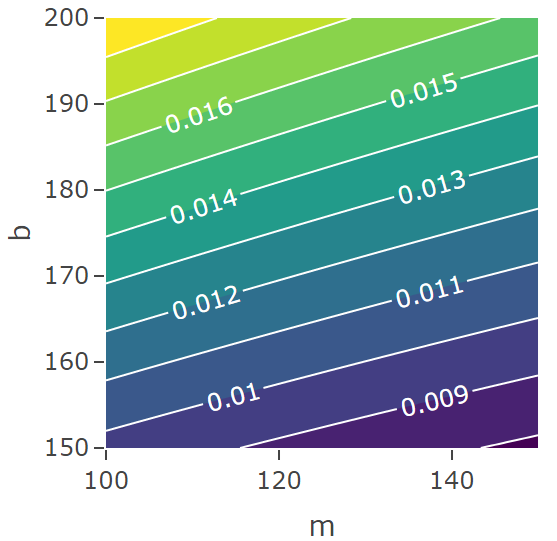}
\includegraphics[width=0.3\textwidth,trim= 0pt 0pt 12pt 0pt, clip]{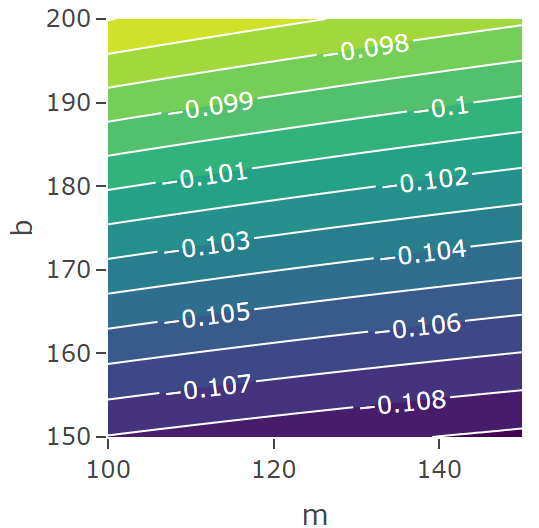}}\\
\raisebox{0.5in}{\textcolor{teal}{\footnotesize{\rotatebox{90}{MUSTww}}}}
\includegraphics[width=0.3\textwidth,trim= 0pt 0pt 12pt 0pt, clip]{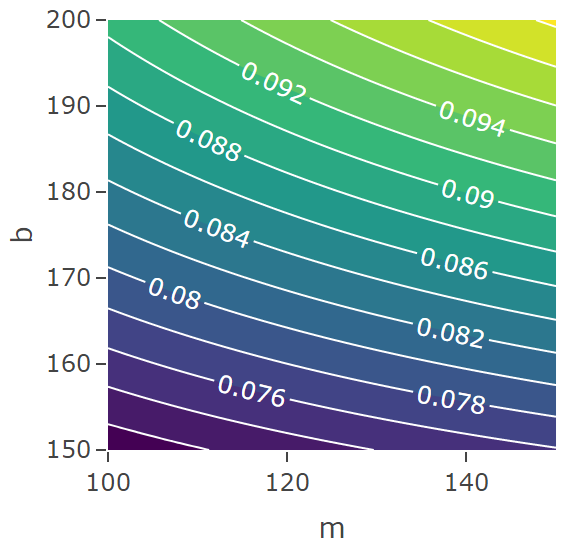}
\includegraphics[width=0.3\textwidth,trim= 0pt 0pt 12pt 0pt, clip]{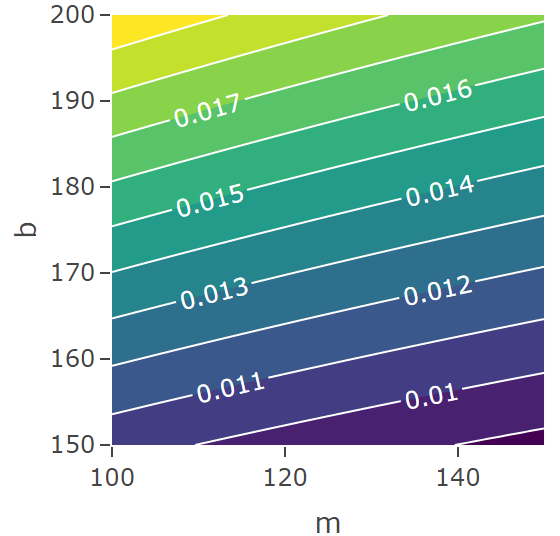}
\includegraphics[width=0.3\textwidth,trim= 0pt 0pt 12pt 0pt, clip]{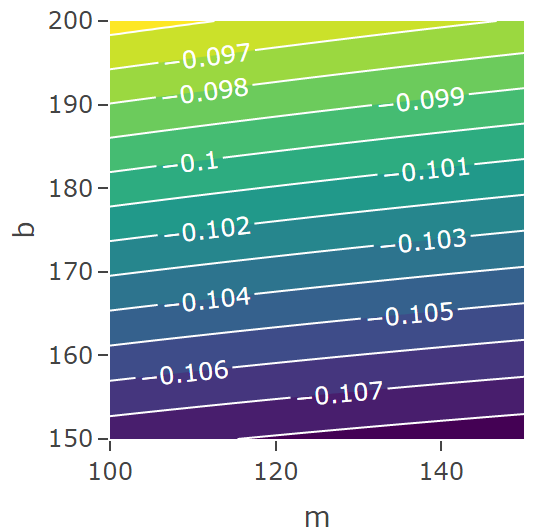}\\

\caption{Contour plots of $\eta$ vs $(b, m)$ for MUST$\circ\M$ ($\M$ is a generic base mechanism)  and $(\delta'-\delta)$ vs $(b, m)$ for  MUST $\circ$ Laplace mechanism with $\Delta_1/\sigma\!=\!1$ and  MUST $\circ$ Gaussian mechanism with $\Delta_2/\sigma\!=\!1$ ($n=1000$, $\epsilon=1$ for the base mechanism)}\label{fig: contour}
\end{figure}

\emph{Sensitivity of PA effects to MUST hyperparameters.} The 1-stage sampling procedures has one hyperparameter that affects the size of the subsample and as well as their PA effects through $\eta$, whereas MUST has two hyperparameters ($b,m$). We set $n=1000$, $\epsilon=1$ for a generic $\M$, $\Delta_1/\sigma=1$ for the Laplace mechanism, and $\Delta_2/\sigma=1$ for the Gaussian mechanism; and examine the relations between  ($b,m$) vs $\eta$ that is approximated by $\epsilon'/\epsilon$ and  $(\delta'-\delta)$ for $m\in[100, 150]$ and $b\in[150,200]$. The contour plots are provided in Fig.~\ref{fig: contour} (the 3D versions of the plots are provided in the suppl. mat.). %As depicted in Fig \ref{fig:PAeffect}, the smaller the $\eta$ and $\delta'-\delta$ are, the larger the PA effects.  present the contour plots in Fig.~\ref{fig: contour} of $\eta$ vs $(b, m)$ for MUST$\circ\M$, where $\M$ is a generic base mechanism, and $(\delta'-\delta)$ vs $(b, m)$ for  MUST$\circ\M$, where $\M$ is the Laplace mechanism and  Gaussian mechanism, respectively, based on the analytical results in Theorem \ref{thm:existing}. 
The observations can be summarized as follows. First, MUSTww yields smaller $\eta$ and thus a stronger PA effect on $\epsilon$ than MUSTow;  MUSTwo yields smaller $\delta'-\delta$ and thus a stronger PA effect on $\delta$ than MUSTww; however, the differences in $\eta$ and $\delta'-\delta$ between MUSTow and MUSTww are minor. Second,  $\eta$ increases with $b$ and decreases with $m$, and $\delta'-\delta$ increases with both $m$ and $b$, implying that the PA effects for MUSTow and MUSTww weaken as $b$ increases, strengthen on $\epsilon$ and weaken on $\delta$ as $m$ increases. Overall speaking, while the PA effects of MUST change with $b$ and $m$ but the rate of the changes are rather slow.

\subsection{Utility Analysis}\label{sec: utility}
We compare the utility of PP outputs from several DP procedures with subsampling in four experiments: subsampling bootstrap for parameter estimation and sensitivity analysis of MUST to hyperparameters (Sec.~\ref{sec: utility bootstrap}), parameter estimation and outcome prediction via linear regression (in Sec.~\ref{sec: utility lm}), and two classification experiments (Sec.~\ref{sec: utility logistic}). %The first two experiments are run on simulated data and the latter two are performed on real data.  
In all experiments, the non-private procedure with Poisson subsampling is used as the baseline; the DP procedures employ  Gaussian mechanisms composited with subsampling procedures WR, WOR, Poisson, MUSTow, and MUSTww\footnote{\hspace{1pt}We also run MUSTwo in each experiment and its utility results resembled those from WR, as they should.}. 
% 1) Specify delta per query before PA  (should be based on n or m? less important, user-specified)
% 2) Specify eps' per query after PA --> back-solve for epsilon before PA
% 3) calculate Delta_2^2 (should be based on n or m?)
% 1), 2), 3) fed to Eqn (9) to calculate sigma^2 for the base mechanism 

% to calculate delta_c after composition, we feed in 
% 1) epsilon_c, which is user-specified. 
% 2) PLD omega based on Def 3 and Eqn 8 which uses sigma^2 calculated above (this also suggests that different (delta, eps', and Delta_2^2 combinations can lead to the same sigma^2, which is what matters). 
% Ideally, it should be calculate sigma^2 given (delta_c^2, and epsilon_c^2), instead of calculating delta_c^2 given sigma^2 and epsilon_c^2.
% In addition, given that both delta' and delta_c are user-specified, sigma_c could > or < sigma'
% for delta_c, there should be a positive relation with delta. the larger delta, the smaller sigma^2, the larger delta should be. However, whether delta_c > or < delta depends on the PA effect vs the convolution. For these involving WR, I would say delta_C > delta for sure. 

%---------------------------------------------------
\vspace{-6pt}\subsubsection{Utility experiment  1: statistical inference} %Subsampling Bootstrap for Parameter Estimation and Sensitivity of MUST to Its Hyperparameters}
\label{sec: utility bootstrap}
We simulate 200 datasets, each with $n\!=\!300$ samples from $\mathcal{N}(\mu\!=\!0, \tau^2\!=\!1)$ and estimate $\mu$ and $\tau^2$ using subsampling bootstrap\footnote{\hspace{1pt}  Subsampling bootstrap is a class of procedures that incorporate subsampling and bootstrap to obtain valid statistical inference with computational efficiency, especially in large-scale data. Common procedures include $m$ out of $n$ bootstrap \citep{bickel1997resampling}, subsampling \citep{politis1999subsampling}, double bootstrap \citep{lee1999class}, and the BLB procedure \citep{kleiner2014scalable}; the former two are one-stage subsampling procedures (WR and WOR, respectively) whereas the latter two are two-stage (corresponding to MUSTww and a special case of Fig.~\ref{fig:MUST2}(a) with a small $s$  and a large $r$.}  in each repetition. We generate $T\!=\!500$ bootstrap samples, each sized at $m\!=\!30$ for each subsampling method.  For the two MUST procedures, we use different $b$ values to examine its impact on PP inference.

We sanitized the sample mean and variance in each bootstrap sample via the Gaussian mechanism and released their averages over $T=500$ bootstrap samples as the final PP estimates for the population mean and variance. We set $\epsilon'=0.1$ and $\delta=1/n$ per query and calculated $\sigma$ for the Gaussian mechanism (Theorem \ref{thm:LapGau} \cite{dwork2014algorithmic}).  %For estimating both sample mean and sample variance, WOR and Poisson sampling require the same largest $\sigma$, followed by WR and MUSTwo, MUSTow, and MUSTww. MUSTwo requires the same noise scale as WR regardless of $b$. For MUSTow and MUSTww, a larger $b$ requires a larger $\sigma$, while their largest $\sigma$s over all the $b$ values under investigation are still smaller than any other sampling methods.

The values of $\sigma$ of the base Gaussian mechanisms are given in Table \ref{tab:bootstrap noise}.  A smaller $\sigma$ value indicates a larger PA effect of a sampling method since less noise is required to achieve the same privacy level after subsampling. Prop.~\ref{prop:epsH} suggests that MUSTow has the strongest PA effect on $\epsilon$ compared to Poisson, WOR,  and WR and thus has the smallest $\sigma$. %The scale for Poisson and WOR can be 2-fold higher than  MUSTow, depending on $b$.  The $\eta$ values of MUSTww is hard to compare to others analytically due to the lack of a convenient closed form. 
MUSTww has a smaller $\sigma$ than MUSTow in some cases, consistent with the observations in Fig.~\ref{fig: contour}, but overall they are similar.  As $b$ increases, $\sigma$ for MUSTww and MUSTow increases, as expected.
\begin{table}[!htb]
\centering
\caption{Scale $\sigma$ of the base Gaussian mechanisms and per-query $\delta'$ in utility experiment 1 ($n\!=\!300, m\!=\!30, T\!=\!500$ bootstrap subsamples, $\epsilon'\!=\!0.1, \delta\!=\!1/300$ per statistic and per subsample)} \label{tab:bootstrap noise}
\resizebox{0.65\columnwidth}{!}{
\begin{tabular}{@{} l|c@{\hspace{6pt}}c@{\hspace{6pt}}c@{\hspace{6pt}}c@{\hspace{6pt}}c@{\hspace{6pt}}|
c@{\hspace{6pt}}c@{\hspace{6pt}}c@{\hspace{6pt}}c@{\hspace{6pt}}c@{\hspace{6pt}}}
\toprule
$\mathcal{S}$ & \multicolumn{5}{c|}{$\sigma$ for PP estimation of $\mu$} & \multicolumn{5}{c}{$\sigma$ for PP estimation of $\tau^2$} \\
\hline
Poisson & \multicolumn{5}{c|}{0.13} & \multicolumn{5}{c}{1.02}\\
WOR & \multicolumn{5}{c|}{0.13} & \multicolumn{5}{c}{1.02}\\
WR & \multicolumn{5}{c|}{0.12} & \multicolumn{5}{c}{0.99}\\
\hline
$\quad  b$ & $10$ & $20$ & $30$ & $50$ & $100$ &
$10$ & $20$ & $30$ & $50$ & $100$\\
\hline
MUSTow & 0.06&	0.08&	0.09&	0.11 & 0.12 &0.50	&0.67&	0.75&	0.84&	0.93\\
MUSTww & 0.06&	0.08&	0.09	&0.10	&0.11&	0.50&	0.66&	0.74&	0.82&	0.90\\
\midrule
\end{tabular}}
\resizebox{0.65\columnwidth}{!}{
\begin{tabular}{@{}c@{\hspace{7pt}}|c@{\hspace{7pt}}|c@{\hspace{7pt}}|c@{\hspace{7pt}}|c@{\hspace{7pt}}c@{\hspace{7pt}}c@{\hspace{7pt}}c@{\hspace{7pt}}c@{\hspace{7pt}}@{\hspace{7pt}}|c@{\hspace{7pt}}c@{\hspace{7pt}}c@{\hspace{7pt}}c@{\hspace{7pt}}c@{}}
\multicolumn{14}{c}{$\delta' (\times10^{-4}$) at $\epsilon'=0.1$}\\
\hline Poisson & WOR & WR & & \multicolumn{5}{c|}{MUSTow} & \multicolumn{5}{c}{MUSTww }\\
\cline{1-3}\cline{5-14}
& & & $b=$ & 10 & 20 & 30 & 50 & 100 & 10 & 20 & 30 & 50 & 100\\
\cline{5-14}
3.0 & 3.0 & 5.7 & & 64 & 22 & 11 & 5.0 & 2.0 & 65 & 24 & 13 & 7.0& 3.0\\
\bottomrule
\end{tabular}}

%\resizebox{0.9\textwidth}{!}{\begin{tabular}{l}
%\footnotesize{NA: not applicable as MUSTwo  requires $b\geq m$.$\hspace{2in}$}\\
%\hline
%\end{tabular}}\vspace{-3pt}
\end{table}

The PP estimates of $\mu$ and $\tau^2$ are presented in  Tab.~\ref{tab:bootstrap}.  The main observations are summarized as follows.
First, the PP mean estimates and variance are close to their respective true values of 0 and 1 for all sampling methods with close-to-0 bias except for MUST in the case of variance estimation, where there is some noticeable downward bias when $b\le m$.  
Second, MUSTow and MUSTww offer slightly smaller SD and thus higher stability around the PP mean and variance estimates, especially the latter, than WR, WOR, and Poisson.  The stability decreases with $b$ for MUST, consistent with the increase in the scale parameter of the base Gaussian mechanism in Tab.~\ref{tab:bootstrap noise}. Taken together with the accuracy of the estimates, this suggests that practical applications of MUST should avoid using either very small  $b$ as inferential bias may be large or very large $b$ as the stability of the estimates may decrease or the PA effect weakens (Fig.~\ref{fig: contour}).
\begin{table}[!htb]
\centering
\caption{Mean (SD) of PP estimates of $(\mu, \tau^2)$ in 200 simulated datasets ($n\!=\!300, m\!=\!30,  T\!=\!500$ bootstrap samples) in utility experiment 1} \label{tab:bootstrap} \vspace{-3pt} %, $\epsilon'=0.1, \delta=1/300$ per bootstrap sample
\resizebox{0.8\textwidth}{!}{
\begin{tabular}{@{} l @{\hspace{3pt}} c @{\hspace{3pt}} c @{\hspace{3pt}} c @{\hspace{3pt}} c @{\hspace{3pt}} c @{}}
\toprule  
$\mathcal{S}$ & &\multicolumn{2}{c}{estimation of $\mu$ (0) } &\multicolumn{2}{c}{estimation of $\tau^2$ (1)} \\
\hline
Poisson &  (non-private) &\multicolumn{2}{c}{$0.0020 \;(0.059)$} & \multicolumn{2}{c}{$0.990 \;(0.084)$}\\
Poisson & & \multicolumn{2}{c}{$0.0024 \;(0.058)$} & \multicolumn{2}{c}{$0.989  \;(0.093)$}\\
WOR && \multicolumn{2}{c}{$0.0025 \;(0.058)$} & \multicolumn{2}{c}{$0.991 \;(0.092)$}\\
WR && \multicolumn{2}{c}{$0.0026 \;(0.058)$} & \multicolumn{2}{c}{$0.984  \;(0.094)$}\\
\cline{2-6}
MUST& $b=10$ & $b=20$ & $b=30$ & $b=50$ & $b=100$ \\
\cline{2-6}
&\multicolumn{5}{c}{estimation of $\mu$}\\
\cline{2-6}
%MUSTwo & NA  & NA  & $0.0006 \;(0.058)$ &	$0.0024 \;(0.059)$ &	$0.0023  \;(0.058)$ &	$0.0013 \;(0.059)$\\
ow & $ 0.0027 \;(0.058)$ &	$0.0007 \;(0.061)$ & $0.0013 \;(0.058)$ &	$0.0025 \;(0.057)$ & $0.0022\;(0.060)$\\
ww & $0.0003 \;(0.060)$ &	$0.0013 \;(0.059)$ &	$0.0009 \;(0.059)$ &	$0.0010	 \;(0.059)$ &	$0.0025 \;(0.058)$\\
\cline{2-6}
&\multicolumn{5}{c}{estimation of $\tau^2$}\\
\cline{2-6}
%MUST& $b=10$ & $b=20$ & $b=30$ & $b=50$ & $b=100$ \\
%\cline{2-6}
%MUSTwo & NA  & NA  & $0.987 \;(0.093)$ &	$0.988 \;(0.094)$ & $0.983 \;(0.091)$ &	$0.985 \;(0.099)$\\  
ow & $0.886 \;(0.082)$ &	$0.943 \;(0.086)$ &	$0.958 \;(0.090)$ &	$0.969	 \;(0.093)$ &	$0.984 \;(0.096)$\\
ww & $0.889 \;(0.081)$ &	$0.935 \;(0.086)$ &	$0.956 \;(0.090)$ &	$0.965 \;(0.088)$ &	$0.972 \;(0.096)$\\
\bottomrule
\end{tabular}}
%\resizebox{0.5\textwidth}{!}{\begin{tabular}{l}
%\small{NA: not applicable as MUSTwo  requires $b\geq m$.$\qquad\hspace{1.2in}\qquad$}\\
\vspace{-12pt}
\end{table}

\subsubsection{Utility experiment 2: linear regression}\label{sec: utility lm}
We simulated 200 datasets of size $n\!=\!1000$ from a linear model $y\!=\!\beta_0\!+\!\beta_1x_1\!+\!\beta_2x_2\!+\!\mathcal{N}(0,1)$ with $\beta_0\!=\!1,\beta_1\!=\!0.5$, and $\beta_2\!=\!0.2$ and trained a linear regression model on each simulated dataset using DP-SGD \cite{abadi2016deep} (Alg.~S1 in the suppl. mat.) to obtain PP estimates on $\boldsymbol{\beta}=(\beta_0, \beta_1, \beta_2)^{\top}$ and outcome prediction with a mean squared loss between the observed and predicted $y$. We set $b=200$, $m=100$, the clipping constant $C$ in the DP-SGD procedure at 3, and a learning rate of $0.04$ .

The training loss profiles in a single repeat during DP-SGD at the per-iteration privacy loss $\epsilon'=0.01$ and $0.001$ for the $\mathcal{S}\;\circ\;$Gaussian are presented in Fig.~S4 in the suppl. mat.. The training loss from all the DP-SGD procedures converges to similar levels as the non-private SGD and there is not much difference among different $\mathcal{S}$. %the training loss fluctuated at smaller $\epsilon'$ due to larger $\sigma$ of the base Gaussian mechanism (Tab.~\ref{tab:lm_rmse_main}). 
$\sigma$ of the base Gaussian mechanism at the prespecified $\epsilon'$ and $\delta$, the calculated per-query $\delta'$, and the mean and SD of the root mean squared error (RMSE) of predicted $\hat{y}$ over the $200$ repeats are presented in Tab.~\ref{tab:lm_rmse_main}.  MUSTww the smallest $\sigma$, attributed to its strongest PA effect, at $\epsilon'=0.01$, followed by MUSTow, WR, and WOR. The difference in $\sigma$ among the sampling procedures is more pronounced at  $\epsilon'=0.001$. The RMSE mean and SD based on DP-SGD are slightly larger than the non-private baseline. MUSTww yields the lowest mean RMSE at $\epsilon'=0.001$. The per-query $\delta'$ values for WR, MUSTow and MUSTww, all of which generate multisets, are larger than Poisson and WOR, as expected.
\begin{table}[!htb]
\centering
\caption{$\sigma$ of the base Gaussian mechanism, per-query $\delta'$, and average RMSE (SD) of  PP predicted $\hat{y}$ over 200  simulated datasets in utility experiment 2 ($n=1000, b=200,  m=100, T=200$)} \label{tab:lm_rmse_main}\vspace{-5pt}
\resizebox{0.7\textwidth}{!}{
\begin{tabular}{@{}l@{\hspace{4pt}}c@{\hspace{4pt}}@{\hspace{4pt}}c@{\hspace{4pt}}c@{\hspace{4pt}}c@{\hspace{4pt}} c@{\hspace{4pt}}c@{\hspace{4pt}}c@{}}
\toprule
&\multicolumn{3}{c}{$\epsilon'=0.01$}
&&\multicolumn{3}{c}{$\epsilon'=0.001$}\\
\cline{2-4}\cline{6-8}
$\mathcal{S}$ & $\sigma$ &$\delta'$ &RMSE && $\sigma$ & $\delta'$ &RMSE  \\
&&&  mean (SD) &&&& mean (SD)\\
\hline 
Poisson & 0.118& 0.0001&1.000 (0.022)&& 1.138 & 0.0001&	1.036 (0.037)\\
WOR & 0.118& 0.0001&1.001 (0.023)&& 1.138 &	0.0001&1.030 (0.033)\\
WR & 0.113&	0.095&0.997 (0.022)&& 1.084	&0.080  &1.032 (0.029)\\
%MUSTwo & 0.113& 0.998 (0.022)&& 1.084	&1.031 (0.032)\\
MUSTow & 0.094&0.079	&0.997 (0.024)&& 0.898&	0.073&1.031 (0.032)\\
MUSTww & 0.091&0.076	&0.999 (0.023)&& 0.865&	0.071 &1.017 (0.025)\\
\hline
\multicolumn{8}{l}{non-private Poisson: mean RMSE (SD) was 0.998 (0.024)}\\
\bottomrule
\end{tabular}}
\end{table}

The bias and RMSE of the PP  estimates  $\hat{\boldsymbol{\beta}}^*=(\hat{\beta}^*_0,\hat{\beta}^*_1,\hat{\beta}^*_2)$ over the $200$ repeats are provided in Tab.~\ref{tab:lm_beta_main}. The biases in the DP procedures and the non-private baseline are all close to 0. The RMSE values for DP- procedures are larger than the non-private baseline, especially at $\epsilon'=0.001$; but those with MUSTww and MUSTow are noticeably smaller than the rest sampling procedures.

\subsubsection{Utility experiment 3:  classification}\label{sec: utility logistic}
We run two classification experiments  with DP guarantees: 1) logistic regression on the adult dataset \citep{miscadult2} to predict whether the annual income is over $50K$; 2) training a neural network (NN) on the MNIST dataset \citep{726791} to predict digits 0 to 9. In both experiments, we employ the DP-SGD algorithm with the cross-entropy loss.  In the adult data,  we use the $41,292$ complete records whose native country is the U.S. with $11$ predictors (the details of the model are provided in the suppl. mat.).  The dataset is split into a training set of size $n=30,969$ and a test set of size $10,323$. We set $m=100, b=200$, the gradient clipping constant $C=1.5$,  learning rate $0.4$,  $\epsilon'=5\times10^{-5}$ in each iteration of the DP-SGD algorithm.  
For the MNIST experiment, the detail of the NN architecture is provided in the suppl. mat.. The training set consists of $n=60,000$ images and the test set has 10,000 images. We set $b=3000, m=2000, C=2$, learning rate $0.3$  and $\epsilon'=10^{-5}$ in the DP-SGD algorithm. In both experiments, we calculate $\sigma$ of the base Gaussian mechanism via the formula in Theorem \ref{thm:LapGau} \citep{balle2018improving} at $\delta=1/n$ and examined the prediction performance on the test data after training for $15$ iterations in the adult experiment and  for 100 iterations in the MNIST experiment.

\begin{figure}[!htb]
\centering
\includegraphics[width=0.6\linewidth, trim= 0.5in 0.1in 0.8in 0.6in, clip]{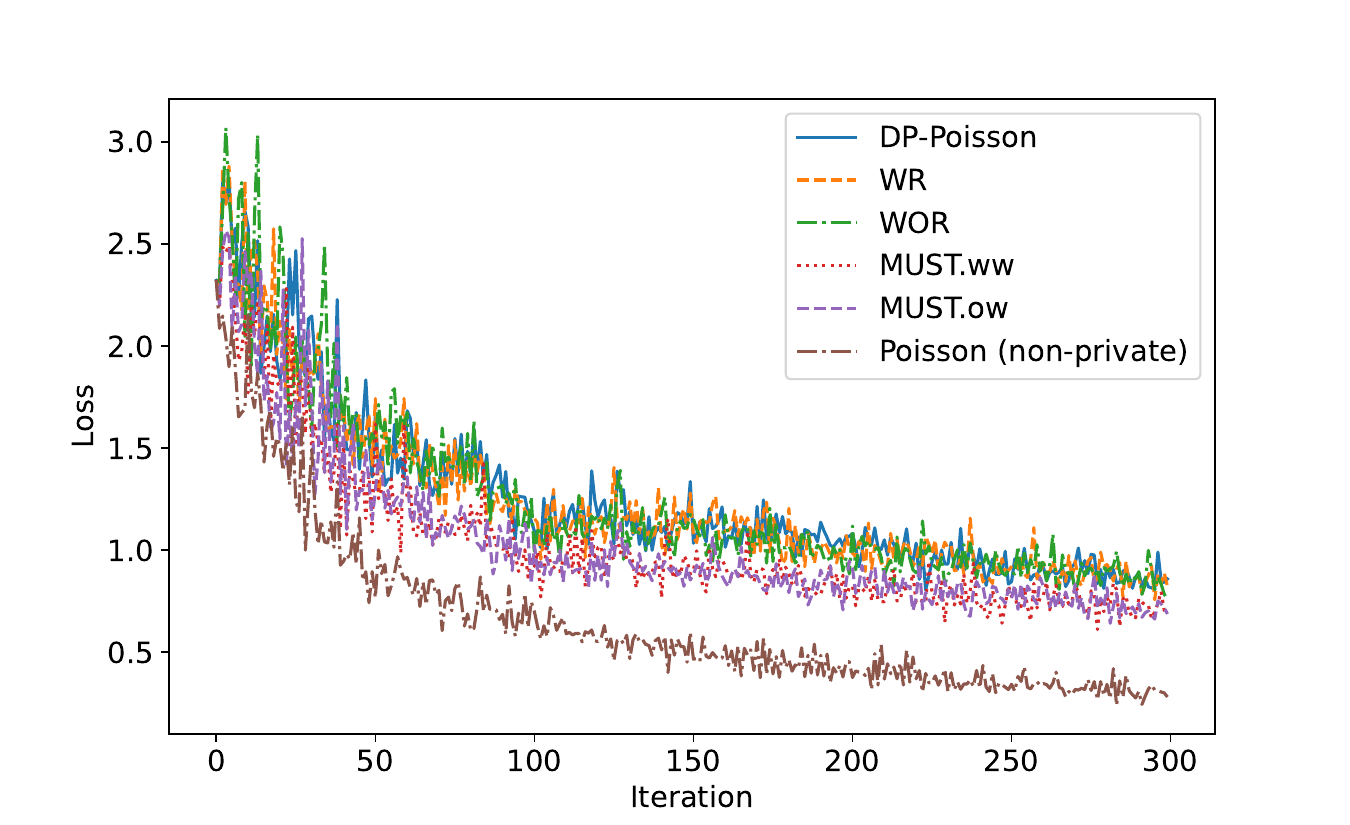}\vspace{-3pt}
\caption{Training loss in MNIST experiment}\label{fig:loss}
\end{figure}

Fig.~\ref{fig:loss} present the he training loss in MNIST data. Since MUSTow and MUSTww require smaller $\sigma$ to achieve the same $\epsilon'$ per iteration compared to WR, WOR, and Poisson, their gradients are less perturbed and their  loss is smaller and closer to the non-private loss. The training loss in the Adult data is in the suppl. mat. due to space constraint, with a similar but less pronounced trend as Fig.~\ref{fig:loss}. %$\sigma m$ in the Adult is $\sim$2.68 for the single-step sampling, and $\sim$2.21 for the two MUST programs 

The prediction accuracy on the test set in each experiment is presented in Tab.~\ref{tab:experiment3}. For the adult data, while the performance of DP-SGD based on the 5 subsampling schemes is similar to the non-private baseline, MUSTow is the best; for the MNIST data,  MUSTow and MUSTww achieve higher accuracy than WR, WOR, and Poisson.
\begin{table}[!htb]
\centering\vspace{-3pt}
\caption{Test set prediction accuracy in the adult data and MNIST data based on the classified trained via DP-SGD } \label{tab:experiment3} \vspace{-3pt}
\setlength{\tabcolsep}{2pt}
\resizebox{0.7\columnwidth}{!}{
\begin{tabular}{l|c|ccc|cc}
\toprule  
 & Poisson (non-private) & Poisson & WOR & WR & MUSTow& MUSTww \\
\midrule
Adult & 75.19 & 74.83 & 74.93 & 74.92 & 75.33 & 74.90\\
MNIST & 88.28 & 76.14 & 78.60 & 78.92 & 80.09 & 81.43\\
\midrule
\multicolumn{7}{l}{15 and 100 iterations for the adult and MINIST experiments, respectively}\\
\bottomrule
\end{tabular}}
\vspace{-15pt}
\end{table}

%-----------------------------------------------------
\vspace{-3pt}\subsection{Computational Efficiency}\label{sec:computation}
For subsampling procedures that involve WR (WR and MUST), the number of distinct samples $m_0$ in the final subset is smaller than that in WOR and Poisson which do not use WR. Tab.~\ref{tab:unique} provides the $m_0$ values in the final subsample generated by each subsampling method and  the time needed to generate a subset of the same size via a $\mathcal{S}$, relative to WR based on 10,000 Monte Carlo evaluations in the utility experiments. The main observations are as follows.  1) MUSTww yields subsets with the lowest $m_0$, followed closely by MUSTow. 2) Poisson sampling on average has the same number of unique samples as WOR, but the actual $m_0$ can vary greatly across different subsets. If it is combined with an algorithm with a superlinear time complexity in the number of samples, Poisson sampling would be associated with a higher computational cost than WOR. 3) WR and MUSTwo are (and should be) the same in terms of $m_0$, which is in between Poisson/WOR and MUSTow/MUSTww. 4)  WOR in general is the fastest, followed by WR which is either similar to WOR (experiment 1) or notably longer (experiment 3). 5)  Poisson sampling in experiments 2 and 3 takes significantly longer than the other subsampling procedures. 6)  The time for MUSTow and MUSTww is either similar to WR (experiment 3) or between WR and Poisson sampling (experiments 1 and 2); and MUSTww takes less time than MUSTow in all three experiments. 
\begin{table}[!htb]
\centering
\caption{Mean unique samples in the final subset generated by different subsampling methods and \%increase in time taken to generate a subset relative to WR (based on 10,000 Monte Carlo evaluations)}\label{tab:unique}\vspace{-3pt}
\resizebox{0.7\textwidth}{!}{
\begin{tabular}{@{}c@{\hspace{3pt}}c@{}c@{\hspace{6pt}}c@{\hspace{6pt}}c@{}@{\hspace{6pt}}ccc@{}}
\toprule
exp. & $(n,b,m)$ &  & WOR  & Poisson& WR  & \multicolumn{2}{c}{MUST}\\
\cline{7-8}
 &  &  &  &    & & ow & ww\\
\hline
1 &  (300,  & min & 30 & 14 & 22 & 15 & 15\\
    &50,    & mean & 30 & 30 & 29 & 23 & 22 \\
    &30)    & max & 30 & 52 & 30 & 29 & 29\\
\cline{2-8}
 &\%time & increase & 41.67   & 233.33 & --  & 66.67 & 75.00\\
\hline
2 &  (1000, & min & 100 & 61  & 87 & 67 & 64\\ 
    &200,   & mean& 100 & 100 & 95  & 79 & 76 \\
    &100)   & max & 100 & 138 & 100  & 90 & 87\\
\cline{2-8}
&\%time & increase & 18.75 & 437.50  & -- & 106.25 & 87.50\\
\hline
3 (adult) & (30969, & min  & 300 & 241 & 292 & 204 & 205\\ 
            &500,   & mean & 300 & 300 & 299 & 226 & 225 \\
            &300)   & max  & 300 & 373 & 300 & 246 & 250\\
\cline{2-8}
&\%time & increase & 146.43  & 7825.00   & -- &  232.14  & 103.57  \\
\hline
3 (MNIST) &  (60000, & min  & 2000 & 1839 & 1945 & 1409 & 1378\\ 
&3000, & mean               & 2000 & 2000 & 1967 & 1460 & 1442\\
&2000) & max                & 2000 & 2174 & 1986 & 1513 & 1501\\
\cline{2-8}
&\%time & increase & 46.24  & 2162.92  & -- &  140.32 &  91.94  \\
\midrule
\multicolumn{8}{p{0.75\linewidth}}{\footnotesize
WR and MUST were realized using R function \texttt{sample} with option \texttt{replace} at \texttt{T} and \texttt{F}, respectively, whereas Poisson sampling was real-
ized by performing independent  Bernoulli trials with R function  \texttt{rbinom}. 
 The time complexity of each function is provided in the Suppl. Mat.}\\
\bottomrule
\end{tabular}}
%\resizebox{0.5\textwidth}{!}{\begin{tabular}{l}
%\footnotesize{WR and MUSTwo are equivalent; the small numerical differences in some }\\
%\footnotesize{values above  are due to  Monte Carlo errors.}\\
%\hline
%\end{tabular}}
\vspace{-10pt}
\end{table}

%Tab.~\ref{tab:time} displays the  %The variation also suggests that the time for MUST depends on ($n,b,m$). 
\begin{comment}
\begin{table}[!htb]
\centering
\caption{\%increase in time taken to generate a subset relative to WOR (averaged over 10,000 Monte Carlo evaluations)} \label{tab:time}\vspace{-3pt}
\resizebox{0.48\textwidth}{!}{\begin{tabular}{@{}c@{\hspace{-6pt}}ccc ccc@{}}
\hline
experiment & $(n,b,m)$  & WR  & MUSTww  & MUSTow & Poisson\\

\hline
1 &  (300, 50, 30) & 23.08 & 53.85 &  69.23 & 115.38\\
2 &  (1000, 200, 100)  & 11.11 & 72.22 &  111.11  & 350.00\\
3 &  (30969, 500, 300)  & 110.34 & 93.10 &  189.66 & 6748.28\\
\hline
\end{tabular}}
%\resizebox{0.5\textwidth}{!}{\begin{tabular}{l}
%\footnotesize{The mean \% increase in time for  MUSTwo are 61.54\%, 88.89\%, 100.00\%,} \\
%\footnotesize{in experiments 1 to 3, respectively, which are notably larger than its }\\ 
%\footnotesize{equivalent  procedure WR, except for in experiment 3.}\\
%\hline
%\end{tabular}}
\resizebox{0.5\textwidth}{!}{
\begin{tabular}{@{}l@{}}
WR and MUST were realized using R function \texttt{sample} with option \\ \texttt{replace} at \texttt{T} and \texttt{F}, respectively, whereas Poisson sampling was realized\\
by performing independent  Bernoulli trials with R function  \texttt{rbinom}. The \\
time complexity of each function is provided in the Suppl. Mat.\\
\hline
\end{tabular}}

\end{table}
\end{comment}

We may translate the number of unique samples $m_0$ in Tab.~\ref{tab:unique} to actual time complexity to examine the impact of each subsampling procedure on computational efficiency. For example, $k=3,000$ and $p=14$  in utility experiment 3,  at which gradients were evaluated. Given the time complexity of SGD is $O(kpm_0)$, it implies the computational efficiency for MUSTww was 33\% higher than WOR and Poisson, leading to significant savings in running time.   Table \ref{tab:time} list the computational time in the two classification tasks in experiment 3, showing that DP-SGD with MUSTwo and MUSTww requires less time than the single-step sampling procedures. In the MNIST experiment, the saving in time by USTwo and MUSTww is significant compared to the single-step procedures, with or without DP guarantees. Fig \ref{fig:loss} and the loss plot in Suppl. Mat. suggest  similar iterations to convergence across all procedures.
%but also significantly reduced the total execution time from approximately $860$ seconds to about $630$ seconds.
\begin{table}[!htb]
\centering
\caption{Total computation time of the DP-SGD algorithm in the adult (15 iterations) and MNIST (100 iterations) experiments} \label{tab:time} \vspace{-3pt}
\setlength{\tabcolsep}{2pt}
\resizebox{0.75\columnwidth}{!}{
\begin{tabular}{l|c|ccc|cc}
\toprule  
 & Poisson (non-priv) & Poisson & WOR & WR & MUSTow& MUSTww \\
\midrule
Adult ($\times10^{-2}$) & 4.63 & 5.65 & 8.82 & 7.20 & 5.02 & 4.78 \\
MNIST & 867.25 & 874.90 & 862.46 & 869.21 & 636.65 & 632.02\\
\bottomrule
\end{tabular}}
\vspace{-15pt}
\end{table}
\begin{comment}
\begin{table}[!htb]
\centering\vspace{-3pt}
\caption{Comparison of accuracy and training time for classification on the adult data and MNIST data} \label{tab:experiment3} \vspace{-3pt}
\setlength{\tabcolsep}{2pt}
\resizebox{0.5\textwidth}{!}{
\begin{tabular}{lccc|ccc}
\toprule  
 & \multicolumn{3}{c|}{\textbf{Adult}} & \multicolumn{3}{c}{\textbf{MNIST}}\\
\cline{2-7}
$\mathcal{S}$& Test & $\sigma m$ & Total time & Test  & Time per iteration (sec) & Total time  \\
 &accuracy&&($\times 10^{-2}$ sec)& accuracy & mean $\pm$ SD (min, max) & (sec) \\
\midrule
Poisson &75.19\%&NA&4.63& 88.28\%  & 2.89 $\pm$ 0.40 (2.42, 4.12) & 867.25\\
DP-Poisson &74.83\%&2.68&5.65& 76.14\% & 2.92 $\pm$ 0.40 (2.45, 4.56) & 874.90\\
WOR &74.93\%&2.68&8.82& 78.60\% & 2.88 $\pm$ 0.39 (2.48, 4.35) & 862.46\\
WR &74.92\%&2.67&7.20& 78.92\% &2.90 $\pm$ 0.41 (2.51, 4.26) & 869.21\\
\midrule
MUST.ow&75.33\%&2.21&5.02& 80.09\% & 2.12 $\pm$ 0.31 (1.82, 3.15) & 636.65\\
MUST.ww &74.90\%&2.21&4.78& 81.43\%& 2.11 $\pm$ 0.31 (1.80, 3.38) & 632.02\\
\bottomrule
\end{tabular}}
\vspace{-15pt}
\end{table}
\end{comment}

\vspace{-3pt}\section{Discussion}\label{sec:conclusion}
We introduced a subsampling family named MUST.
%mechanism comprises two steps: sample without replacement from the entire dataset and then sample counts for the distinct data points in the previous step from a multinomial distribution with equal probabilities. Similarly, the MUSTwo mechanism is composed of sampling with replacement and sampling without replacement, and the MUSTww mechanism is composed of two sampling with replacement steps. 
We studied the PA effects of 2-stage MUST procedures on a generic DP mechanism theoretically and empirically. We showed that  MUSTwo is equivalent to the one-stage WR in PA, and MUSTww and MUSTow  yield the strongest PA effect on $\epsilon$ compared to the commonly used one-stage sampling procedures (Poisson sampling, WOR, WR) while the PA effect on $\delta'$ varies case by case. %In our simulation examples, it seems that MUST has the best PA effects when $\epsilon$ is either large  or small. In the former case, MUST shrinks $\epsilon$ significantly with a small sacrifice on $\delta$; in the latter case, MUST decreases $\delta$ significantly and as always shrinks $\epsilon$.
We introduced new notions including strong, type I, and type II weak PA effects, and aligned privacy profiles to effectively study and understand the PA of subsampling on a base DP mechanism.
We conducted the privacy loss analysis over the $k$-fold composition of MUST-subsampled Gaussian mechanisms and provided numerical results using FFA.  
%The R codes for the PA effect comparison and Python codes for privacy composition analysis in the simulation study are available at \url{https://github.com/zhao-xingyuan/Hsubsampling}. The latter part is based on the Github repository \footnote{\hspace{1pt}\url{https://github.com/DPBayes/PLD-Accountant}} for \cite{koskela2021tight}, which covers the privacy profile analysis for two cases of using Poisson subsampling and WOR in the subsampled Gaussian mechanism. We add the codes for computing the privacy profile for the case of using WR in the subsampled Gaussian mechanism and provide the privacy profile computation codes for the subsampled Gaussian mechanism with MUST.
%There are some limitations in the current work which we expect can be resolved with further investigation.  First, this work is largely conceptual and theoretical and does not apply  MUST to any real applications.
Our empirical results on the utility of MUST suggest that  MUST can deliver similar or improved utility and stability in PP outputs compared to one-stage subsampling schemes at similar privacy loss with improved computational efficiency.
%Toward that end, we set the amplified privacy loss per iteration to be the same across different subsampling schemes and back-calculate the scale for the Gaussian mechanism that perturbs SGD in each iteration. % we would study how to determine the corresponding Gaussian noise variance that needs to be injected in each iteration of DP-SGD given total privacy costs $(\epsilon, \delta)$, which serves as an inverse problem of calculating $\delta(\epsilon)$ given the noise scale implemented in this paper. In a broader sense, whether there are possible analytical relationships between $\delta(\epsilon)$ and the Gaussian noise variance for DP-SGD with MUST, or with other sampling schemes, in addition to the moment accounting results for Poisson subsampling.
MUST can be used in procedures such as stochastic optimization, boosting, or subsampling bootstrap due to its saving in privacy and computational cost, especially in large datasets.

As for choosing between  MUSTwo and MUSTww, % the 2-step MUST procedures, MUSTwo is equivalent to WR in PA, utility, and composition of generated subsets, despite the differences in the actual sampling scheme. Between MUSTwo and MUSTww, 
it depends on the trade-offs among the PA effects, utility, and computational efficiency. If $\delta'$ is on the same order, MUSTww may be preferred given its smaller $\epsilon'$ and less time for sampling, though  the experiments suggest the two are largely similar in utility. The choice may also depend on specific problems. For example,  MUSTww should be used   in the double bootstrap inferential procedure \cite{lee1999class} as this is how double bootstrap is defined. The BLB procedure is a MUSTow procedure that generates $\ge2$ subsets in stage II per stage I subset, where the number of subset in each sampling stage is motivated by efficient resampling-based statistical inference in large  datasets.

Regarding the choice of hyperparameters in MUST, $m$ is driven by specific problems. For example, $m$ in in stochastic optimization (e.g DP-SGD) may depend on the data type/size, computational cost, model accuracy, and privacy budget, etc, but often ranges from 0.001 to 0.1 \citep{abadi2016deep, bu2020deep}. In general, the larger $m/n$ or $b$ is, the weaker the PA effect is in 2-stage MUST. The sensitivity analysis in utility experiment 1 suggests that the stability of an output may decrease with increasing $b$ while the accuracy is rather stable or improved. %depending on the parameters being estimated. The trend may change if $m/n$ takes different values. Hence,  From a utility standpoint, the choice of 
Choosing $b$ would  also consider $m/n$, output types, and the trade-off with privacy loss. In summary, we recommend that users examine multiple $(m, b)$ values when employing a MUST procedure to assess the privacy-utility trade-off and identify appropriate values for their specific problems. %provided that the privacy loss, if any, during this decision-making process is taken into consideration.

There are several future research directions we plan to explore: 1)  improve PP interval estimation via double bootstrap, leveraging its inherent PA effect; 2)  integrate the inherent PA effect of BLB to generate PP inference; 3) extend the current work to other DP variants such as RDP, GDP, zCDP and other DP mechanisms such as $k$-fold MUST$\;\circ\;$Laplace mechanism. %this work focuses on the PA of MUST in the setting of $(\epsilon,\delta)$-DP and the $k$-fold  privacy loss profile of MUST$\circ$Gaussian mechanism, it would be interesting to 
The Python and R code in this work is available
at \url{https://github.com/CZchikage}.  %\url{https://github.com/zhao-xingyuan/MUST.git}.   and

%\section*{Supplementary Materials}

\bibliographystyle{apalike}
\bibliography{ref}

%%%%%%%%%%%%%%%%%%%%%%%%%%%%%%%%%%%%%%%%%%%%%%%%%%%%%%%%%%%%
\newpage
\section*{Supplementary Materials}
\setcounter{page}{1}
\setcounter{figure}{0}
\setcounter{table}{0}
\renewcommand{\thetable}{S\arabic{table}}
\renewcommand{\thefigure}{S\arabic{figure}}

\crefalias{sectiom}{supp}

\newcounter{suppcounter}
\setcounter{suppcounter}{0}
 \refstepcounter{suppcounter}%Increases the counter by 1 and makes it visible for the referencing mechanism so that you can use \label afterward.
\def\thesection{S\thesuppcounter}

\setcounter{tocdepth}{0} % Temporarily disable ToC entries
\addtocontents{toc}{\protect\setcounter{tocdepth}{2}} 
\renewcommand{\contentsname}{}
\vspace{-20pt}
\newcommand{\supptoc}{
  \begingroup
  \setlength{\cftbeforesubsecskip}{10pt}
  \tableofcontents
  \endgroup
}
\supptoc

\section{Proof of Theorem 4}\label{proof:H amplification}
\begin{proof}
The proof of MUSTow follows the proof of Theorem 10 on the PA bound for WR in \cite{balle2018privacy}. Both WR and MUSTow output a multiset. Consider $\mbox{MUST}(n,b,m)(\cdot): X\to\mathbb{P}(Y)$ of releasing a multiset $y$ of total sample size $m$ with respect to relation $X\simeq_{S}X'$. The total variation distance between $\omega=\mbox{MUST}(n,b,m;X)$ and $\omega'=\mbox{MUST}(n,b,m;X')$ is given by
\begin{align*}
    \eta=\mbox{TV}(\omega,\omega')=\sup_{A\in\mathcal{F}}|P_{\omega}(A)-Q_{\omega'}(A)|=\frac{b}{n}\left(1-\left(1-\frac{1}{b}\right)^m\right).
\end{align*}
Let $v$ and $v'$ be the different elements and $x_0=X\cap X'$ be the comment parts between $X$ and $X'$. The supreme in the equation above is achieved when $P_{\omega}(A)$ and $Q_{\omega'}(A)$ are the empirical distributions with non-zero occurrence of $v$, in which $P_{\omega}(A)=\frac{b}{n}\left(1-\left(1-\frac{1}{b}\right)^m\right)$ and $Q_{\omega'}(A)=0$. Let $\omega_0=\mbox{MUST}(n,b,m;x_0)$. $\omega_1$ is the distribution constructed from sampling $\tilde{y}$ from $\tilde{\omega}_1=\mbox{MUST}(n,b-1,m-1;X)$ and adding one $v$ to $\tilde{y}$ to obtain $y$ (i.e., $y=\tilde{y}\cup\{v\}$). Likewise, sampling $y'$ from $\omega'_1$ is obtained by adding one $v'$ to a multiset sampled from $\tilde{\omega}'_1=\mbox{MUST}(n,b-1,m-1;X')$.

We construct appropriate distance-compatible couplings as follows. Let $\pi_{1,1}$ be the distribution of $(y,y')$ where $y$ is sampled from $\omega_1$ and $y'$ is obtained from replacing each $v$ in $y$ by $v'$. $\pi_{1,1}$ is a $d_{\simeq_s}-$compatible coupling between $\omega_1$ and $\omega'_1$ ($d_Y(y,y')$: number of $v'$ in $y'$). Using $\omega=(1-\eta)\omega_0+\eta\omega_1$ from the construction of the maximal coupling, we get for every $j\geq1$,
$$
   \omega_1(Y_j)=\frac{\omega(Y_j)-(1-\eta)\omega_0(Y_j)}{\eta}=\frac{\omega(Y_j)}{\eta}=\frac{\Pr_{y\sim\omega}[y_v=j]}{\eta}=\frac{\frac{b}{n}\binom{m}{j}(\frac{1}{b})^j(1-\frac{1}{b})^{m-j}}{\eta}. 
$$
$\omega_0(Y_j)=0$ since the support of $\omega_0$ are multisets not containing $v$. By Theorem 7 in \citet{balle2018privacy}, the distributions $\mu_1=\M\circ\omega_1$ and $\mu'_1= \M\circ\omega'_1$ satisfy
\begin{equation*}
    \eta D_{e^\epsilon}(\mu_1||\mu'_1)\leq \eta\sum_{j=1}^m\omega_1(Y_j)\delta_{\mathcal{M},j}(\epsilon)=
    \sum_{j=1}^m\frac{b}{n}\binom{m}{j}\left(\frac{1}{b}\right)^j\left(1-\frac{1}{b}\right)^{m-j}\delta_{\mathcal{M},j}(\epsilon).
\end{equation*}
We can also build a $d_{\simeq_s}$-compatible coupling between $\omega_1$ and $\omega_0$ by first sampling $y$ from $\omega_1$ and then replacing each $v$ in $y$ by an element sampled uniformly at random from $x_0$. We have $\eta D_{e^\epsilon}(\mu_1||\mu_0)\leq\eta\sum_{j=1}^m\omega_1(Y_j)\delta_{\mathcal{M},j}(\epsilon)=\sum_{j=1}^m\frac{b}{n}\binom{m}{j}\left(\frac{1}{b}\right)^j\left(1-\frac{1}{b}\right)^{m-j}\delta_{\mathcal{M},j}(\epsilon).$
By Theorem 2 in \citet{balle2018privacy}, 
\begin{equation*}
    \delta_{\mathcal{M}'}(\epsilon')\leq\sum_{j=1}^m\frac{b}{n}\binom{m}{j}\left(\frac{1}{b}\right)^j\left(1-\frac{1}{b}\right)^{m-j}\delta_{\mathcal{M},j}(\epsilon).
\end{equation*}

For MUSTwo and MUSTww, the steps of the proofs are similar to  MUSTow. The probability an element appears $j$ times after the intermediate WR($n,b$) step, which is $\binom{b}{j}\left(\frac{1}{n}\right)^j\left(1-\frac{1}{n}\right)^{b-j}$ for $1\leq j\leq b$, and the probability of the element appears in the subsample after the subsampling in the second stage given its $j$ occurrences is $1-\frac{\binom{b-j}{m}}{\binom{b}{m}}$ for WOR($b,m$) and $\left(1-\left(1\!-\!\frac{j}{b}\right)^m\right)$ for WR($b,m$), respectively, leading to the probability that an element appears in the final subsample  $\eta\!=\!\sum_{j=1}^b\binom{b}{j}\left(\frac{1}{n}\right)^j\left(1\!-\!\frac{1}{n}\right)^{b-j}\left(1\!-\!\frac{\binom{b-j}{m}}{\binom{b}{m}}\right)$ for MUSTwo and
$\eta\!=\!\sum_{j=1}^b\!\binom{b}{j}\!\left(\frac{1}{n}\right)^j\!\left(1\!-\!\frac{1}{n}\right)^{b-j}\!\left(1\!-\!\left(1\!-\!\frac{j}{b}\right)^m\right)$ for MUSTww, respectively. % Algebraically,  the binomial coefficient $\binom{n}{a}=\frac{n!}{a!(n-a)!}$ if $n, a$ are integers and $0\leq a\leq n$; $\binom{n}{a}=0$ otherwise.

The coefficient of the group privacy terms $\delta_u$ in the $\delta'$ formula is the probability that an element appears $u$ ($1\leq u\leq m$) times in the final subsample. The probability an element appears $j$ times after the intermediate WR($n,b$) step is $\binom{b}{j}\left(\frac{1}{n}\right)^j\left(1-\frac{1}{n}\right)^{b-j}$ for $1\leq j\leq b$. The probability of the element appearing $u$ times in the final subsample after the subsampling in the second stage given its $j$ occurrences is $\frac{\binom{j}{u}\binom{b-j}{m-u}}{\binom{b}{m}}$ for WOR($b,m$) and $\binom{m}{u}\left(\frac{j}{b}\right)^u\left(1-\frac{j}{b}\right)^{m-u}$ for WR($b,m$), respectively. Since WOR($b,m$) requires $b\geq m$, the possible values of $u$ depend on the number of occurrences $j$ of the element available after the first subsampling stage: when $j\leq m$, $1\leq u\leq j$; when $j>m$, $1\leq u\leq m$. Thus $\delta'=\sum_{j=1}^m\binom{b}{j}\left(\frac{1}{n}\right)^j\left(1-\frac{1}{n}\right)^{b-j}\sum_{u=1}^j\frac{\binom{j}{u}\binom{b-j}{m-u}}{\binom{b}{m}}\delta_u+\sum_{j=m+1}^b\binom{b}{j}\left(\frac{1}{n}\right)^j\left(1-\frac{1}{n}\right)^{b-j}\sum_{u=1}^m\frac{\binom{j}{u}\binom{b-j}{m-u}}{\binom{b}{m}}\delta_u$ for MUSTwo. While WR($b,m$) has no restrictions on the relative magnitude between $b$ and $m$, thus $1\leq u\leq m$ regardless of $j$  and $\delta'=\sum_{j=1}^b\binom{b}{j}\left(\frac{1}{n}\right)^j\left(1-\frac{1}{n}\right)^{b-j}\sum_{u=1}^m\binom{m}{u}\left(\frac{j}{b}\right)^u\left(1-\frac{j}{b}\right)^{m-u}\delta_u$ for MUSTww.
\end{proof}

\vspace{-9pt}\refstepcounter{suppcounter}
\def\thesection{S\thesuppcounter}
\section{Proof of Corollary 5} 
\begin{proof}
    \begin{align*}
    &\eta_{\text{MUSTwo}}=\sum_{j=1}^b\binom{b}{j}\left(\frac{1}{n}\right)^j\left(1-\frac{1}{n}\right)^{b-j}\left(1-\frac{\binom{b-j}{m}}{\binom{b}{m}}\right)\\
    &=1-\left(1-\frac{1}{n}\right)^b-\sum_{j=1}^b\binom{b}{j}\left(\frac{1}{n}\right)^j\left(1-\frac{1}{n}\right)^{b-j}\frac{\binom{b-j}{m}}{\binom{b}{m}}\\
    &=1-\left(1-\frac{1}{n}\right)^b-\sum_{j=1}^b\frac{b!}{j!(b-j)!}\frac{(b-j)!}{m!(b-j-m)!}\frac{m!(b-m)!}{b!}\left(\frac{1}{n}\right)^j\left(1-\frac{1}{n}\right)^{b-j}\\
    &=1-\left(1-\frac{1}{n}\right)^b-\sum_{j=1}^b\frac{(b-m)!}{j!(b-j-m)!}\left(\frac{1}{n}\right)^j\left(1-\frac{1}{n}\right)^{b-j}\\
    &=-\left(\sum_{j=1}^{b-m}\frac{(b\!-\!m)!}{j!(b\!-\!j\!-\!m)!}\left(\frac{1}{n}\right)^j\left(1\!-\!\frac{1}{n}\right)^{b-m-j}\!+\!\sum_{j=b-m+1}^{b}\frac{(b\!-\!m)!}{j!(b\!-\!j\!-\!m)!}\left(\frac{1}{n}\right)^j\left(1\!-\!\frac{1}{n}\right)^{b-m-j}\right)\!\left(1-\frac{1}{n}\right)^m+1-\left(1-\frac{1}{n}\right)^b\\
    &=1-\left(1-\frac{1}{n}\right)^b-\left(1-\left(1-\frac{1}{n}\right)^{b-m}+0\right)\left(1-\frac{1}{n}\right)^m\\
    &=1-\left(1-\frac{1}{n}\right)^b-\left(1-\frac{1}{n}\right)^m+\left(1-\frac{1}{n}\right)^b\\
    &=1-\left(1-\frac{1}{n}\right)^m=\eta_{\text{WR}}.
\end{align*}
\begin{align*}   &\delta'_{\text{MUSTwo}}=\sum_{j=1}^b\sum_{u=1}^{\min(j, m)}\binom{b}{j}\frac{\binom{j}{u}\binom{b-j}{m-u}}{\binom{b}{m}}\left(\frac{1}{n}\right)^j\left(1-\frac{1}{n}\right)^{b-j}\delta_u\\
    &=\sum_{j=1}^b\sum_{u=1}^{\min(j, m)}\frac{b!}{j!(b\!-\!j)!}\frac{j!}{(j\!-\!u)!u!}\frac{(b\!-\!j)!}{(m\!-\!u)!(b\!-\!j\!-\!m\!+\!u)!}\frac{m!(b\!-\!m)!}{b!}\left(\frac{1}{n}\right)^j\left(1\!-\!\frac{1}{n}\right)^{b-j}\delta_u\\
    &=\sum_{j=1}^b\sum_{u=1}^{\min(j, m)}\frac{m!}{u!(m-u)!}\frac{(b-m)!}{(j-u)!(b-j-m+u)!}\left(\frac{1}{n}\right)^j\left(1-\frac{1}{n}\right)^{b-j}\delta_u\\
    &=\sum_{j=1}^b\sum_{u=1}^{\min(j, m)}\binom{b-m}{j-u}\binom{m}{u}\left(\frac{1}{n}\right)^u\left(1-\frac{1}{n}\right)^{m-u}\left(\frac{1}{n}\right)^{j-u}\left(1-\frac{1}{n}\right)^{b-m-j+u}\delta_u\\
    &=\sum_{u=1}^m\sum_{j=u}^{b}\binom{b-m}{j-u}\binom{m}{u}\left(\frac{1}{n}\right)^u\left(1-\frac{1}{n}\right)^{m-u}\left(\frac{1}{n}\right)^{j-u}\left(1-\frac{1}{n}\right)^{b-m-j+u}\delta_u\\
    &=\sum_{u=1}^m\sum_{j=u}^{b}\binom{b-m}{j-u}\left(\frac{1}{n}\right)^{j-u}\left(1-\frac{1}{n}\right)^{b-m-j+u}A_u\delta_u, 
\end{align*}
where $A_u=\binom{m}{u}\left(\frac{1}{n}\right)^u\left(1-\frac{1}{n}\right)^{m-u}$. Note that $\sum_{j=u}^b\binom{b-m}{j-u}\left(\frac{1}{n}\right)^{j-u}\left(1-\frac{1}{n}\right)^{b-m-j+u}A_u\delta_u=A_u\delta_u$ for $u=1,\dots, m $, then we have 
\begin{align*}
&\delta'_{\text{MUSTwo}}=A_1\delta_1+A_2\delta_2+\cdots+A_m\delta_m\\
&=\binom{m}{1}\left(\frac{1}{n}\right)^1\left(1-\frac{1}{n}\right)^{m-1}\delta_1+\cdots+\binom{m}{m}\left(\frac{1}{n}\right)^m\left(1-\frac{1}{n}\right)^{m-m}\delta_m\\
&=\sum_{j=1}^m\left(\frac{1}{n}\right)^j\left(1-\frac{1}{n}\right)^{m-j}\delta_j=\delta'_{\text{WR}}.
\end{align*}
Taken together with the interpretation of the coefficient for $\delta_j$, which is the probability that an element in the original data appears $j$ ($1\leq j\leq m$) times in the final subsample, MUSTwo ad WR are equivalent subsampling procedures.
\end{proof}

\refstepcounter{suppcounter}
\def\thesection{S\thesuppcounter}
\section{Proof of Proposition 6}\label{proof:epsH}
\begin{proof}
$\eta_{\text{WR}}=1-(1-\frac{1}{n})^m$, $\eta_{\text{MUSTow}}=\frac{b}{n}(1-(1-\frac{1}{b})^m)=:f(b), b\in[1,n]$. The first and second derivatives of $f(b)$ are
\begin{equation*}
    f'(b)=\frac{1}{n}-\frac{1}{n}\left(1-\frac{1}{b}\right)^m-\frac{m}{nb}\left(1-\frac{1}{b}\right)^{m-1}.
\end{equation*}
\begin{equation*}
    f''(b)=-\frac{m(m-1)}{nb^3}\left(1-\frac{1}{b}\right)^{m-2}\leq 0,
\end{equation*}
where equality only holds when $b=1$ or $m=1$. Thus, $f'(b)$ is a decreasing function over $[1,n]$. We have
\begin{equation*}
    f'(b)\geq f'(n)=\frac{1}{n}\left(1-\left(1-\frac{1}{n}\right)^{m-1}\left(1-\frac{m+1}{n}\right)\right)>0
\end{equation*}
Thus $f(b)$ is an increasing function over $[1,n]$. we have $f(b)\leq f(n)=1-(1-\frac{1}{n})^m$, i.e., $\eta_{\text{MUSTow}}\leq\eta_{\text{WR}}$. $\eta_{\text{MUSTow}}=\eta_{\text{WR}}$ only in the trivial case of $b=n$, when MUSTow$(n, b, m)$ reduces to WR$(n, m)$.  By binomial expansion, we have
\begin{align*}
    &\eta_{\text{WR}}(m)=1-\left(1-\frac{1}{n}\right)^m\\
    &=\dbinom{m}{1}\frac{1}{n}\left(1-\frac{1}{n}\right)^{m-1}+\dbinom{m}{2}\left(\frac{1}{n}\right)^2\left(1-\frac{1}{n}\right)^{m-2}+\cdots+\dbinom{m}{m}\left(\frac{1}{n}\right)^m\\
    &=\frac{m}{n}\left(\dbinom{m}{0}\left(1-\frac{1}{n}\right)^{m-1}+\frac{1}{2}\dbinom{m\!-\!1}{1}\frac{1}{n}\left(1-\frac{1}{n}\right)^{m-2}+\cdots+\frac{1}{m}\dbinom{m\!-\!1}{m\!-\!1}\left(\frac{1}{n}\right)^{m-1}\right)\\
    &\leq\frac{m}{n}\left(\dbinom{m}{0}\left(1-\frac{1}{n}\right)^{m-1}+\dbinom{m\!-\!1}{1}\frac{1}{n}\left(1-\frac{1}{n}\right)^{m-2}+\cdots+\dbinom{m\!-\!1}{m\!-\!1}\left(\frac{1}{n}\right)^{m-1}\right)\\
    &=\frac{m}{n}\left(\frac{1}{n}+1-\frac{1}{n}\right)^{m-1}=\frac{m}{n}=\eta_{\text{WOR}}(m).
\end{align*}
Thus for any $m\in [1, n]$, $\eta_{\text{WR}}\leq\eta_{\text{WOR}}$, where the equality only holds when $m=1$.
\end{proof}

%-----------------------------------------
\refstepcounter{suppcounter}
\def\thesection{S\thesuppcounter}
\section{3D plots version of Figure 5}\label{sec: 3D plot}
\begin{figure}[!htbp]
\vspace{-6pt}\centering
(a) and (b): $\eta$ as a function of $(b,m)$ in  MUST$\circ\M$ \vspace{-6pt}
\subfloat[MUSTow]{{\includegraphics[width=0.33\textwidth]{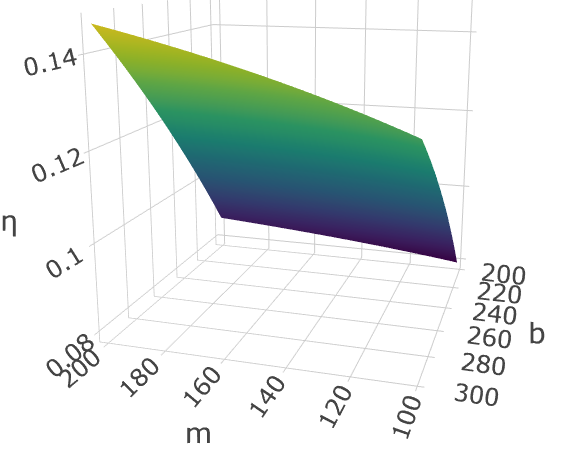}}}
%\subfloat[MUSTwo] {{\includegraphics[width=0.33\textwidth]{figures/3D_eta_WO1.png}}}
\subfloat[MUSTww] {{\includegraphics[width=0.33\textwidth]{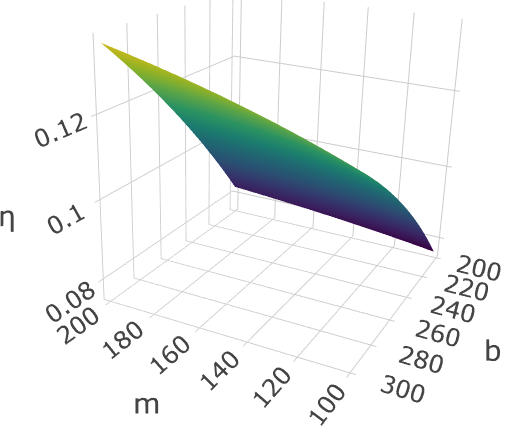}}}\\
\vspace{12pt}
(c) and (d): $\delta'-\delta$ as a function of $(b,m)$ in MUST$\circ$Laplace mechanism ($\Delta/\sigma\!=\!1$)\vspace{-6pt}
\subfloat[MUSTow]{{\includegraphics[width=0.33\textwidth]{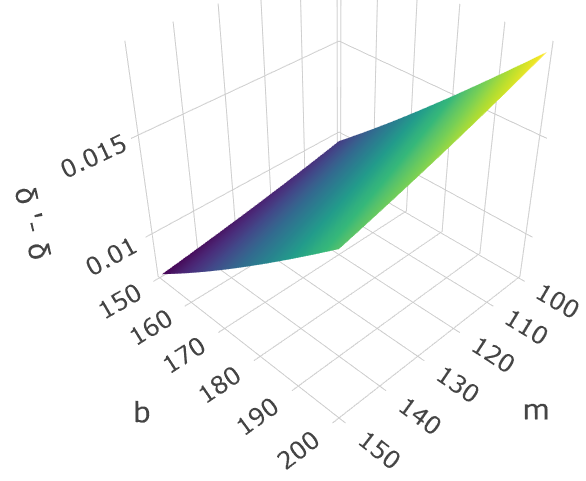}}}
%\subfloat[MUSTwo] {{\includegraphics[width=0.33\textwidth]{figures/3D_delta_WO_Lap1.png}}}
\subfloat[MUSTww] {{\includegraphics[width=0.33\textwidth]{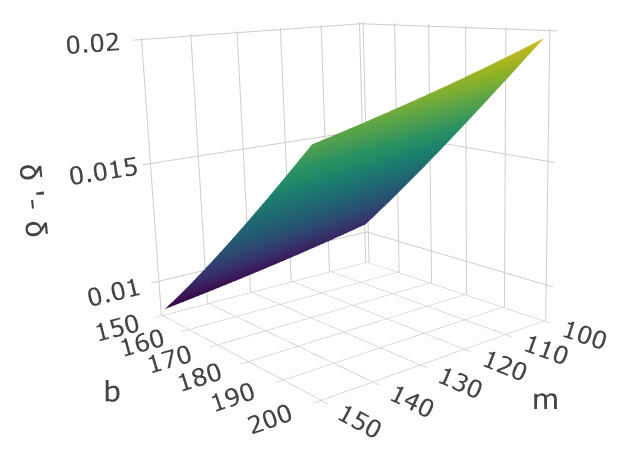}}}\\
\vspace{12pt}
(e) and (f): $\delta'-\delta$ as a function of $(b,m)$ in MUST$\circ$Gaussian mechanism ($\Delta/\sigma\!=\!1$)\\
\subfloat[MUSTow]{{\includegraphics[width=0.33\textwidth]{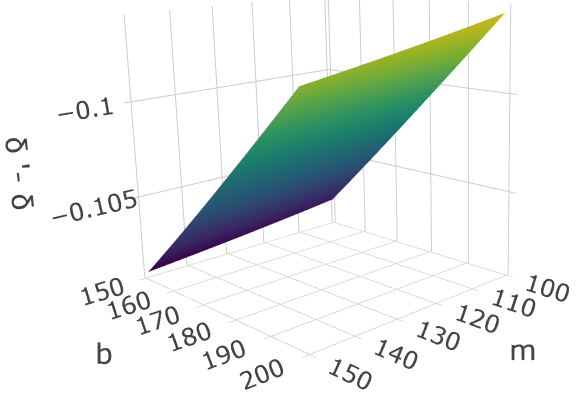}}}
%\subfloat[MUSTwo] {{\includegraphics[width=0.33\textwidth]{figures/3D_delta_WO_Gauss1.png}}}
\subfloat[MUSTww] {{\includegraphics[width=0.33\textwidth]{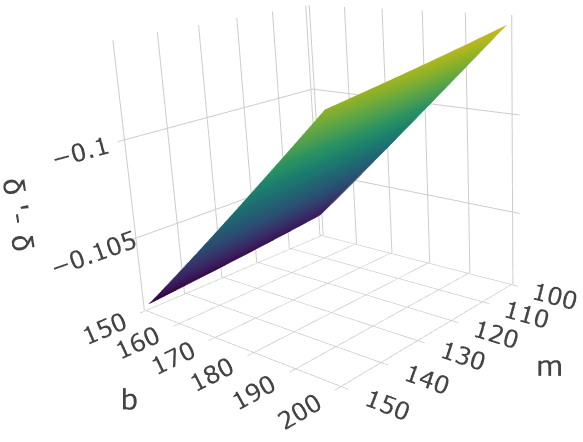}}}\\
\caption{3D plots of $\eta$ vs $(b, m)$ for MUST$\circ\M$ ($\M$ is a generic base mechanism)  and $(\delta'-\delta)$ vs $(b, m)$ for  MUST$\circ$ Laplace mechanism and  MUST$\circ$ Gaussian mechanism ($n=1000$, $\epsilon=1$ for the base mechanism)}\label{fig: 3D}
\end{figure}

\vspace{-18pt}
\refstepcounter{suppcounter}
\def\thesection{S\thesuppcounter}
\section{Privacy Loss Composition for MUST-Subsampled  Gaussian Mechanism}\label{proof:MUSTwoWW}

For MUSTwo$(n,b,m)$ with $\simeq_S$ neighboring relation
\begin{align}
f_{X}(t)=&\frac{1}{\sqrt{2\pi}\sigma}\!\sum_{l=0}^m\!\frac{b}{n}\binom{m}{l}\left(\!\frac{1}{b}\!\right)^l\left(\!1\!-\!\frac{1}{b}\!\right)^{m-l}e^{-\frac{(t-l)^2}{2\sigma^2}}+\frac{1}{\sqrt{2\pi}\sigma}\left(\!1\!-\!\frac{b}{n}\!\right)e^{-\frac{t^2}{2\sigma^2}},\notag\\
f_{X'}(t)=&\frac{1}{\sqrt{2\pi}\sigma}\!\sum_{l=0}^m\!\frac{b}{n}\binom{m}{l}\left(\!\frac{1}{b}\!\right)^l\left(\!1\!-\!\frac{1}{b}\!\right)^{m-l}e^{-\frac{(t+l)^2}{2\sigma^2}}+\frac{1}{\sqrt{2\pi}\sigma}\left(\!1\!-\!\frac{b}{n}\!\right)e^{-\frac{t^2}{2\sigma^2}},\notag\\
\mbox{The privacy loss from} & \mbox{outputting $t$ thus equals to} \\
\mathcal{L}_{X/{X'}}(t)&=\log\left(\frac{f_{X}(t)}{f_{X'}(t)}\right)=\log\!\left(\!\frac{\sum_{l=0}^m\frac{b}{n}\binom{m}{l}\left(\frac{1}{b}\right)^l\left(1\!-\!\frac{1}{b}\right)^{m-l}e^{-\frac{l^2}{2\sigma^2}}e^{\frac{tl}{\sigma^2}}\!+\!1\!-\!\frac{b}{n}}{\sum_{l=0}^m\frac{b}{n}\binom{m}{l}\left(\frac{1}{b}\right)^l\left(1\!-\!\frac{1}{b}\right)^{m-l}e^{-\frac{l^2}{2\sigma^2}}e^{-\frac{tl}{\sigma^2}}\!+\!1\!-\!\frac{b}{n}}\!\right), \notag
\end{align}
For MUSTww$(n,b,m)$ with $\simeq_S$ neighboring relation, the probability that the one data point that is differential between $X$ and $X'$  eventually occurs $l$ times in a subset $Y$ of size $m$ is $\sum_{j=1}^b\binom{b}{j}\left(\frac{1}{n}\right)^j\left(1-\frac{1}{n}\right)^{b-j}\binom{m}{l}\left(\frac{j}{b}\right)^l\left(1-\frac{j}{b}\right)^{m-l}$ for $l=1,\dots,m$ and $\sum_{j=1}^b\binom{b}{j}\left(\frac{1}{n}\right)^j\left(1-\frac{1}{n}\right)^{b-j}\left(1-\frac{j}{b}\right)^{m}+\left(1-\frac{1}{n}\right)^b$ for $l=0$. WLOG, we consider
{\small\begin{align*}
    f_{X}(t)&=\frac{1}{\sqrt{2\pi}\sigma}\sum_{l=0}^m\sum_{j=1}^b\binom{b}{j}\left(\frac{1}{n}\right)^j\left(1-\frac{1}{n}\right)^{b-j}\binom{m}{l}\left(\frac{j}{b}\right)^l\left(1-\frac{j}{b}\right)^{m-l}e^{-\frac{(t-l)^2}{2\sigma^2}}+\frac{1}{\sqrt{2\pi}\sigma}\left(1-\frac{1}{n}\right)^{b}e^{-\frac{t^2}{2\sigma^2}},\\
    f_{X'}(t)&=\frac{1}{\sqrt{2\pi}\sigma}\sum_{l=0}^m\sum_{j=1}^b\binom{b}{j}\left(\frac{1}{n}\right)^j\left(1-\frac{1}{n}\right)^{b-j}\binom{m}{l}\left(\frac{j}{b}\right)^l\left(1-\frac{j}{b}\right)^{m-l}e^{-\frac{(t+l)^2}{2\sigma^2}}+\frac{1}{\sqrt{2\pi}\sigma}\left(1-\frac{1}{n}\right)^{b}e^{-\frac{t^2}{2\sigma^2}},
\end{align*}}%
 The privacy loss from outputting $t$ thus equals to
{\small\begin{equation*}\label{eqn:loss.MUST_WW}
    \mathcal{L}_{X/{X'}}(t)\!=\!\log\left(\frac{f_{X}(t)}{f_{X'}(t)}\right)\!=\!\log\!\!\left(\frac{\sum_{l=0}^m\sum_{j=1}^b\binom{b}{j}\left(\frac{1}{n}\right)^j\left(1\!-\!\frac{1}{n}\right)^{b-j}\binom{m}{l}\left(\frac{j}{b}\right)^l\left(1-\frac{j}{b}\right)^{m-l}e^{-\frac{l^2}{2\sigma^2}}e^{\frac{tl}{\sigma^2}}+\left(1\!-\!\frac{1}{n}\right)^b}{\sum_{l=0}^m\sum_{j=1}^b\binom{b}{j}\left(\frac{1}{n}\right)^j\left(1\!-\!\frac{1}{n}\right)^{b-j}\binom{m}{l}\left(\frac{j}{b}\right)^l\left(1\!-\!\frac{j}{b}\right)^{m-l}e^{-\frac{l^2}{2\sigma^2}}e^{-\frac{tl}{\sigma^2}}+\left(1\!-\!\frac{1}{n}\right)^b}\right).
\end{equation*}}%
Since $\sum_{j=1}^b\binom{b}{j}\left(\frac{1}{n}\right)^j\left(1-\frac{1}{n}\right)^{b-j}\binom{m}{l}\left(\frac{j}{b}\right)^l\left(1-\frac{j}{b}\right)^{m-l}e^{-\frac{l^2}{2\sigma^2}}>0$ for $l=0,\dots,m$, the numerator in the $\mathcal{L}_{X/{X'}}(t)$ formula is a monotonically increasing function of $t$ and the denominator is a monotonically decreasing function of $t$. Therefore, \vspace{-3pt}
$$\mathcal{L}_{X/X'}(t)\to -\infty \mbox{ as }t\to -\infty \mbox{ and }\mathcal{L}_{X/X'}(t)\to \infty \mbox{ as }t\to \infty, $$\vspace{-3pt}
implying that the domain of $\mathcal{L}_{X/{X'}}(\mathbb{R})$ is $\mathbb{R}$.

\refstepcounter{suppcounter}
\def\thesection{S\thesuppcounter}
\section{Privacy Composition for Subsampled Gaussian Mechanisms with Poisson, WOR, and WR Sampling}\label{proof:existing sampling}
Section 6 of \citet{koskela2020computing} provides the privacy analysis of the subsampled Gaussian mechanism with the subsampling mechanisms as Poisson, WOR, and WR sampling. We summarize their main results below.

The privacy analysis of Poisson subsampling with both $\simeq_R$ and  $\simeq_S$ neighboring relations is equivalent to considering 
{\begin{align*}
    f_{X}(t)&=q\frac{1}{\sqrt{2\pi}\sigma}e^{-\frac{(t-1)^2}{2\sigma^2}}+(1-q)\frac{1}{\sqrt{2\pi}\sigma}e^{-\frac{t^2}{2\sigma^2}},\\
    f_{X'}(t)&=\frac{1}{\sqrt{2\pi}\sigma}e^{-\frac{t^2}{2\sigma^2}}.
\end{align*}}%
The privacy loss function then is 
\begin{align*}%\label{eqn:loss.Poisson}
    \mathcal{L}_{X/{X'}}(t)=\log\left(\frac{f_{X}(t)}{f_{X'}(t)}\right)=\log\left(qe^{\frac{2t-1}{2\sigma^2}}+(1-q)\right).
\end{align*}%
The domain $\mathcal{L}_{X/{X'}}(\mathbb{R})=(\log(1-q), \infty)$. The inverse function of $\mathcal{L}_{X/{X'}}(t)$ is 
\begin{align*}
    \mathcal{L}^{-1}_{X/{X'}}(s)=\sigma^2\log\left(\frac{e^s-(1-q)}{q}\right)+\frac{1}{2}.
\end{align*}%
The derivative of $\mathcal{L}^{-1}_{X/{X'}}(s)$ for Poisson subsampling admits a closed-form:
\begin{align*}
    \frac{d}{ds}\mathcal{L}^{-1}_{X/{X'}}(s)=\frac{\sigma^2 e^s}{e^s-(1-q)}.
\end{align*}%

For WOR with $\simeq_S$ neighboring relation, it suffices to consider
{\begin{align*}
    f_{X}(t)&=(m/n)\frac{1}{\sqrt{2\pi}\sigma}e^{-\frac{(t-1)^2}{2\sigma^2}}+(1-m/n)\frac{1}{\sqrt{2\pi}\sigma}e^{-\frac{t^2}{2\sigma^2}},\\
    f_{X'}(t)&=(m/n)\frac{1}{\sqrt{2\pi}\sigma}e^{-\frac{(t+1)^2}{2\sigma^2}}+(1-m/n)\frac{1}{\sqrt{2\pi}\sigma}e^{-\frac{t^2}{2\sigma^2}}.
\end{align*}}%
The privacy loss function is 
\begin{align*}\label{eqn:loss.WOR}
    \mathcal{L}_{X/{X'}}(t)=\log\left(\frac{(m/n)e^\frac{2t-1}{2\sigma^2}+(1-m/n)}{(m/n)e^\frac{-2t-1}{2\sigma^2}+(1-m/n)}\right).
\end{align*}%
$\mathcal{L}_{X/{X'}}(\mathbb{R})=\mathbb{R}$. The inverse function of $\mathcal{L}_{X/{X'}}(t)$ is 
{\small\begin{align*}
    \mathcal{L}^{-1}_{X/{X'}}(s)=\sigma^2\log\left(\left(\frac{n}{2m}e^{\frac{1}{2\sigma^2}}\right)\left(-(1-m/n)(1-e^s)+\sqrt{(1-m/n)^2(1-e^s)^2+4((m/n)e^{-\frac{1}{2\sigma^2}})^2e^s}\right)\right).
\end{align*}}%
Though not analytically provided in \citet{koskela2020computing}, a closed-form formula for the derivative of $\mathcal{L}^{-1}_{X/{X'}}(s)$ is also available for WOR, which we derived as follows: 
\begin{align*}
&\frac{d}{ds}\mathcal{L}^{-1}_{X/{X'}}(s)\\
=\;&\sigma^2
\frac{(1-\frac{m}{n})e^s\sqrt{(1-\frac{m}{n})^2(1-e^s)^2+4((\frac{m}{n})e^{-\frac{1}{2\sigma^2}})^2e^s}+(1-\frac{m}{n})^2(1-e^s)(-e^s)+2((\frac{m}{n})e^{-\frac{1}{2\sigma^2}})^2e^s}{-(1-\frac{m}{n})(1-e^s)\sqrt{(1-\frac{m}{n})^2(1-e^s)^2+4((\frac{m}{n})e^{-\frac{1}{2\sigma^2}})^2e^s}+(1-\frac{m}{n})^2(1-e^s)^2+4((\frac{m}{n})e^{-\frac{1}{2\sigma^2}})^2e^s}.
\end{align*}

For WR with $\simeq_S$ neighboring relation, we consider
\begin{align*}
    f_{X}(t)&=\frac{1}{\sqrt{2\pi}\sigma}\sum_{l=0}^m\binom{m}{l}\left(\frac{1}{n}\right)^l\left(1-\frac{1}{n}\right)^{m-l}e^{-\frac{(t-l)^2}{2\sigma^2}},\\
    f_{X'}(t)&=\frac{1}{\sqrt{2\pi}\sigma}\sum_{l=0}^m\binom{m}{l}\left(\frac{1}{n}\right)^l\left(1-\frac{1}{n}\right)^{m-l}e^{-\frac{(t+l)^2}{2\sigma^2}}.
\end{align*}
The privacy loss function is
\begin{equation*}\label{eqn:loss.WR}
    \mathcal{L}_{X/{X'}}(t)=\log\left(\frac{\sum_{l=0}^m\binom{m}{l}\left(\frac{1}{n}\right)^l\left(1-\frac{1}{n}\right)^{m-l}e^{-\frac{l^2}{2\sigma^2}}e^{\frac{tl}{\sigma^2}}}{\sum_{l=0}^m\binom{m}{l}\left(\frac{1}{n}\right)^l\left(1-\frac{1}{n}\right)^{m-l}e^{-\frac{l^2}{2\sigma^2}}e^{-\frac{tl}{\sigma^2}}}\right).
\end{equation*}

$\mathcal{L}_{X/{X'}}(\mathbb{R})=\mathbb{R}$. There are no closed-form formulas for $\mathcal{L}^{-1}_{X/{X'}}(s)$ and its derivative $\frac{d}{ds}\mathcal{L}^{-1}_{X/{X'}}(s)$ for WR. One needs to solve $\mathcal{L}_{X/{X'}}(t)=s$ to find $\mathcal{L}^{-1}_{X/{X'}}(s)$. We use Newton's method to solve $\mathcal{L}^{-1}_{X/{X'}}(s)$ and use difference quotient to approximate $\frac{d}{ds}\mathcal{L}^{-1}_{X/{X'}}(s)$ numerically.

%-------------------------------------------------
\refstepcounter{suppcounter}
\def\thesection{S\thesuppcounter}
\section{Fast Fourier account Algorithm}\label{sec: eg_composition}
\begin{algorithm}\caption{FA algorithm for privacy loss accounting \citep{koskela2021tight}} \label{algo:FA}
\textbf{Input}: privacy loss pdf $\omega$, overall privacy loss $\epsilon>0$; number of compositions $k$, convolution truncation bound $L$, number of grid points $r$ (even). \\
\textbf{Output}: an approximated  $\delta(\epsilon)$ and lower and upper bounds where the exact $\delta(\epsilon)$ resides.
\begin{algorithmic}[1]
\State Calculate grid length $\Delta x=2L/r$;
\State Set $s_i=-L+i\Delta x$  for $i=0,\ldots,r-1$;
\State Calculate $c_i\!=\!\Delta x \cdot \omega(s_i), c^-_i\!=\!\Delta x \cdot \min_{s\in[s_i,s_{i+1}]}\omega(s)$, and $c^+_i\!=\!\Delta x \cdot \max_{s\in[s_{i},s_{i+1}]}\omega(s)$ for $i=0,\ldots,r-1$;
%\State Define $\omega_{\min}(s)\!=\!\sum_{i=0}^{r-1}c^-_i D_{s_i}(s)$  %$\omega^{\infty}_{\min}(s)\!=\!\sum_{i\in\mathbb{Z}^+}c^-_i D_{s_i}(s)$,\newline
%and $\omega_{\max}(s)\!=\!\sum_{i=0}^{r-1}c^+_i D_{s_i}(s)$, %$\omega^{\infty}_{\max}(s)=\sum_{i\in\mathbb{Z}^+}c^+_i D_{s_i}(s)$,  
%where $D_{s_i}(\cdot)$ is the Dirac delta function centered at $s_i$; 
\State Set $\mathbf{c}=(c_0,...,c_{r-1})^T, \mathbf{c}^-=(c^-_0,...,c^-_{r-1})^T$ and $\mathbf{c}^+=(c^+_0,...,c^+_{r-1})^T$; 
$\mathbf{H}=\small\begin{pmatrix} \boldsymbol{0} & \mathbf{I}_{r/2}\\
\mathbf{I}_{r/2} & \boldsymbol{0} \end{pmatrix}$, where $\mathbf{I}_{r/2}$ is the identity matrix of dimension $r/2$; 
\State Evaluate {\small $\mathbf{u}\!=\!\mathbf{H}\mathcal{F}^{-1}(\mathcal{F}(\mathbf{H}\mathbf{c})^{\bigodot k}), \mathbf{u}^-\!=\!\mathbf{H}\mathcal{F}^{-1}(\mathcal{F}(\mathbf{H}\mathbf{c}^-)^{\bigodot k})$, 
$\mathbf{u}^+\!=\!\mathbf{H}\mathcal{F}^{-1}(\mathcal{F}(\mathbf{H}\mathbf{c}^+)^{\bigodot k})$}, where $\mathcal{F}$ is the Fourier transform, $\mathcal{F}^{-1}$ is the inverse Fourier transform, and $\bigodot k$ denotes the elementwise product of $k$ vectors;
\State Determine the starting point of the integral $i_\epsilon=\min\{i\in [0,r-1]: -L+i\Delta x>\epsilon\}$. 
\State Approximate $\delta(\epsilon)$ by $\sum_{j=i_\epsilon}^{r-1}(1-e^{\epsilon+L-j\Delta x})u_j$ and calculate its lower and upper bounds $(\delta_L(\epsilon),\;\delta_U(\epsilon))= \big(\sum_{j=i_\epsilon}^{r-1}\!(1\!-\!e^{\epsilon+L-j\Delta x})u^-_j,\; \sum_{j=i_\epsilon}^{r-1}\!(1\!-\!e^{\epsilon+L-j\Delta x})u^+_j\big)$\footnotemark.
\end{algorithmic} 
\end{algorithm}
\footnotetext{\hspace{2pt}There is an additional term $1\!-\!(1\!-\!\delta(\infty))^k$ in the formula for $\delta(\epsilon)$ in \cite{koskela2021tight}, which is 0 when $\mathcal{M}$ is Gaussian   mechanism (Lemma 11 in \cite{sommer2018privacy}).}

\vspace{-12pt}\refstepcounter{suppcounter}
\def\thesection{S\thesuppcounter}
\section{Privacy Composition for MUST-Subsampled Gaussian Mechanism} %\label{sec: eg_composition}
We present some numerical privacy loss results as presented in Section 3.3, where  MUST is integrated as a subsampling step in a multiple-release procedure such as DP-SGD algorithm or procedures utilizing simultaneous/parallel subsampling that employs Gaussian mechanisms as the base  $\mathcal{M}$. 

\begin{figure}[!htb]
\vspace{-6pt}\centering
 {\includegraphics[width=0.4\textwidth]{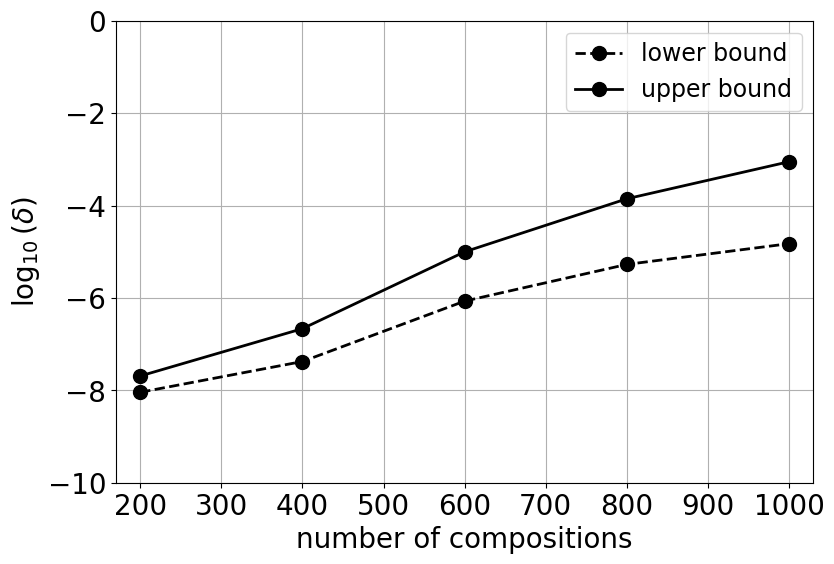}}
\vspace{-6pt}
\caption{The lower and upper bounds for $\delta'(\epsilon=1)$ over $k$-fold composition of Gaussian mechanism $\circ$ MUSTow ($n\!=\!10,000$, $m\!=\!200$, $b\!=\!118$, $\sigma\!=\!4$) calculated via the FA algorithm ($L\!=\!10,r\!=\!5\times10^5$). The FA algorithm produced numerical errors in the case of MUSTww.}\label{fig: compositions}
\vspace{-10pt}\end{figure}

In this example, we set the original data size $n=10,000$, the subset size $m=200$, $b=m^{0.9}=118$, and the scale of the Gaussian mechanism $\sigma=4$. We first obtain $\mathcal{L}^{-1}_{X/{X'}}(s)$ following the procedure in Section 3.3 for one application of the subsampled Gaussian mechanism, To numerically approximate the convolution over $k$ compositions in Theorem 4 via the FA algorithm in Algorithm 1, we need to specify the truncation bound $L$ and the number of grid points $r$. The finer the grid is, the more precise the approximation is but meanwhile the longer the computational time is. We use $L=10$ and $r=5\times10^5$ in this  example. The derivative of $\mathcal{L}^{-1}_{X/{X'}}(s)$ at each grid point $s$ is numerically calculated by difference quotient with the grid length $\Delta x=2L/r=4\times10^{-5}\approx0$,
\begin{equation*}
    d\mathcal{L}^{-1}_{X/{X'}}(s)/ds=\lim_{\Delta x\to 0}(\mathcal{L}^{-1}_{X/{X'}}(s+\Delta x)-\mathcal{L}^{-1}_{X/{X'}}(s))/\Delta x.
\end{equation*}
Finally, we obtain $ \omega_{X/X'}(s)$  per Definition 3 and feed it along with $L, n, k$ to the FA algorithm to get the privacy profile $\delta(\epsilon)$.  We present the $\delta$ values at $\epsilon=1$ for $k=(200, 400, 600,800, 1000)$-fold composition in Figure \ref{fig: compositions}. The lower and upper bounds reflect the numerical (approximation) errors of the FA algorithm in the calculated  $\delta$ values and are strict, meaning the exact $\delta$ falls between the bounds. As expected, the $\delta$ increases with the number of compositions; in other words, The overall privacy cost accumulates over compositions. 

\refstepcounter{suppcounter}
\def\thesection{S\thesuppcounter}
\section{Relations of \texorpdfstring{$\epsilon'$ vs $\epsilon$ and of $\delta'$ vs $\delta$}{}}\label{sec: relation}
We recommend using the aligned privacy profile plots as presented in Figure 4 to examine the PA effect of a subsampling procedure, for completeness, we still plotted  $\epsilon'$ vs $\epsilon$ and $\delta'$ vs $\delta$ in the same settings as used/specified in Figure 4. \textbf{Caution should be to be exercised when reading and interpreting these plots} as focusing on $\epsilon'$ vs $\epsilon$ without considering  $\delta'$ vs $\delta$ and vice versa can be misleading because both change simultaneously rather than one held at a constant. 

\begin{figure}[!htb]
\vspace{-12pt}\centering
\subfloat[$\epsilon'$ vs $\epsilon$: any base mechanism $\mathcal{M}$] {{\includegraphics[width=0.35\textwidth, trim= 0pt 5pt 15pt 30pt, clip]{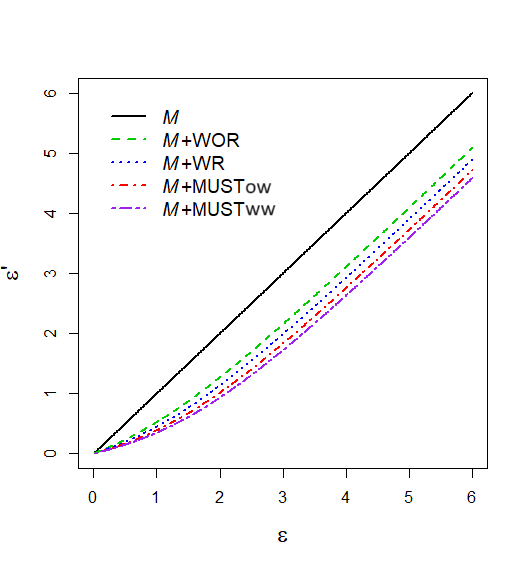}}}\\\vspace{-6pt}
\subfloat[\centering $\delta'$ vs $\delta$: Laplace mechanism ($\Delta_1/\sigma=1$)] {{\includegraphics[width=0.4\textwidth,  trim= 0pt 5pt 15pt 30pt, clip]{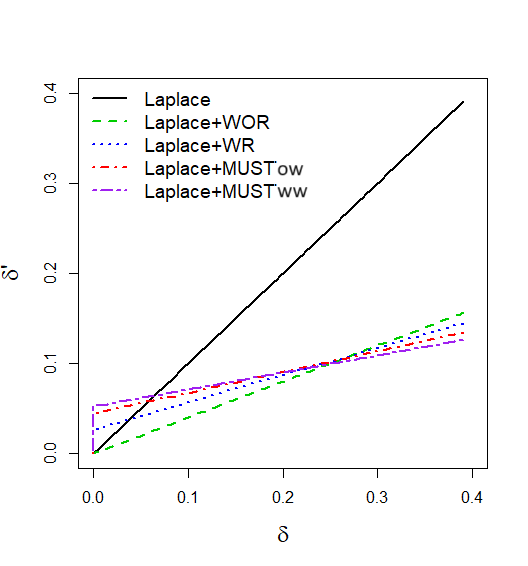}}}\hspace{6pt}
\subfloat[\centering $\delta'$ vs $\delta$: Gaussian mechanism ($\Delta_2/\sigma=1$)] {{\includegraphics[width=0.4\textwidth,  trim= 0pt 5pt 15pt 30pt, clip]{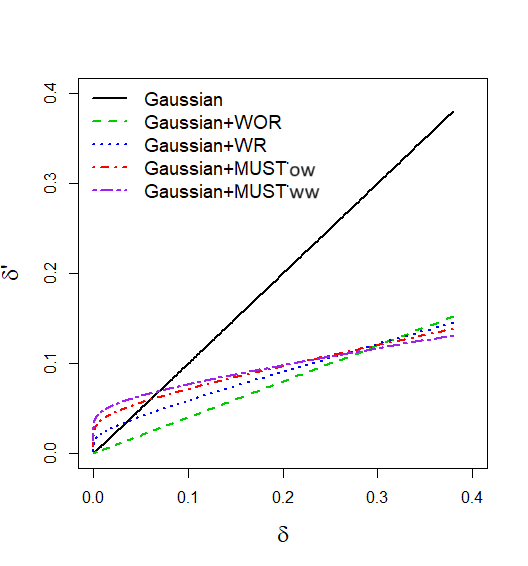}}}\\%
\subfloat[\centering $\delta'$ vs $\delta$: Laplace mechanism ($\Delta_1/\sigma=0.25$)] {{\includegraphics[width=0.4\textwidth,  trim= 0pt 5pt 15pt 30pt, clip]{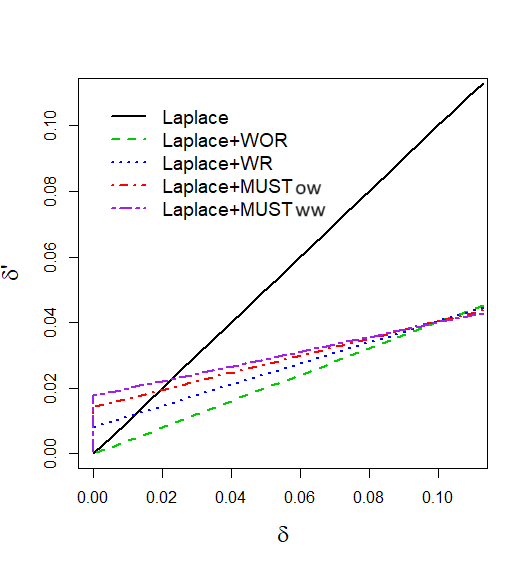}}}\hspace{6pt}
\subfloat[\centering $\delta'$ vs $\delta$: Gaussian mechanism ($\Delta_2/\sigma\!=\!0.25$)] 
{{\includegraphics[width=0.4\textwidth,  trim= 0pt 5pt 15pt 30pt, clip]{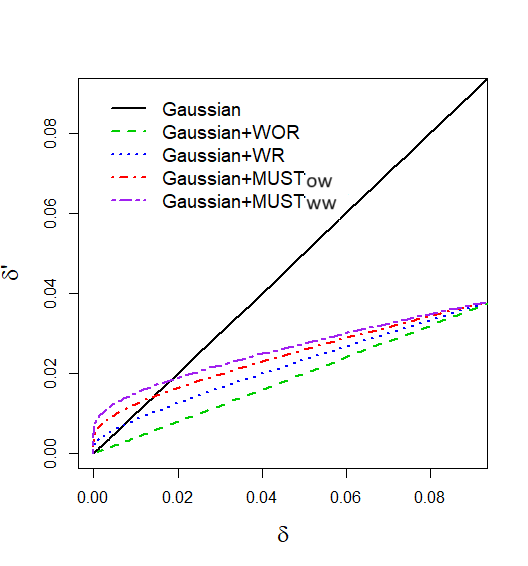}}}%
\vspace{-3pt}
\caption{$\epsilon'$ vs $\epsilon$ for any base mechanism $\mathcal{M}$ and $\delta'$ vs $\delta$ for the Laplace and Gaussian mechanisms for subsampling schemes WOR (Poisson), WR, MUSTow and MUSTww  when $n=1000$, $m=400$, and $b=500$}\label{fig: PA}\vspace{-9pt}
\end{figure}

Figure \ref{fig: PA}(a) depicts the relation between $\epsilon'$ vs $\epsilon$ for a general $\mathcal{M}$. Figure \ref{fig: PA}(b) and \ref{fig: PA}(c) depict the relation between $\delta'$ vs $\delta$ for the Laplace and Gaussian mechanisms, respectively.  The main findings can be summarized as follows. 1) As expected, $\epsilon'<\epsilon$ for all five subsampling schemes and MUSTww has the largest PA, followed by MUSTow, WR/MUSTwo, and WOR. The relations among MUSTow, WR, and WOR are as expected per Proposition 7. 2)  As noted in Remark 2, only  WOR (and Poisson sampling) provides strong PA as both $\epsilon'<\epsilon$  and  $\delta'<\delta$ whereas WR and MUST yield weak PA with $\epsilon'<\epsilon$ whereas $\delta'<\delta$ only for $\delta>0.039$ for WR/MUSTwo,  $\delta>0.058$ for MUSTow,  and $\delta>0.068$ for MUSTww when the base mechanism is Laplace and $\delta>0.036$ for WR/MUSTwo,  $\delta>0.060$ for MUSTow,  and $\delta>0.070$ for MUSTww when the base mechanism is Gaussian. 3)  The crossovers among the five  $\delta'$ curves occur around $(\delta,\delta')=(0.25, 0.10)$ when the base mechanism is Laplace and occur around $(\delta,\delta')=(0.23, 0.10)$ when the base mechanism is Gaussian. For both base mechanisms, before the crossover, the $\delta'$ value from the smallest to the largest is WOR, MUSTwo/WR, MUSTow, and MUSTww; after the crossover, the order is completely flipped.

%Notice that for a range of $\delta$ around $0$, the WR and MUST mechanisms inflate the baseline $\delta$ without subsampling, while the inflation decreases as $\delta$ increases. Figure \ref{fig: PA} (c) for the subsampled Gaussian mechanism has a similar pattern as Figure \ref{fig: PA} (b).

\clearpage

%----------------------------------------------
\refstepcounter{suppcounter}
\def\thesection{S\thesuppcounter}
\section{Examples of PA of $\mathcal{S}\!\circ\!\M$}\label{sec:PAexample}

\begin{table}[!h]
\centering
\caption{Examples of PA of $\mathcal{S}\!\circ\!\M$, where $\M$ is the Laplace and Gaussian mechanisms and $\mathcal{S}$ is WR, WOR (Poisson), and MUST ($n\!=\!1000, m\!=\!400, b\!=\!500$)}\label{tab:somepairs}
\resizebox{1\textwidth}{!}{
\begin{tabular}{@{}lll llll@{}}
\hline
$\mathcal{M}\circ\mathcal{S}$ & \multicolumn{6}{c}{$(\epsilon, \delta)$ for $\M$ and $(\epsilon', \delta')$ for $\M\circ\mathcal{S}$} \\
\hline 
\multicolumn{7}{c}{$\Delta/\sigma=0.25$}\\
\hline 
Laplace & 0.050, 0.095 & 0.500, 0 & 1.000, 0 & 2.000, 0 & 3.000, 0 & 4.500, 0 \\
\cline{2-7}
$\circ$ WOR & 0.020, 0.038 & 0.231, 0 & 0.523, 0 & 1.269, 0 & 2.156, 0 & 3.600, 0\\
$\circ$ WR & 0.017, 0.039 & 0.194, 0.001 & 0.449, $7.47\times 10^{-6}$ & 1.134, $5.65\times 10^{-11}$ & 1.987, $7.44\times 10^{-17}$ & 3.413, $1.22\times 10^{-26}$\\
%$\circ$ MUSTwo & 0.017, 0.039 & 0.194, 0.001 & 0.449, $7.47\times 10^{-6}$ & 1.134, $5.65\times 10^{-11}$ & 1.987, $7.44\times 10^{-17}$ & 3.413, $1.22\times 10^{-26}$\\
$\circ$ MUSTow & 0.014, 0.039 & 0.164, 0.003 & 0.388, $9.18\times 10^{-5}$ & 1.015, $1.06\times 10^{-8}$ & 1.834, $2.18\times 10^{-13}$ & 3.240, $2.26\times 10^{-21}$\\
$\circ$ MUSTww & 0.012, 0.039 & 0.145, 0.006 & 0.346, $6.07\times 10^{-4}$ & 0.932, $4.05\times 10^{-6}$ & 1.722, $1.84\times 10^{-8}$ & 3.111, $3.44\times 10^{-12}$\\
\hline
Gaussian & 0.050, 0.078 & 0.500, 0.003 & 1.000, $2.92\times10^{-6}$ & 2.000, $5.09\times10^{-17}$ & 3.000, $1.62\times10^{-34}$ & 4.500, $1.27\times10^{-73}$ \\
\cline{2-7}
$\circ$ WOR &0.020, 0.031 & 0.231, 0.001 & 0.523, $1.17\times10^{-6}$ & 1.269, $2.04\times10^{-17}$ & 2.156, $6.50\times 10^{-35}$ & 3.600, $5.06\times 10^{-74}$\\
$\circ$ WR & 0.017, 0.033 & 0.194, 0.005 & 0.449, 0.001 & 1.134, $3.59\times10^{-5}$ & 1.987, $2.17\times10^{-6}$ & 3.413, $4.64\times10^{-8}$\\
%$\circ$ MUSTwo & 0.017, 0.033 & 0.194, 0.005 & 0.449, 0.001 & 1.134, $3.59\times10^{-5}$ & 1.987, $2.17\times10^{-6}$ & 3.413, $4.64\times10^{-8}$\\
$\circ$ MUSTow & 0.014, 0.034 & 0.164, 0.008 &0.388, 0.002& 1.015, $1.79\times10^{-4}$ & 1.834, $1.89\times10^{-5}$ & 3.240, $8.28\times10^{-7}$\\
$\circ$ MUSTww & 0.012, 0.034 & 0.145, 0.011 &0.346, 0.004& 0.932, $6.21\times10^{-4}$ & 1.722, $1.31\times10^{-4}$ & 3.111, $1.66\times10^{-5}$\\
\hline
\multicolumn{7}{c}{$\Delta/\sigma=1$}\\
\hline 
Laplace & 0.050, 0.378 & 0.500, 0.221 & 1.000, 0 & 2.000, 0 & 3.000, 0 & 4.500, 0 \\
\cline{2-7}
$\circ$ WOR & 0.020, 0.151 & 0.231, 0.088 & 0.523, 0 & 1.269, 0 & 2.156, 0 & 3.600, 0\\
$\circ$ WR & 0.017, 0.141 & 0.194, 0.093 & 0.449, 0.026 & 1.134, 0.003 & 1.987, $3.17\times 10^{-4}$ & 3.413, $1.45\times 10^{-5}$\\
%$\circ$ MUSTwo & 0.017, 0.141 & 0.194, 0.093 & 0.449, 0.026 & 1.134, 0.003 & 1.987, $3.17\times 10^{-4}$ & 3.413, $1.45\times 10^{-5}$\\
$\circ$ MUSTow & 0.014, 0.132 & 0.164, 0.095 & 0.388, 0.044 & 1.015, 0.010 & 1.834, 0.002 & 3.240, $1.83\times 10^{-4}$\\
$\circ$ MUSTww & 0.012, 0.123 & 0.145, 0.094 & 0.346, 0.052 & 0.932, 0.018 & 1.722, 0.006 & 3.111, 0.001\\
\hline
Gaussian  & 0.050, 0.368 & 0.500, 0.238 & 1.000, 0.127 & 2.000, 0.021 & 3.000, 0.002 & 4.500, $5.87\times10^{-6}$ \\
\cline{2-7}
$\circ$ WOR &0.020, 0.147 & 0.231, 0.095 & 0.523, 0.051 & 1.269, 0.008 & 2.156, $6.15\times 10^{-4}$ & 3.600, $2.35\times 10^{-6}$\\
$\circ$ WR & 0.017, 0.142 & 0.194, 0.103 & 0.449, 0.068 & 1.134, 0.029 & 1.987, 0.015 & 3.413, 0.006\\
%$\circ$ MUSTwo & 0.017, 0.142 & 0.194, 0.103 & 0.449, 0.068 & 1.134, 0.029 & 1.987, 0.015 & 3.413, 0.006\\
$\circ$ MUSTow & 0.014, 0.136 & 0.164, 0.106 &0.388, 0.079& 1.015, 0.045 & 1.834, 0.028 & 3.240, 0.015\\
$\circ$ MUSTww & 0.012, 0.123 & 0.145, 0.094 &0.346, 0.052& 0.932, 0.018 & 1.722, 0.006 & 3.111, 0.001\\
\hline
\end{tabular}}
\end{table}

%----------------------------------------------------
\refstepcounter{suppcounter}
\def\thesection{S\thesuppcounter}
\section{Differentially private stochastic gradient descent algorithm in utility experiments 2 and 3}\label{sec:DP-SGD}
\begin{algorithm}[H]\caption{differentially private stochastic gradient descent algorithm in utility experiments 2 and 3 with overall privacy loss calculation via the FA Algorithm} \label{algo:DP-SGD}
\textbf{Input}: Training data $X=\{\boldsymbol{x_1},\dots,\boldsymbol{x_n}\}$, loss function $l(\boldsymbol{\beta})=\sum_{i=1}^n l(\boldsymbol{\beta},\boldsymbol{x_i})$, subsampling scheme $\mathcal{S}$ (a 2-step MUST, Possion, WR, WOR),  subsampling hyperparameters ($m,b$ for MUST; $m$ for WO and WOR; $\gamma$ for Poisson sampling),  per-iteration privacy loss  $\epsilon'$ (after subsampling) and $\delta$ (before subsampling), overall privacy budget $\epsilon_c$, number of iterations $T$, gradient clipping bound $C$, learning rate at iteration $t$ $\eta_t$ ($t=1,\ldots,T$), convolution truncation bound $L$ and number of grid points  $r$ for the  the FA algorithm\\
\textbf{Output}: privacy-preserving parameter estimate $\hat{\boldsymbol{\beta}}^*$; the overall $\delta_c(\epsilon_c)$ with lower and upper bounds ($\delta_{c,L}, \delta_{c,R}$).
\begin{algorithmic}[1]
\State Calculate the per-query privacy loss $\epsilon$ before  PA given $\epsilon'$ using the results in Theorems 5 and 7;
\State Calculate the scale of the base Gaussian mechanism $\M$ of $(\epsilon,\delta)$-DP for gradient sensitization $\sigma=C\sqrt{2\log(1.25/\delta)}/(n\epsilon)$;
\State Initialize  $\boldsymbol{\beta}_0$; 
\For {$t=1$ to $T$}
\State Take a random subsample $Y$ via $\mathcal{S}$;
\For {$i=1$ to $|Y|$}
\State Compute the gradient $g_t(\boldsymbol{x_i})=\nabla_{\boldsymbol{\beta}_t}l(\boldsymbol{\beta}_t, \boldsymbol{x_i})$;
\State Clip the gradient  $\bar{g}_t(\boldsymbol{x_i})=g_t(\boldsymbol{x_i})/\max(1, ||g_t(\boldsymbol{x_i})||_2/C)$.
\EndFor
\State Add Gaussian noise to the gradient %$\tilde{g}_t=\sum^{|Y|}_{i=1}\bar{g}_t(x_i)/|Y|+\mathcal{N}(0, \sigma^2 \mathbf{I})$.
$\tilde{g}_t\!=\!|Y|^{-1}(\sum^{|Y|}_{i=1}\bar{g}_t(\boldsymbol{x_i})\!+\!(|Y|/m)\mathcal{N}(0, (m\sigma)^2 \mathbf{I}))$.
\State Update parameter $\boldsymbol{\beta}_{t+1}=\boldsymbol{\beta}_{t}-\eta_t\tilde{g}_t$.
\EndFor
\State  $\hat{\boldsymbol{\beta}}^*\leftarrow \boldsymbol{\beta}_T$
\State Calculate $\delta_c(\epsilon_c)$  and its  lower and upper bounds ($\delta_{c,L}, \delta_{c,R}$) using Algorithm 1 with input $\epsilon=\epsilon_c, k=T, L=L, r=r$ and privacy loss pdf $\omega_{X/X'}(s)=f_X(\mathcal{L}^{-1}_{X/{X'}}(s))d\mathcal{L}^{-1}_{X/{X'}}(s)/ds$ over $s\in\mathcal{L}_{X/{X'}}(\mathbb{R})$ ($f_X(\cdot)$ and $\mathcal{L}_{X/{X'}}(s)$ vary by $\mathcal{S}$).
\State \textbf{Return}  $\hat{\boldsymbol{\beta}}^*$, $\delta_c(\epsilon_c)$, and ($\delta_{c,L}, \delta_{c,R}$).
\end{algorithmic} 
\end{algorithm}

%-------------------------------------
\refstepcounter{suppcounter}
\def\thesection{S\thesuppcounter}
\section{Implementation details in the experiments}\label{sec:implementation}
\textbf{Section IV.A: Privacy Amplification Comparison.} We calculate $\epsilon'$ for a range of $\epsilon$ values $\in(0,6)$ based on Table 1 and  the $\delta_j$ and $\delta_u$ terms in the $\delta'$ formulas for WR and MUST using Theorem 1. Specifically, we  plug $\theta=j\Delta_p$ and $\theta=u\Delta_p$ for $p=1,2$ in the formulas to calculate $\delta_j$ and $\delta_u$, respectively.

\textbf{Section IV.B: Utility Analysis.} In each deployment of the subsampled Gaussian mechanism, we pre-specify the same $\epsilon'$, the per-query privacy loss after PA by subsampling, for all subsampling procedures. We then back-calculate the corresponding privacy loss before subsampling $\epsilon$ via the relation $e^{\epsilon'}=1+\eta(e^\epsilon-1)$. Since the PA effect is different across subsampling procedures, $\epsilon$ back-calculated from the same $\epsilon'$ is different for WR, WOR, Poisson, MUSTow,  and MUSTww. 
Fixing $\delta$ before subsampling at $1/n$ for each DP application, we can calculate the standard deviation (SD) $\sigma$ of the noise in the base Gaussian mechanism of $(\epsilon,\delta)$-DP using the privacy profile relations in Theorem 1. Since $\epsilon$ is different across the subsampling procedures, so is $\sigma^2$. The stronger the PA effect a subsampling procedure has, the smaller $\sigma^2$ will be needed for the base Gaussian mechanism to yield the same $\epsilon'$  after being composited with the base mechanism.

\emph{In Utility experiment 1},  $\Delta_2$ is  $(U-L)/n$ for the sample mean and $(U-L)^2/n$ for the sample variance, where $U$ and $L$ are the global upper and lower bounds of the data. We use $[L, U]=[-4, 4]$ where $>99.99\%$ of data points generated from $\mathcal{N}(0, 1)$ would fall within.

%\underline{In Utility experiment 2}, the loss function is $l(\boldsymbol{\beta})=\sum_{i=1}^n(\mathbf{x}_i^T\boldsymbol{\beta}-y_i)^2/n$, where $\mathbf{x}_i=(1, x_{i, 1}, x_{i, 2})^T$. 
%The gradient is $g_t(\boldsymbol{x_i})=2\boldsymbol{x_i}(\boldsymbol{x_i}^{\top}\boldsymbol{\beta}_t-y_i)$. 

\emph{In utility experiment 3 with the adult data}, the $11$ features in the logistic regression model included (age, workclass, fnlwgt, educational number, marital status, race,	gender,	capital gain, capital loss,	hours per week, occupation).  Three attributes relationship, education, and native country were not used since  relationship is highly correlated with gender,  education number contains similar information to the education variable. %and only the subset with  country ``United States'' was used in this experiment. 
The attribute occupation has many detailed categories and we collapsed them into four categories (``Professional'', ``Office Staff'', ``Worker'', and ``Other Service''); similarly for workclass (combined into two categories ``Private'' and ``Others'' ), marital status (``married'' vs ``not-married''),  and race (``White'' vs ``Others'').  We normalized each variable to be within $[0,1]$ by dividing each with its respective maximum (each variable is non-negative in the dataset) for model training. The logistic regression model contains 14 parameters in total.

In this  experiment, we also calculate the overall privacy loss $\delta(\epsilon_c)$ at $\epsilon_c=2$ in the five $\mathcal{S}\circ\M$ procedures for the DP-SGD algorithm after $1000$ iterations using the FFA algorithm with $L=6, r=3\times10^5, k=1,000$ for all the  subsampling schemes except for WR, where $r=3\times10^5$ led to numerical error and $r=2\times10^5$ was used instead.

\emph{In Utility experiment 3 with the MNIST data}, the employed neural network architecture contains two fully connected hidden layers.  The first layer maps the flattened 784-dimensional input (from 28$\times$ 28 pixel images) to a hidden layer with 128 nodes via a ReLU activation function, and the second hidden layer that connects with 10 output outputs also contains 128 nodes. The output of the network is passed through a logarithmic softmax function to compute a normalized output distribution over the classes for classification tasks. The neural network contains 17,802 parameters parameters in total.

\refstepcounter{suppcounter}
\def\thesection{S\thesuppcounter}
\section{Additional Experiment Results in Section IV.B}%\label{sec:DP-SGD}
\begin{table}[!htb]
\vspace{-3pt}\centering
\caption{Bias and RMSE of PP parameter estimates $\hat{\boldsymbol{\beta}}^*=(\hat{\beta}^*_0,\hat{\beta}^*_1,\hat{\beta}^*_2)$ over 200 simulated datasets in utility experiment 2 in Section IV.B} \label{tab:lm_beta_main}\vspace{-3pt}
\resizebox{0.65\textwidth}{!}{
\begin{tabular}{l cc}
\toprule
$\mathcal{S}$ & Bias & RMSE  \\
\hline
&\multicolumn{2}{c}{$\epsilon'=0.01$}\\
\cline{2-3}
Poisson &  -0.0024, 0.0029, 0.0004 & 0.044, 0.044, 0.042\\
WOR &  0.0002, 0.0023, -0.0001 &	0.040, 0.043, 0.041\\
WR &  0.0011, -0.0001, 0.0043 &	0.039, 0.042, 0.043\\
%MUSTwo & 0.0042, -0.00003, 0.0034	&0.045, 0.045, 0.045&& -0.0039, -0.0017, 0.0109 &	0.145, 0.159, 0.154\\
MUSTow &  0.0008, 0.0046, -0.0036
	&0.038, 0.043, 0.042\\
MUSTww &  0.0035, 0.0012, -0.0006
	&0.041,	0.042, 0.044\\
 \cline{2-3}
&\multicolumn{2}{c}{$\epsilon'=0.001$}\\
\cline{2-3}
Poisson  & -0.0038, -0.0055, -0.0122 &	0.152, 0.174, 0.168\\
WOR& -0.0059, 0.0028, -0.0068 & 0.150, 0.164, 0.159\\
WR & 0.0061, -0.0082, -0.0200 & 0.142, 0.164, 0.156\\
MUSTow& 0.0010, -0.0078, -0.0023 & 0.107, 0.125, 0.125\\
MUSTww& 0.0175, -0.0144, 0.0018 & 0.106, 0.125, 0.120\\
\hline
\multicolumn{3}{l}{True parameter $\beta_0, \beta_1, \beta_2 =1, 0.5, 0.2$; for non-private Poisson,}\\
\multicolumn{3}{l}{ bias $= (0.006, 0.001, 0.001)$; RMSE $=(0.037, 0.034, 0.038)$.}\\
\multicolumn{3}{l}{The results for MUSTwo are similar to WR, as expected, }\\
\multicolumn{3}{l}{ except for some numerical difference due to Monte Carlo errors.}\\
\bottomrule
\end{tabular}}\vspace{-6pt}
\end{table}

In the adult experiment, we also calculate the range $(\delta_{c,L},\delta_{c,U}$ for $\delta_c$. It is  $(0,4.17\!\times\!10^{-6}), (0, 3.79\!\times\!10^{-7}), (0, 1.28\!\times\!10^{-6}), (0, 8.49\!\times\!10^{-6}), (0, 1.37\!\times\!10^{-5})$ for Poisson,  WOR,  WR, 
MUSTow, and  MUSTww, respectively. Though there are numerical differences across different subsampling procedures, the  $\delta_c$  values themselves are small enough to make the difference practically ignorable from a privacy perspective. In fact, the upper bound of $\delta_c$  at $\epsilon_c=2$ in every subsampling method was below $ 1/n\approx 3.2\times10^{-5}$. 

\begin{figure}[!htb]
\vspace{-3pt}\centering
{\includegraphics[width=0.7\textwidth]{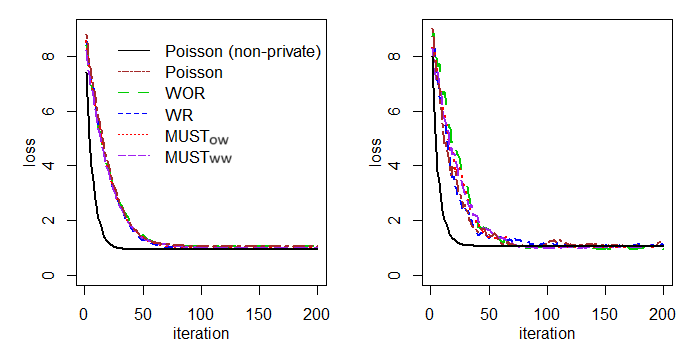}}\\\vspace{-9pt}
$\qquad(a)\; \epsilon'=0.01\mbox{\hspace{1in}} (b)\; \epsilon'=0.001$
%\qquad\epsilon'=0.1
\caption{Training loss for linear regression using DP-SGD in a single dataset in utility experiment 2 in Section IV.B}\label{fig: lm_loss_main}\vspace{-6pt}
\end{figure}

\begin{figure}[!htb]
\vspace{-6pt}\centering
{\includegraphics[width=0.49\textwidth, trim=12pt 32pt 16pt 32pt, clip]{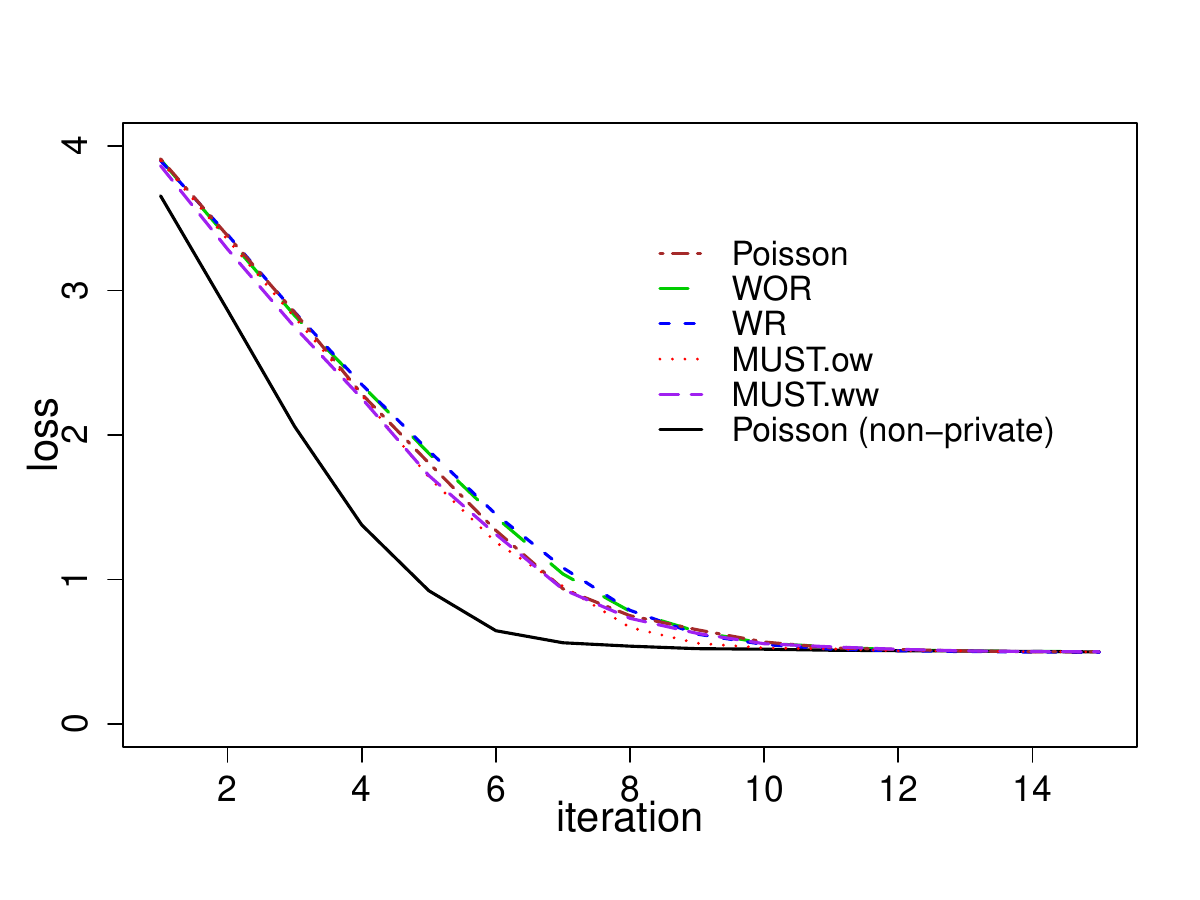}}
\includegraphics[width=0.49\linewidth, trim= 0.5in 0.1in 0.8in 0.6in, clip]{figures/final_loss_plot.pdf}\\
(a) Adult Data \hspace{2in}(b) MNIST data
\caption{Training loss of the logistic regression in Utility experiment 3 in Section IV.B}\label{fig: logistic loss}
\vspace{-6pt}\end{figure}

\refstepcounter{suppcounter}
\def\thesection{S\thesuppcounter}
\section{Computtional complexity of R functions \texttt{sample} and \texttt{rbinom}}%\label{sec:DP-SGD}
When sampling without replacement, each element is drawn from the remaining pool, which decreases in size after each selection. The algorithm is essentially a shuffle and draw process. %The Fisher-Yates shuffle (or Knuth shuffle)
The time complexity of  \texttt{sample(x, m, replace = F)} for sampling $m$ elements without replacement from a set of size $n$ is $O(n$) as the algorithm  shuffles an array $x$ in $O(n)$ time before selecting the first $m$ elements. This complexity is independent of $m$, meaning that even if you're sampling only a few elements ($k << n$), the entire vector is still shuffled first.
In the case of sampling with replacement, each element is selected independently, with the full set available for each draw. That is, generating a sample of size $m$ requires $m$ independent random selections, each taking constant time.  The time complexity of  \texttt{sample(x, m, replace = T)} for is $O(m)$.
For Poisson sampling, which often refers to sampling each element independently with the same probability $\gamma$ in the PA literature we apply \texttt{rbinom($n$, 1, prob=$\gamma$)} in R to draw sample from Bernoulli($\gamma$) and its time complexity is $O(n)$.

\end{document}